\documentclass[times, 10pt]{elsarticle}

\usepackage{amssymb}
\usepackage{amsthm}

\usepackage{orcidlink}
\usepackage{subcaption}
\usepackage{algpseudocode}
\usepackage[ruled,vlined,linesnumbered]{algorithm2e}
\usepackage{amsmath}
\usepackage{amsfonts}
\usepackage{physics}
\usepackage[capitalize,noabbrev]{cleveref}
\usepackage{orcidlink}
\usepackage{booktabs}
\usepackage{duckuments}
\usepackage{enumitem}
\usepackage[T1]{fontenc}
\usepackage[title]{appendix}
\usepackage{dsfont}
\usepackage{graphicx}
\usepackage{xcolor}
\usepackage{multirow}
\usepackage{array}
\usepackage{lipsum}
\usepackage{setspace}
\usepackage{makecell}

\newcommand{\pif}{\textsc{PIF}\xspace}
\newcommand{\spif}{Sliding-\textsc{PIF}\xspace}
\newcommand{\pspace}{Preference Space\xspace}
\newcommand{\pembedding}{Preference Embedding\xspace}
\newcommand{\pisolation}{Preference Isolation\xspace}
\newcommand{\mhash}{\textsc{MinHash}\xspace}
\newcommand{\rzhash}{\textsc{RuzHash}\xspace}

\newcommand{\vitrees}{Voronoi-\textsc{iTrees}\xspace}
\newcommand{\vitree}{Voronoi-\textsc{iTree}\xspace}
\newcommand{\rzhitree}{\rzhash-\textsc{iTree}\xspace}

\newcommand{\itree}{\textsc{iTree}\xspace}

\newcommand{\viforest}{Voronoi-\textsc{iForest}\xspace}
\newcommand{\rzhiforest}{\rzhash-\textsc{iForest}\xspace}
\newcommand{\iforest}{\textsc{iForest}\xspace}
\newcommand{\vifor}{\textsc{ViFor}\xspace}
\newcommand{\rzhifor}{\textsc{RzHiFor}\xspace}
\newcommand{\ifor}{\textsc{iFor}\xspace}
\newcommand{\eifor}{\textsc{EiFor}\xspace}
\newcommand{\lof}{\textsc{LOF}\xspace}
\newcommand{\lsh}{\textsc{LSH}\xspace}

\DeclareRobustCommand{\vect}[1]{
  \ifcat#1\relax
    \boldsymbol{#1}
  \else
    \vb*{#1}
\fi}

\newtheorem{theorem}{Theorem}
\newtheorem{corollary}{Corollary}
\newtheorem{lemma}{Lemma}
\journal{Pattern Recognition}
\begin{document}
    \begin{frontmatter}
        \title{Preference Isolation Forest for Structure-based Anomaly Detection}
        
        \author[polimi]{Filippo Leveni\texorpdfstring{\corref{fl}}{*}\orcidlink{0009-0007-7745-5686}}
        \ead{filippo.leveni@polimi.it}
        \cortext[fl]{Corresponding author.}
        \author[polimi]{Luca Magri\orcidlink{0000-0002-0598-8279}}
        \ead{luca.magri@polimi.it}
        \author[polimi,usi]{Cesare Alippi\orcidlink{0000-0003-3819-0025}}
        \ead{cesare.alippi@polimi.it}
        \author[polimi]{Giacomo Boracchi\orcidlink{0000-0002-1650-3054}}
        \ead{giacomo.boracchi@polimi.it}
        \affiliation[polimi]{organization={Department of Electronics, Information and Bioengineering, Politecnico di Milano},
                             addressline={Piazza Leonardo da Vinci, 32},
                             city={Milano},
                             postcode={20133},
                             country={Italy}}
        \affiliation[usi]{organization={Faculty of Informatics, Università della Svizzera Italiana},
                          addressline={Via Giuseppe Buffi, 13},
                          city={Lugano},
                          postcode={6900},
                          country={Switzerland}}
        
        \begin{abstract}
            We address the problem of detecting anomalies as samples that do not conform to structured patterns represented by low-dimensional manifolds.
            To this end, we conceive a general anomaly detection framework called Preference Isolation Forest (\pif), that combines the benefits of adaptive isolation-based methods with the flexibility of preference embedding. The key intuition is to embed the data into a high-dimensional preference space by fitting low-dimensional manifolds, and to identify anomalies as isolated points.
            We propose three isolation approaches to identify anomalies: \emph{i}) \viforest, the most general solution, \emph{ii}) \rzhiforest, that avoids explicit computation of distances via Local Sensitive Hashing, and \emph{iii}) \spif, that leverages a locality prior to improve efficiency and effectiveness. 
        \end{abstract}
        
        \begin{graphicalabstract}
            \includegraphics[width=\linewidth]{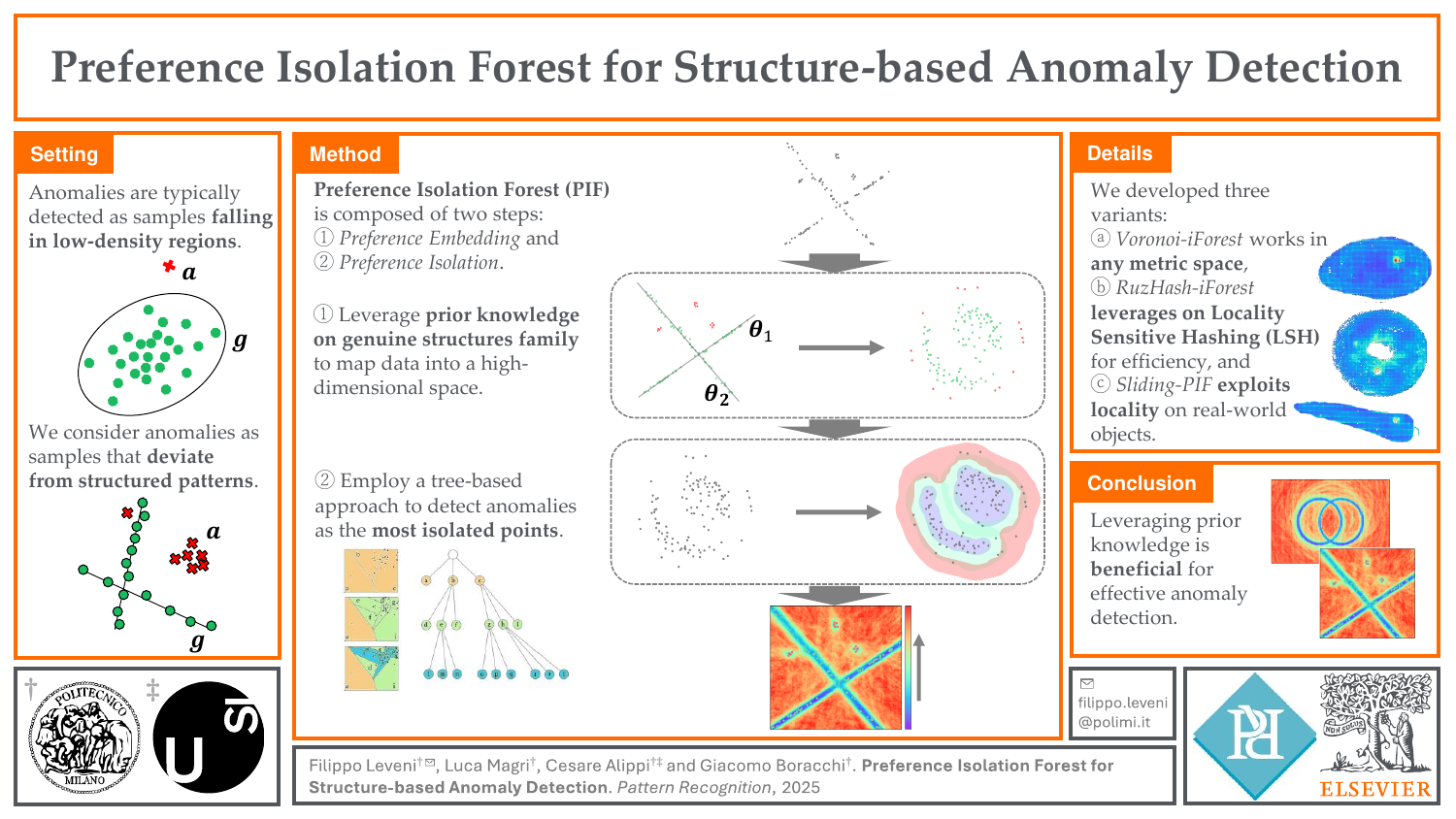}
        \end{graphicalabstract}
        
        \begin{highlights}
            \item We propose a general framework to detect anomalies with respect to parametric structured patterns.
            \item We enhance separability between genuine and anomalous data via embedding into a high-dimensional preference space, leveraging on prior knowledge on genuine data.
            \item We detect anomalies as the most isolated in the embedding, by proposing tailored isolation-based methods able to exploit the correct distance measures in the preference space.
            \item We propose a sliding window extension able to deal with unknown but smooth patterns by leveraging on locality priors.
            \item Extensive experiments on synthetic and real-world data demonstrate the effectiveness of structure-based anomaly detection.
        \end{highlights}
        
    \end{frontmatter}

    \section{Introduction}
        \label{sec:introduction}

        \begin{figure}[t]
            \centering
            \begin{subfigure}[t]{.325\linewidth}
                \centering
                \includegraphics[height=.8\linewidth]{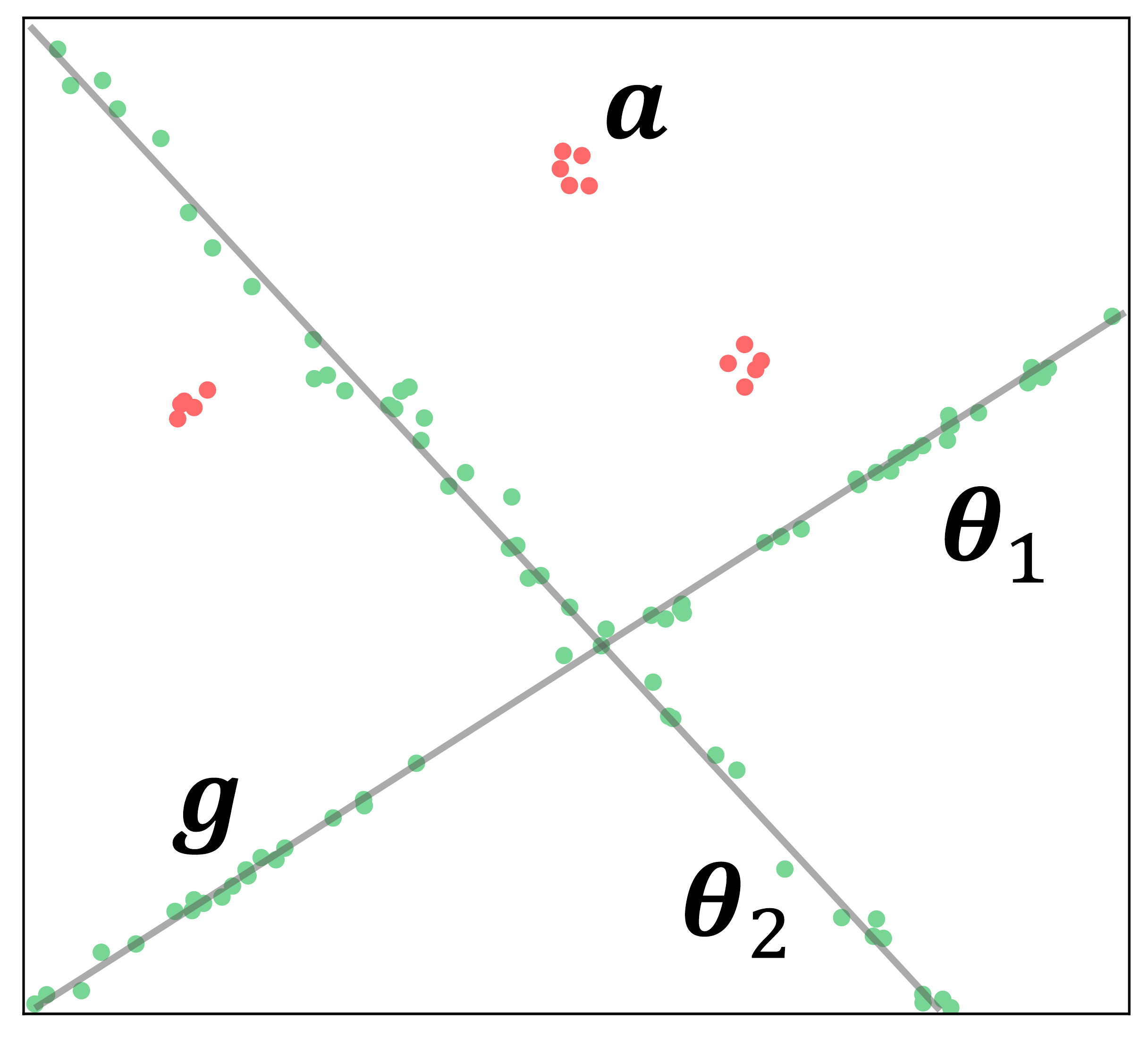}
                \caption{Input data $X = G \cup A$.}
                \label{subfig:input}
            \end{subfigure}
            \hfill
            \begin{subfigure}[t]{.325\linewidth}
                \centering
                \hspace{-0.4cm}
                \includegraphics[height=.8\linewidth]{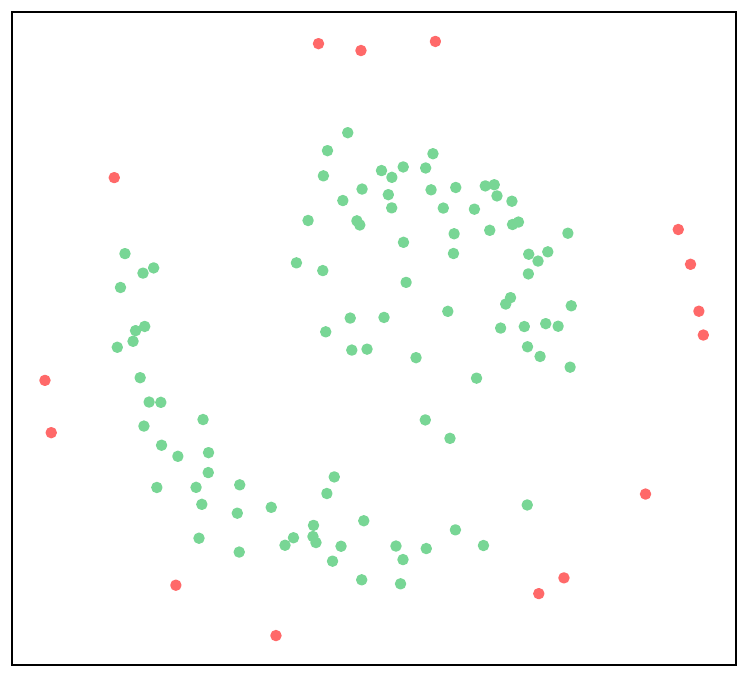}
                \caption{Preference space $\mathcal{P}$.}
                \label{subfig:embedding}
            \end{subfigure}
            \hfill
            \begin{subfigure}[t]{.325\linewidth}
                \centering
                \includegraphics[height=.8\linewidth]{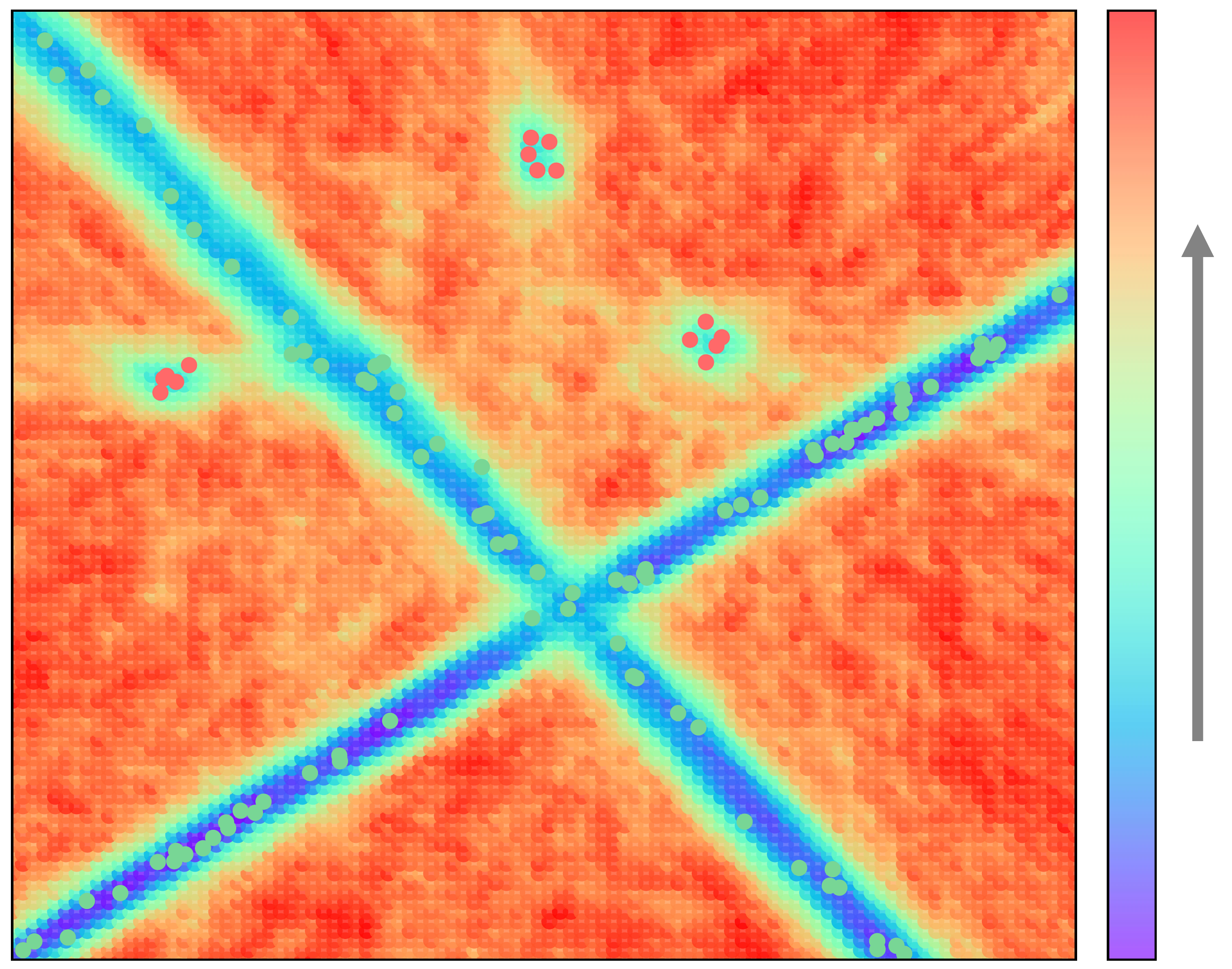}
                \caption{Anomaly scores $\alpha$.}
                \label{subfig:output}
            \end{subfigure}
            \caption{\pif detects anomalies that deviate from structures. (a) Genuine points $G$, in green, described by two lines $\vect{\theta}_1$ and $\vect{\theta}_2$, and anomalies $A$ in red. (b) \pembedding maps $X$ to a high-dimensional space where anomalies result in isolated points (visualized via MultiDimensional Scaling). (c) Anomaly scores $\alpha(\cdot)$ computed via \pisolation.}
            \label{fig:input_output}
        \end{figure}

        \emph{Anomaly detection} (AD) deals with the problem of identifying data that do not conform to an expected behavior~\cite{ChandolaBanerjeeAl09}.
        In the statistical and data-mining literature, \emph{anomalies} are typically detected as samples falling in low-density regions~\cite{ChandolaBanerjeeAl09}. However, in many real-world scenarios, genuine data lie on low-dimensional manifolds and anomalies are best characterized in terms of non-conformity to those structures.
        For example, in 3D scans~\cite{BergmannXinAl22} anomalies can be recognized as defects that deviate from the smooth surfaces that characterized the object. This remains true regardless of their local density, which may vary depending on the distance from the sensing device. In such a situation, the density analysis in the ambient space becomes misleading: scanning artifacts may lie in a dense region but still be structurally inconsistent with the surrounding data.
        Similarly, in intrusion detection~\cite{ZhongLinAl24}, where the purpose is to identify faces of unauthorized subjects, the images of intruders do not conform with the subspace of images of known subjects. In this context, it is the deviation from this subspace -- rather than local data density -- that serves as the key indicator for detecting anomalies.
        
        \emph{Structure-based AD} addresses these situations where a \emph{prior}, expressed as a model family $\mathcal{F}$, describes genuine data and can be used to detect anomalies in a truly \emph{unsupervised} way, \emph{i.e.}, without any training data acting as a reference for genuine or anomalous data and without involving any training process. Structure-based AD overcomes the shortcoming of density- and distance-based techniques, which operate in the ambient data space and struggle when data lie on complex manifolds, and the shortcomings of isolation-based methods, which rely on random axis-aligned splits that ignore the implicit  structure of genuine data.
        Structure-based AD also contrasts with the \emph{supervised} or \emph{semi-supervised} neural network-based AD approaches, which require a large amount of data -- often unavailable or expensive to obtain -- to infer the distribution of genuine data and cannot easily incorporate prior knowledge. In fact, our approach effectively leverages an explicit analytic prior, making the anomaly detection process more controllable, interpretable, and independent on data-driven training.

        \cref{subfig:input} illustrates a simplified example, where the prior characterizing genuine data is the family $\mathcal{F}$ of lines, while anomalies are points deviating from instances of $\mathcal{F}$, and a density-based AD solution would fall short.
        In pattern recognition, rather than lines, other model families $\mathcal{F}$ describe structures or regularities in genuine data. For instance, $\mathcal{F}$ can be rototranslations in 3D registration~\cite{BeslMcKay92} and homographies in template matching~\cite{Brunelli09}. Here, anomalies, \emph{a.k.a.} outliers, are often detected as a byproduct of a multi-model fitting process~\cite{MagriLeveniAl21}, where structures of inliers are first identified, and points that deviate are labeled as anomalous. Unfortunately, when multiple local structures are present, solving complex multi-model fitting problems is computationally expensive and sensitive to initialization or model assumptions. When AD is the primary goal, is preferable to employ algorithms that direclty detect anomlaies thus sidestespping the issues of multi-model fitting.
       
        In this work, we present Preference Isolation Forest (\pif), a novel and general unsupervised AD framework designed to identify anomalies in terms of the \emph{preferences} they grant to genuine structures described by instances of a parametric model family $\mathcal{F}$. \pif allows for the integration of prior knowledge about genuine data, enhancing AD without requiring any training data nor trained model. \pif is composed of two main components: \emph{i}) \pembedding, to map data to a high-dimensional space, called \pspace, by fitting models from $\mathcal{F}$ (from~\cref{subfig:input} to~\cref{subfig:embedding}), and \emph{ii}) \pisolation, to detect anomalies as the most \emph{isolated} points in the \pspace (from~\cref{subfig:embedding} to~\cref{subfig:output}). At high level, the rationale of our approach is that the preference space  captures structural consistency of points, without requiring the models to be identified first and avoiding the need of ground-truth labels or supervised training. We demonstrate that in the \pspace, when endowed with an appropriate distance (e.g., Jaccard, Tanimoto or Ruziska), anomalies naturally emerge as isolated points due to their lack of consensus with any model.
        We propose three main tree-based algorithms for \pisolation: \emph{a}) \viforest, that partitions the space via nested Voronoi tessellations. This is the most general one, as it can cope with any distance defined in the preference space. Then, when efficiency is a concern, we present \emph{b}) \rzhiforest, that exploit  \rzhash, a Locality Sensitive Hashing (LSH)~\cite{GionisIndyk99} scheme we designed for the \pspace to achieve a speed-up factor of $\times 35\%$ to $\times 70\%$ over \viforest.
        Finally, since in real scenario the model family $\mathcal{F}$ might  only approximates the genuine structures locally (see~\cref{subfig:models_manifold_local_family}), we introduce \emph{c})  \spif a sliding window mechanism to leverage the spatial proximity in AD problems for spatial data. \spif promotes locality, and enables scalability of AD tasks on large datasets by solving many smaller, localized AD problems, instead of a large one.
        
        This work extends~\cite{LeveniMagriAl21,LeveniMagriAl23} which introduced \pif and \rzhiforest respectively. In particular, regarding \pif, we provide a comprehensive overview  along with a stronger and broader experimental assessment. For \rzhiforest, we formally prove the correctness of the \rzhash scheme, reinforcing its theoretical foundations. Moreover, the design of the \spif algorithm is completely novel.
        
        All in all, our contributions can be summarized as follows:
        \begin{itemize}
        \item We introduce \pif, a \emph{general framework} for structure-based anomaly detection that leverages prior knowledge through \pembedding, enabling truly \textbf{unsupervised} AD without explicit model fitting.
        \item Building on this, we develop \viforest, a \emph{generic} isolation-based method for arbitrary metric spaces, suitable for preference-based representations.
        \item To improve efficiency, we propose \rzhiforest, which replaces explicit distance computations in \viforest with a tailored LSH scheme for the \pspace.
        \item Finally, we extend \pif with \spif, incorporating data locality to enhance scalability and accuracy in settings with spatially varying structures.
        \end{itemize}

    \section{Problem Formulation}
        \label{sec:problem_formulation}
        \begin{figure}[t]
            \centering
            \begin{subfigure}[t]{.275\linewidth}
                \centering
                \hspace*{-0.6cm}
                \includegraphics[height=\linewidth]{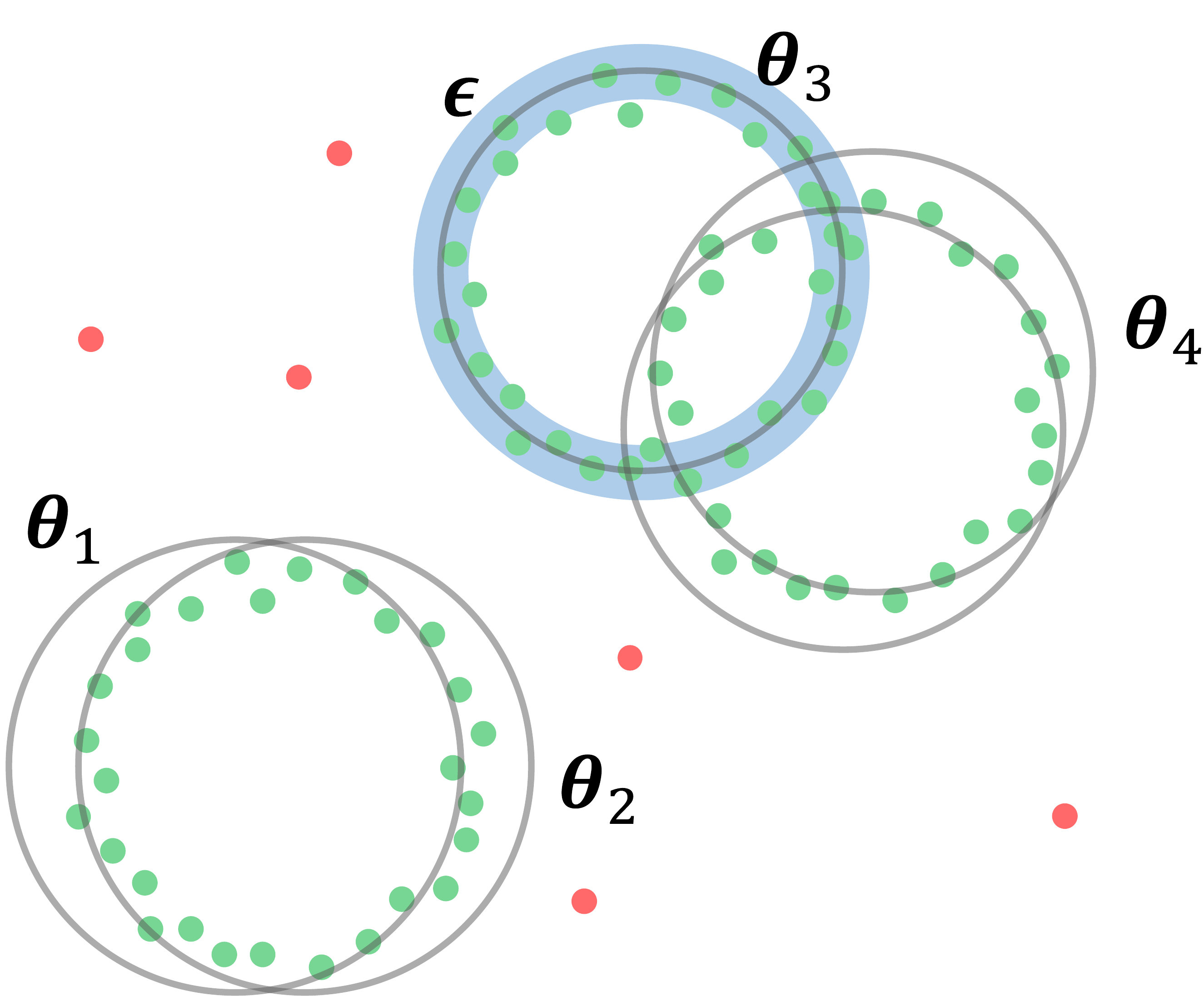}
                \caption{}
                \label{subfig:models_global_family}
            \end{subfigure}
            \hfill
            \hspace*{0.6cm}
            \begin{subfigure}[t]{.3\linewidth}
                \centering
                \hspace*{-0.6cm}
                \includegraphics[height=\linewidth]{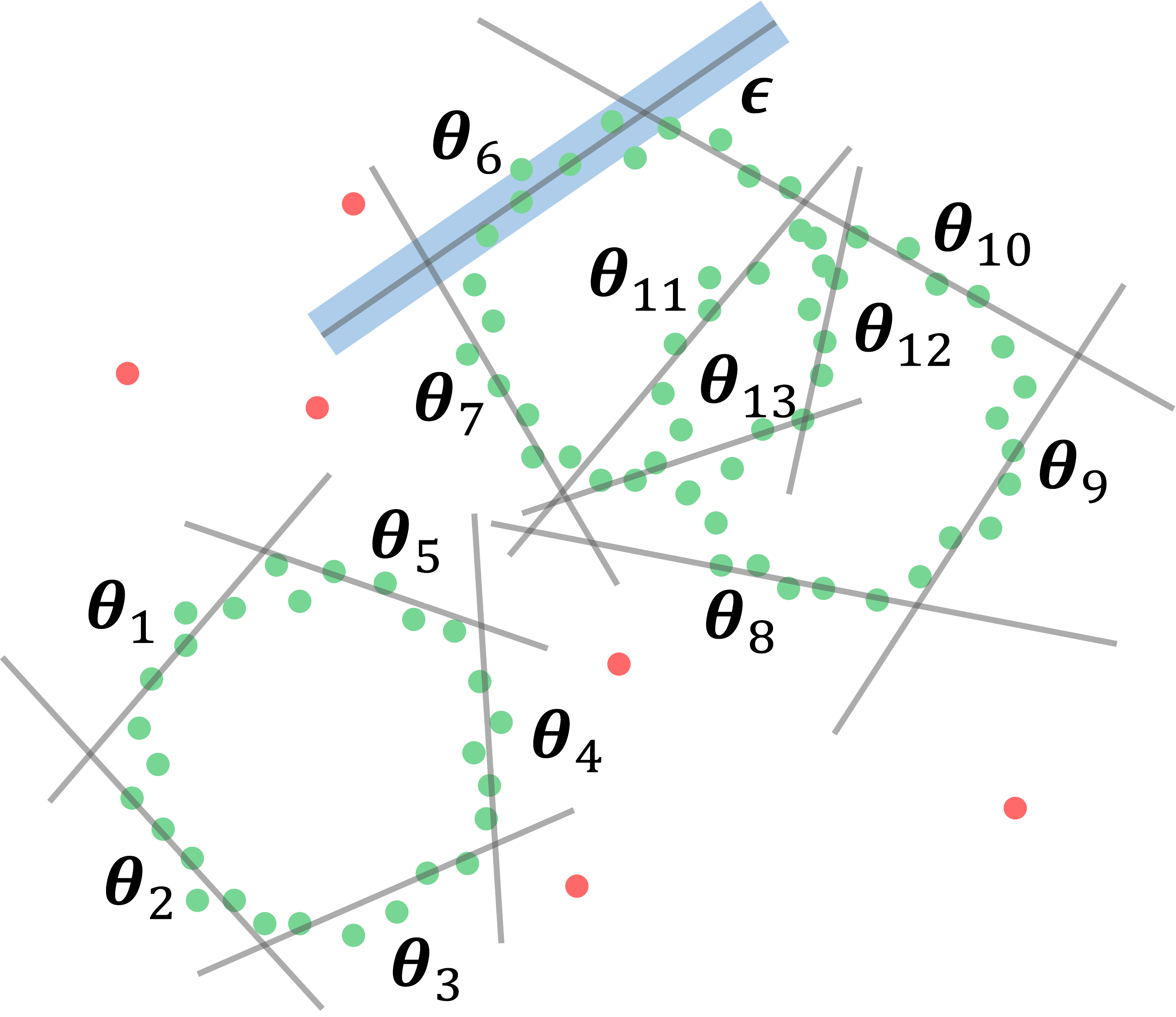}
                \caption{}
                \label{subfig:models_local_family}
            \end{subfigure}
            \hfill
            \hspace*{0.2cm}
            \begin{subfigure}[t]{.325\linewidth}
                \centering
                \includegraphics[height=\linewidth]{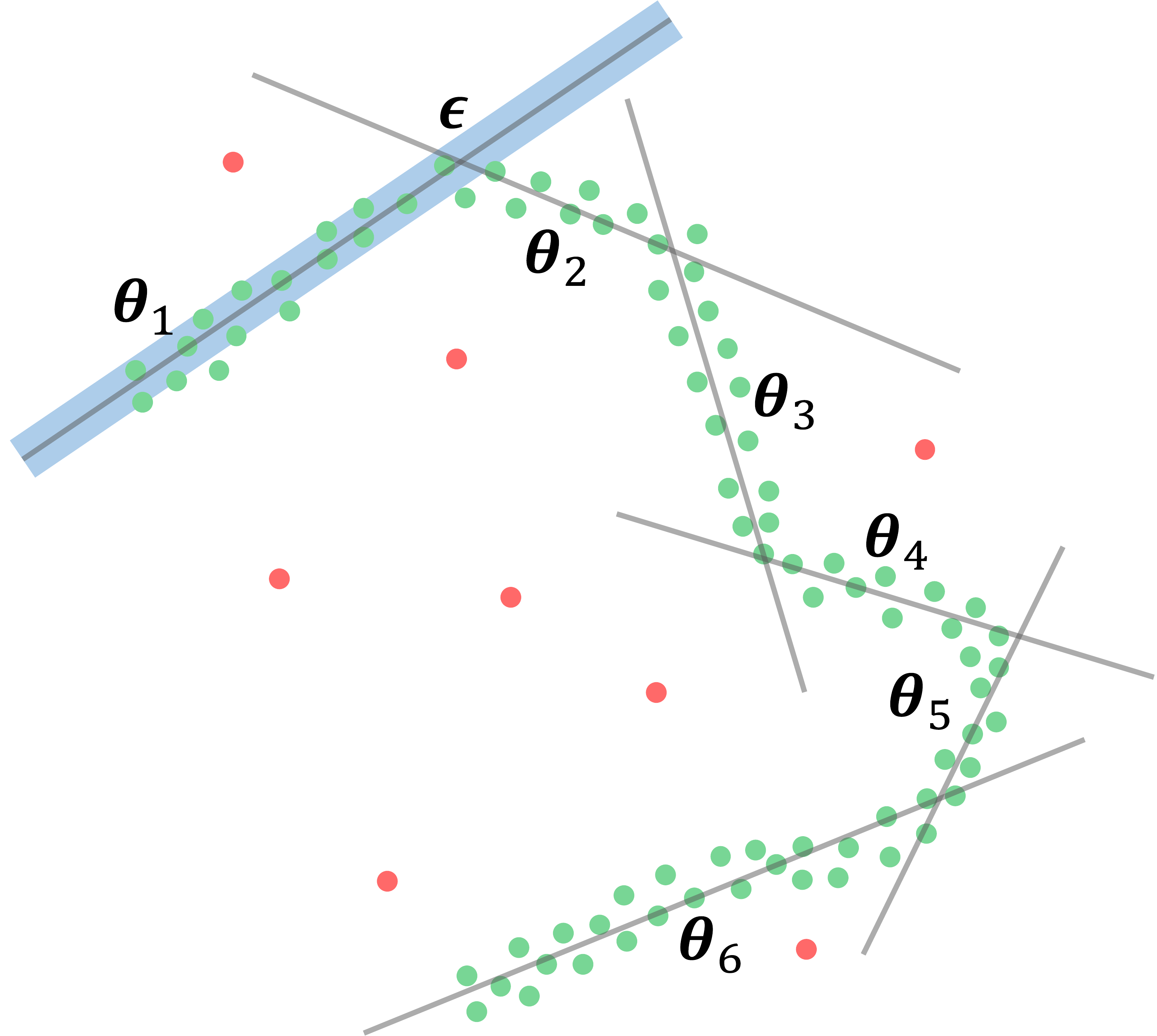}
                \caption{}
                \label{subfig:models_manifold_local_family}
            \end{subfigure}
            \caption{The same data set $G$ can be described by different model families $\mathcal{F}$, circles in (a) and lines in (b). (c) Unknown complex patterns can be described by models that locally approximate the underlying manifold of genuine data. The inlier thresholds for some models are indicated in blue.}
            \label{fig:models_family}
        \end{figure}

        We consider a data set $X = G \cup A \subset \mathcal{X}$, where $G$ and $A$ are the sets of genuine and anomalous data respectively, and $\mathcal{X}$ is the ambient space. We assume that $G$ can be described by a family $\mathcal{F}$ of known parametric models whose number is unknown, while anomalies $A$ cannot.
        Specifically, each $\vect{g} \in G$ is close to the solutions of a parametric equation $\mathcal{F}(\vect{g}, \vect{\theta}) \approx 0$, that depends on unknown parameters $\vect{\theta}$. For example, in~\cref{subfig:models_global_family}, $\mathcal{F}$ is the family of circles $\mathcal{F}(\vect{g}, \vect{\theta}) = (\theta_1 - g_1)^2 + (\theta_2 - g_2)^2 - \theta_3$ and, due to noise, genuine data satisfy the equation up to a tolerance, namely $|\mathcal{F}(\vect{g}, \vect{\theta})| < \epsilon$ (indicated in~\cref{fig:models_family} in blue), where $\epsilon > 0$ depends on the noise, whose distribution is unknown. On the other hand, anomalous data $\vect{a} \in A$ poorly satisfy the equation, \emph{i.e.}, $|\mathcal{F}(\vect{a}, \vect{\theta})| \gg \epsilon$.
        Note that the same data $G$ can be described by different model families. In~\cref{subfig:models_local_family}, $\mathcal{F}$ consists of lines, expressed as $\mathcal{F}(\vect{g}, \vect{\theta}) = \theta_1 g_1 + \theta_2 g_2 + \theta_3$. The flexibility of our formulation allows $\mathcal{F}$ to \textit{locally} describe complex structures whose underlying manifold is unknown as in \cref{subfig:models_manifold_local_family}, where $G$ is approximated by several lines in $\mathcal{F}$ up to $\epsilon$.
        
        The goal of Structured-based AD is to produce an anomaly scoring function $\alpha: X \rightarrow \mathbb{R}^+$ such that $\alpha(\vect{a}) \gg \alpha(\vect{g}) \ \forall \vect{a} \in A, \; \forall \vect{g} \in G$, so that anomalies are detected by setting an appropriate threshold (\cref{subfig:output}).

    \section{Related Work}
        \label{sec:related_work}
        
        Several AD approaches have been proposed, as extensively described in~\cite{ChandolaBanerjeeAl09}. A possible taxonomy envisages four main categories: distance-based, density-based, isolation-based, and model-based including the approaches that exploit deep neural networks.
        
        In \emph{distance-based AD}~\cite{SamariyaThakkar23} approaches, such as K-Nearest Neighbors (K-NN)~\cite{HalderUddinAl24}, a point is considered anomalous when its $k$-nearest neighborhood is very large, thereby implicitly assuming that anomalies lie in low-density regions. Additionally, they cannot exploit prior knowledge about the family $\mathcal{F}$, and when data lie on structured manifolds (\emph{e.g.}, lines, planes, or more complex geometric models), low-density assumption break down. While distance-based methods are independent from data distribution and can be improved by adopting data-dependent distances~\cite{TingZhuAl16}, they are computationally intensive and perform poorly with high-dimensional data.
        \emph{Density-based AD} methods, like \lof~\cite{BreunigKriegel00}, identify anomalies by means of their local density, and are closely related to distance-based ones. Density-based AD methods share the computational drawbacks of distance-based ones when dealing with high-dimensional data and cannot leverage prior knowledge about family $\mathcal{F}$ as well.
        
        \emph{Isolation-based AD} methods can be traced back to Isolation Forest~\cite{LiuTingAl12} (\ifor), where anomalies are detected as the most \textit{``isolated''} points. \ifor builds a forest of random trees (\itree) that recursively partition the space, and the number of splits necessary to \emph{isolate} a point is inversely related to its anomaly score. Several extensions of \ifor were proposed: Extended Isolation Forest~\cite{HaririKindAl21} and Generalized Isolation Forest~\cite{LesoupleBaudoinAl21} overcome the limitation of axis-parallel splits, while~\cite{StaermanMozharovskyiAl19} extends \ifor to detect functional anomalies. \ifor was also extended to directly operate on pairwise point distances~\cite{MensiTaxAl23}, or by different anomaly scores computation~\cite{MensiBicego21}. Analogously to distance and density-based ones, isolation-based methods do not take into account the underlying structure of the data, and cannot leverage priors about model family $\mathcal{F}$.
        
        \emph{Model-based AD} often treat anomaly detection as a byproduct of another task (\emph{e.g.}, clustering), rather than as a primary objective, by first fitting a model on the data and then detecting anomalies as points that do not conform to it. When multiple or local structures are present, they typically require solving complex (multi-)model fitting problems, which are computationally expensive and sensitive to initialization or model assumptions. Examples of this approach are classification-based~\cite{XuWangAl24,QiaoTongAl25}, reconstruction-based~\cite{ZuoWuAl24,KorniszukSawicki24}, and clustering-based methods~\cite{XuWangAl24a}. The model-based category also encompasses deep learning methods for anomaly detection. These approaches often rely on large amounts of labeled or genuine training data, which are often unavailable or expensive to obtain.
        For example, recent zero-shot methods such as PointAD~\cite{ZhouYanAl24} or MVP-PClip~\cite{ChengCaoAl24}, which, like our work, claim to operate without target-domain genuine data, rely on significantly different assumptions. Despite being labeled ''zero-shot``, they are based on extensively pre-trained Vision-Language Models like CLIP~\cite{RadfordKimAl21} to incorporate data-driven priors. These methods cannot incorporate explicit parametric structural priors (\emph{e.g.}, geometric constraints), which are central to many real-world tasks (\emph{e.g.}, 3D reconstruction and geometric registration). Therefore, although operating without any training example, deep learning approaches fall outside the scope of our work, which focuses on leveraging explicit structural priors, that are controllable and can be chosen to define the characteristics of genuine data.

        Our proposed \emph{structure-based AD} method \pif stands out from these approaches as it allows us to incorporate prior knowledge about the model family $\mathcal{F}$. This feature enables \pif to operate in a truly \emph{unsupervised} manner, without requiring any training data nor trained model, thereby contrasting with the \emph{supervised} or \emph{semi-supervised} approach typical of classification and reconstruction-based methods. Additionally, \pif directly targets AD, in contrast to clustering-based methods that address it as a byproduct.

    \section{Method}
        \label{sec:method}

        \begin{algorithm}[t]
            \footnotesize
            \setstretch{1.35}
            \caption{Preference Isolation Forest (\pif) \label{alg:main}}
            \DontPrintSemicolon
            \SetNoFillComment
            \KwIn{$X$ - data, $t$ - number of trees, $\psi$ - sub-sampling size, $b$ - branching factor}
            \KwOut{$\{\alpha_\psi(\vect{x}_j)\}_{j=1,\ldots,n}$ - anomaly scores}
            \begin{small}
                \tcc{\pembedding\!\!\!}
            \end{small}
            Sample $m$ models $\{\vect{\theta}_i\}_{i=1,\ldots,m}$ from $X$ \label{line:begin_embedding}\\
            $P \leftarrow \{\vect{p}_{\vect{x}_j} \,|\, \vect{p}_{\vect{x}_j} = \mathcal{E}(\vect{x}_j)\}_{j=1,\ldots,n}$ \label{line:end_embedding}\\
            \begin{small}
                \tcc{\viforest construction\!\!\!}
            \end{small}
            $F \leftarrow \text{\viforest}(P, t, \psi, b)$ \label{line:forest_construction}\\
            \begin{small}
                \tcc{Anomaly scores computation}
            \end{small}
            \For{$j = 1$ \normalfont{to} $n$ \label{line:begin_detection}}
                {$D_{\vect{x}_j} \leftarrow \emptyset$ \label{line:set} \\
                 $\vect{p}_{\vect{x}_j} \leftarrow \text{$j$-th point in $P$}$ \label{line:point} \\
                 \For{$k = 1$ \normalfont{to} $t$}
                     {$T_k \leftarrow \text{$k$-th \vitree in $F$}$ \label{line:tree} \\
                      $D_{\vect{x}_j} \leftarrow D_{\vect{x}_j} \cup \text{\textsc{PathLength}}(\vect{p}_{\vect{x}_j}, T_k, 0)$ \label{line:height}}
                $\alpha_\psi(\vect{x}_j) \leftarrow 2^{-\frac{E(D_{\vect{x}_j})}{c(\psi)}}$ \label{line:anomaly_score}
                }
            \Return $\{\alpha_\psi(\vect{x}_j)\}_{j=1,\ldots,n}$ \label{line:end_detection} \\
        \end{algorithm}
        
        Our Preference Isolation Forest (\pif) is a general framework for Structure-based Anomaly Detection that leverages parametric priors regarding genuine data structures in order to make unsupervised anomaly detection more effective. \pif is composed of two steps: \emph{i}) \pembedding (\cref{subsec:preference_embedding}), which maps points $X$ to a high-dimensional \pspace by means of a pool of parametric models $\{\vect{\theta}_i\}_{i = 1, \dots, m}$ belonging to the family $\mathcal{F}$, and \emph{ii}) \pisolation (\cref{subsec:preference_isolation}), which assigns an anomaly score $\alpha(\vect{x})$ to each point $\vect{x}$ exploiting an isolation mechanism.
        At the core of \pisolation lies the construction of a forest, which can be instantiated through three  variants, each designed to address different practical needs and scalability constraints.
        We introduce first \viforest, a novel and general-purpose isolation-based method that serves as the backbone of our approach. Its key strength consists in supporting arbitrary distance functions between preferences through the use of Voronoi tessellations. Building upon this, we introduce \rzhiforest (\cref{subsec:efficiency}), which retains the general philosophy of \viforest but is explicitly designed to improve efficiency in large-scale settings. By leveraging LSH-based partitioning via the \rzhash mechanism, \rzhiforest significantly reduces computational cost and inference time, making it ideal for scenarios with massive datasets or real-time requirements.
        Finally, to address scenarios where genuine structures vary locally across the input space (\emph{e.g.}, 3D surfaces), we propose \spif (\cref{subsec:sliding_pif}). This variant enhances both scalability and accuracy of \pif by operating within spatially overlapping windows, allowing adaptation to locally structured patterns that global methods may overlook. The key features of each variants, from the generality and flexibility of \viforest, to the efficiency of \rzhiforest, and ultimately to the local adaptability of \spif are detailed in ~\cref{tab:differences_methods}.

        \begin{table}[t]
            \footnotesize
            \centering
            \caption{Differences between our proposed methods \viforest, \rzhiforest and \spif.}
            \resizebox{0.7\textwidth}{!}{
            \begin{tabular}{c||c|c|c|}
                             & Purpose                  & Key Feature                                                           & Scenario addressed                                                \\ \hline
                \viforest    & Accuracy \& generality   & \makecell{Supports arbitrary distances \\ via Voronoi tessellations}  & \makecell{General purpose Structure-based \\ Anomaly Detection}   \\ [.3cm]
                \rzhiforest  & Efficiency               & \makecell{Employs LSH-based \\ partitions via \rzhash}                & \makecell{Large-scale problems, \\ faster inference}              \\ [.3cm]
                \spif        & Scalability \& locality  & \makecell{Operates in spatially \\ overlapping windows}               & \makecell{Locally structured data \\ (\emph{e.g.}, 3D surfaces)}  \\ [.3cm]
            \end{tabular}
            }
            \label{tab:differences_methods}
        \end{table}
        
        \subsection{\pembedding}
            \label{subsec:preference_embedding}
            
            \pembedding is a function $\mathcal{E}\colon \mathcal{X} \to \mathcal{P}$ that maps data $X$ from the ambient space $\mathcal{X}$ to a high-dimensional \pspace $\mathcal{P} = [0, 1]^m$ by leveraging prior knowledge on the parametric family $\mathcal{F}$ that well describes genuine data $G$.
            Such mapping is obtained by fitting a pool $\{\vect{\theta}_i\}_{i=1,\ldots,m}$ of $m$ models (\cref{alg:main}, line~\ref{line:begin_embedding}) from family $\mathcal{F}$ on data $X$ via a RanSaC-like strategy~\cite{FischlerBolles81}. We sample the minimal sample set -- \emph{i.e.}, the minimum number of points to constrain a model in $\mathcal{F}$ -- and fit parameters $\vect{\theta}_i$.
            Then (line~\ref{line:end_embedding}), we embed each sample $\vect{x} \in \mathcal{X}$ to a vector $\vect{p}_{\vect{x}} = \mathcal{E}(\vect{x}) \in \mathcal{P}$ whose $i$-th component is defined as:
            \begin{equation}
                \label{eq:preference}
                p_{\vect{x}, i}  =
                \begin{cases}
                    \phi(\delta_{\vect{x}, i}) &\text{if $|\delta_{\vect{x}, i}| \leq \epsilon$}\\
                                    0 &\text{otherwise}
                \end{cases},
            \end{equation}
            where $\delta_{\vect{x}, i} = \mathcal{F}(\vect{x}, \vect{\theta}_i)$ is the residual of $\vect{x}$ with respect to $\vect{\theta}_i$, and $\epsilon = k\sigma$ is an inlier threshold proportional to the noise standard deviation $\sigma > 0$. We assume the amount of noise to be unknown, but directly estimable from data (\emph{e.g.},~\cite{WangSuter04}). As a consequence, setting the inlier threshold can often be treated as an educated guess, guided by established statistical heuristics. A common practice is to choose $k$ such that a desired percentage of inliers fall below the threshold—following, for instance, the three-sigma rule, which retains approximately 99.7\% of data under the assumption of Gaussian noise. This principle can be also adapted to other noise distributions. The preference function $\phi$ is defined as:
            \begin{equation*}
                \phi(\delta_{\vect{x}, i}) = e^{-\frac{\delta^2_{\vect{x}, i}}{2 \sigma^2}}.
            \end{equation*}
            We considered also a \emph{binary} preference function, that is $\phi(\delta_{\vect{x}, i}) = 1$ when $|\delta_{\vect{x}, i}| \leq \epsilon$ and $0$ otherwise, giving rise to a discrete preference space $\{0, 1\}^m$. Unless otherwise specified, we refer to $\mathcal{P} = [0, 1]^m$ as the \emph{continuous} \pspace.
            As illustrated in~\cref{fig:input_output}, a family $\mathcal{F}$ describing genuine data can isolate anomalies in $\mathcal{P}$, making them detectable via \pisolation. 
            
            \subsubsection{Choice of model family \texorpdfstring{$\mathcal{F}$}{F}}
                \label{subsubsec:model_types}
                
                The model family $\mathcal{F}$ in~\eqref{eq:preference} encodes the prior knowledge on genuine data $G$ in the \pspace, which can be either \emph{global} or \emph{local}.
                In~\cref{subfig:models_global_family}, where $G$ refers to circles, choosing $\mathcal{F}$ as the family of circles provides a \emph{global} description of genuine data.
                In~\cref{subfig:models_manifold_local_family} we show that, even if the family of $G$ is unknown, it is possible to adopt lines to describe $G$ locally. This results in \emph{local} models $\vect{\theta}_i$ that describe accurately the manifold only in a neighborhood of $g \in G$, but in~\cref{subsec:sliding_pif} we will show that this is enough to detect anomalous data.
            
            \paragraph{\textbf{Remark}}
            In principle, if the genuine data were generated by a single underlying model, one could first estimate this model robustly, and then detect anomalies as points with the highest deviation from the model.
            However, in our scenario, the data might arise from a mixture of unknown multiple structures, which may be  valid only locally. Even if all genuine models were available, it would not be trivial to determine which residual $\delta_{x,i}$ (with respect to model $\theta_i$) should be considered for a given point $x$.
            Rather than solving the difficult and ill-posed problem of multi-model fitting, and instead of relying on residuals directly, we embed points into the preference space, which captures how consistently each point agrees with the model set randomly sampled.
            This sidesteps model estimation entirely, and enables robust anomaly detection through isolation in preference space, where outliers naturally emerge as structurally inconsistent points.
        
        \subsection{\pisolation}
            \label{subsec:preference_isolation}

            \begin{figure}[t]
                \centering
                \hspace{-0.525cm}
                \begin{subfigure}[t]{.2\linewidth}
                    \centering
                    \includegraphics[width=\linewidth, angle=90]{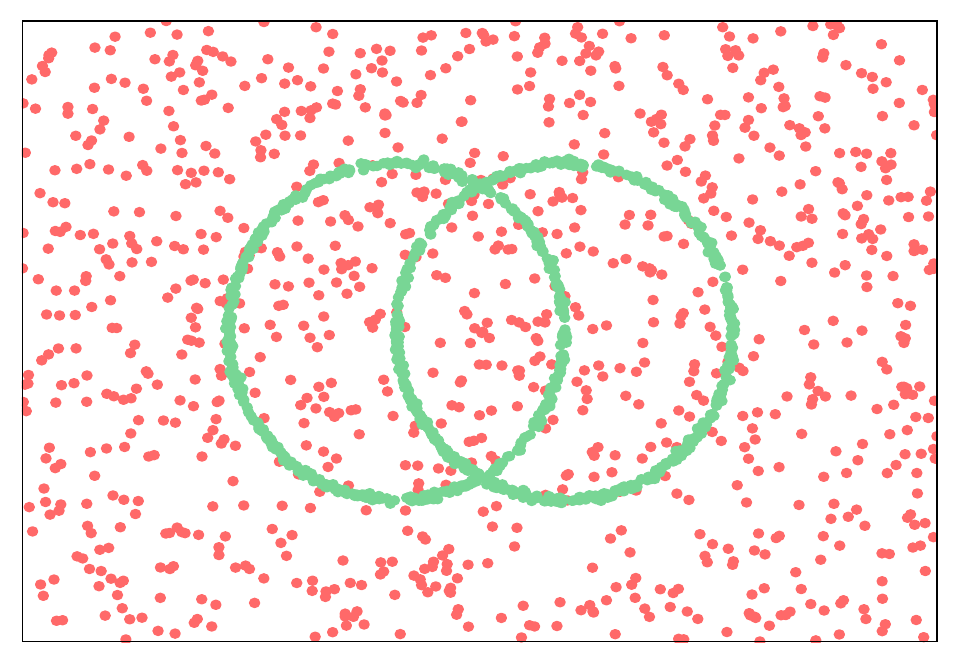}
                    \caption{Input data.}
                    \label{fig:input}
                \end{subfigure}
                \hfill
                \hspace{-1cm}
                \begin{subfigure}[t]{.2\linewidth}
                    \centering
                    \includegraphics[width=\linewidth, angle=90]{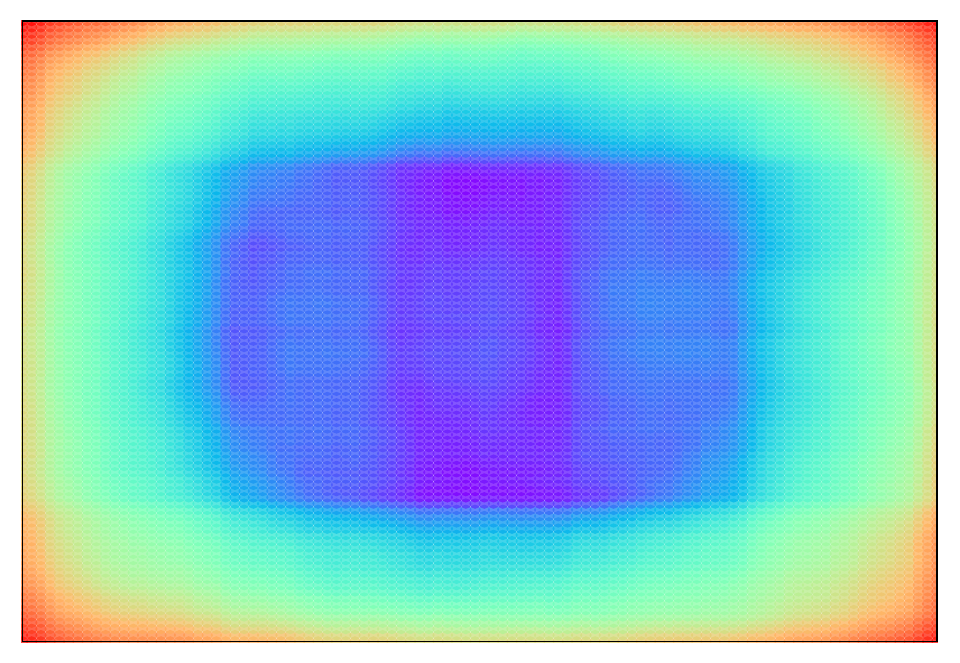}
                    \caption{\ifor~\cite{LiuTingAl12}}
                    \label{fig:ifor}
                \end{subfigure}
                \hfill
                \hspace{-1cm}
                \begin{subfigure}[t]{.2\linewidth}
                    \centering
                    \includegraphics[width=\linewidth, angle=90]{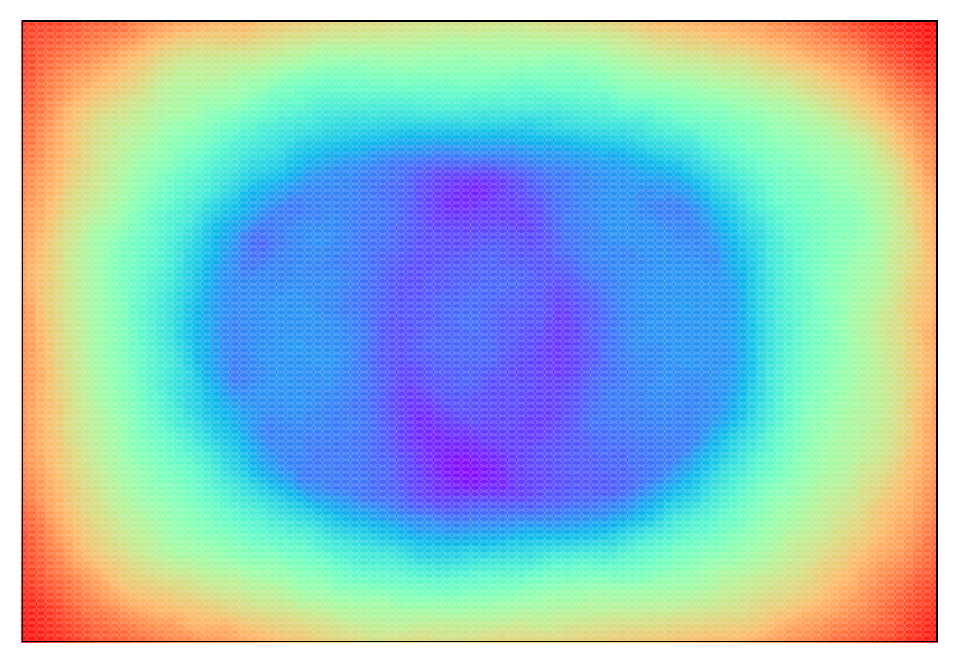}
                    \caption{\eifor~\cite{HaririKindAl21}}
                    \label{fig:eifor}
                \end{subfigure}
                \hfill
                \hspace{-1cm}
                \begin{subfigure}[t]{.2\linewidth}
                    \centering
                    \includegraphics[width=\linewidth, angle=90]{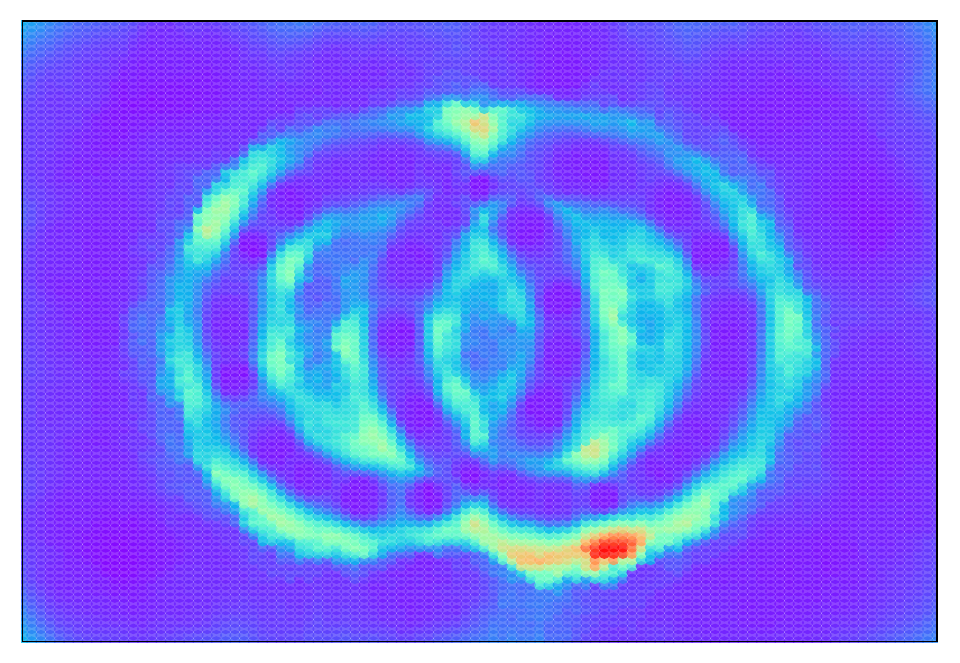}
                    \caption{\lof~\cite{BreunigKriegel00}}
                    \label{fig:lof}
                \end{subfigure}
                \hfill
                \hspace{-1cm}
                \begin{subfigure}[t]{.2\linewidth}
                    \centering
                    \includegraphics[width=\linewidth, angle=90]{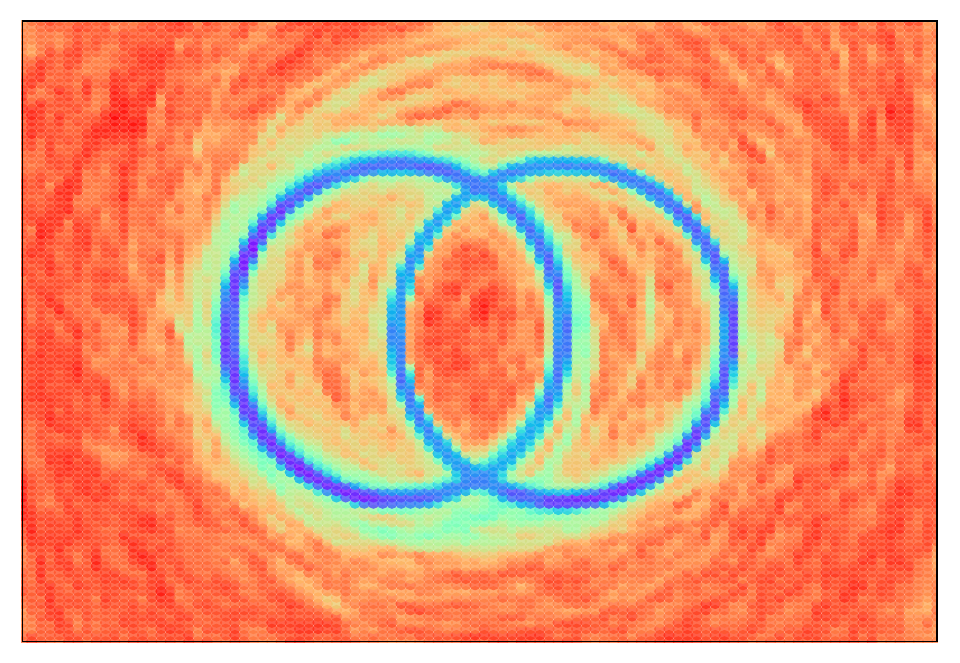}
                    \caption{\pif}
                    \label{fig:pif}
                \end{subfigure}
                \caption{Anomaly scores $\alpha(\cdot)$, color coded, computed by different algorithms. \pif effectively detects anomalies (red points in (a)) when a suitable  family $\mathcal{F}$ (circles) is chosen for \pembedding.}
                \label{fig:anomaly_scores}
            \end{figure}
            
            \pisolation consists in a scheme to isolate anomalies in the \pspace $\mathcal{P}$, resulting in better detection with respect to methods that operate in the ambient space $\mathcal{X}$. This is illustrated in~\cref{fig:anomaly_scores}, where traditional isolation-based methods operating in $\mathcal{X}$ struggle to detect anomalies inside the circle, while \pif succeeds in this challenging task.
            
            \subsubsection{Distances in \pspace}
                \label{subsubsec:distance_measures}
                
                The effectiveness of \pif is closely related to the ad-hoc distance functions adopted to isolate anomalies in the \pspace. Several works in the literature~\cite{ToldoFusiello08,MagriFusiello14,LeveniMagriAl21,LeveniMagriAl23} investigates distances in the \pspace:

                \paragraph{Jaccard}
                The Jaccard~\cite{Jaccard01} distance is an indicator of dissimilarity between two sets. Given $\vect{p}, \vect{q}$ belonging to the binary \pspace $\{0, 1\}^m$~\cite{ToldoFusiello08}, their Jaccard distance is defined as:
                \begin{equation}
                    \label{eq:jaccard_vect}
                    d_J(\vect{p}, \vect{q}) = 1 - \frac{\sum_{i=1}^{m} (p_i \wedge q_i)}{\sum_{i=1}^{m} (p_i \vee q_i)},
                \end{equation}
                where $\wedge$ and $\vee$ denote the logical $intersection$ and $union$, respectively.

                \paragraph{Ruzicka}
                The Ruzicka~\cite{Ruzicka58} distance extends Jaccard distance to vectors with values in $[0, 1]^m$. Analogously to~\eqref{eq:jaccard_vect}, the Ruzicka distance is defined as:
                \begin{equation*}
                    d_R(\vect{p}, \vect{q}) = 1 - \frac{\sum_{i=1}^{m} \min\{p_i, q_i\}}{\sum_{i=1}^{m} \max\{p_i, q_i\}}.
                \end{equation*}

                \paragraph{Tanimoto}
                The Tanimoto~\cite{Tanimoto57} distance extends Jaccard as follows:
                \begin{equation*}
                    d_T(\vect{p},\vect{q}) = 1 - \frac{\langle\vect{p},\vect{q}\rangle}{\|\vect{p}\|^{2}+\|\vect{q}\|^{2}- \langle\vect{p},\vect{q}\rangle} = 1 - \frac{\sum_{i=1}^{m} p_i q_i}{\sum_{i=1}^{m} (p_i^{2} + q_i^{2}) - \sum_{i=1}^{m} p_i q_i}.
                \end{equation*}

                It is worth noting that these three distances reach their maximum when $\vect{p}, \vect{q}$ are orthogonal ($\vect{p} \perp \vect{q}$). Since anomalies become sparse vectors in $\mathcal{P}$, as they are typically poorly described by fitted models, their preference vectors are nearly orthogonal to those of genuine data, becoming isolated points in $\mathcal{P}$.
                \cref{subfig:embedding} shows the MDS of the points in the \pspace with reference to Tanimoto distance. Genuine data $G$ from line $\vect{\theta}_1$ are close in the \pspace, but distant to those from line $\vect{\theta}_2$, while anomalies $A$ are both distant to each other and to genuine data $G$, resulting \emph{isolated}.

            \subsubsection{\vitree}
                \label{subsubsec:voronoi_itree}
                
                \begin{figure}[t]
                    \centering
                    \includegraphics[width=\linewidth]{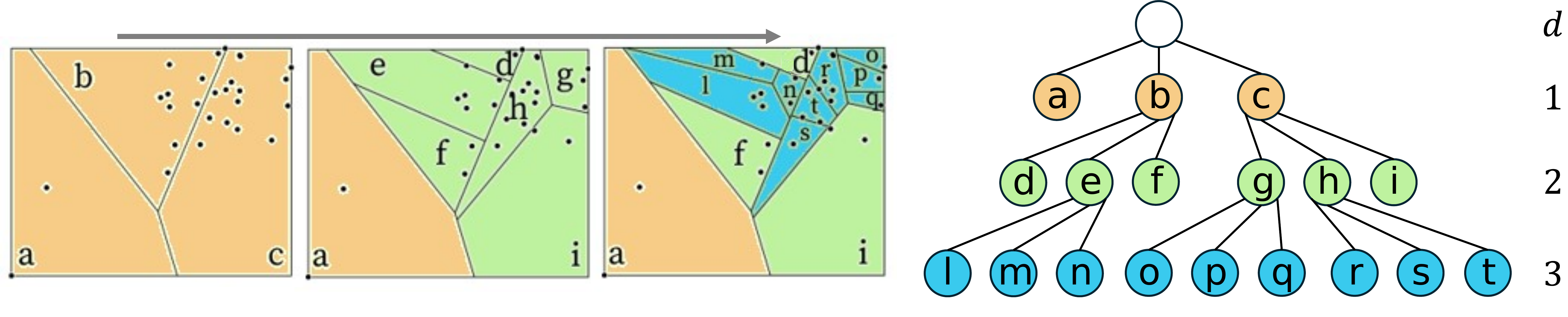}
                    \caption{\vitree with branching factor $b=3$ and depth limit $l=3$ built on points in $\mathbb{R}^2$. Regions are recursively split in $b$ sub-regions. The most isolated samples fall in leaves at lowest depth, such as \texttt{a} and \texttt{i}.}
                    \label{fig:voronoi_itree}
                \end{figure}

                \begin{algorithm}[t]
                    \footnotesize
                    \setstretch{1.35}
                    \caption{\vitree \label{alg:vitree}}
                    \KwIn{$P$ - preferences, $e$ - tree depth, $l$ - depth limit, $b$ - branching factor}
                    \KwOut{$T$ - a \vitree}
                        \eIf{$e \geq l$ \normalfont{or} $|P| < b$\label{line:stopping_criteria}}
                            {\Return $exNode\{Size \leftarrow |P|\}$\label{line:node_creation}}
                            {$\{\vect{s}_i\}_{i=1,\ldots,b} \leftarrow \text{\textsc{Subsample}}(P, b)$\label{line:subsample_seeds}\\
                             $\{P_i\}_{i=1,\ldots, b} \leftarrow \text{\textsc{voronoiPartition}}(P, \{\vect{s}_i\}_{i=1,\ldots,b})$\label{line:vsplit}\\
                             $chNodes \leftarrow \emptyset$ \label{line:begin_recursive}\\
                             \For{$i = 1$  \normalfont{to} $b$}
                                 {$chNodes \leftarrow chNodes \cup \text{\vitree}(P_i, e + 1, l, b)$} \label{line:end_recursive}
                            \Return $inNode\{ChildNodes \leftarrow chNodes, SplitPoints \leftarrow \{\vect{s}_i\}_{i=1,\ldots,b}\}$}
                \end{algorithm}

                \begin{algorithm}[t]
                    \footnotesize
                    \setstretch{1.35}
                    \caption{\viforest \label{alg:viforest}}
                    \KwIn{$P$ - preference representations, $t$ - number of trees, $\psi$ - sub-sampling size, $b$ - branching factor}
                    \KwOut{$F = \{T_k\}_{k = 1, \dots, t}$ - set of \vitrees}
                        $F \leftarrow \emptyset$ \\
                        $l \leftarrow ceiling(\log_{b}\psi)$ \\
                        \For{$k = 1$ \normalfont{to} $t$}
                            {$P_\psi \leftarrow \text{\textsc{Subsample}}(P, \psi)$ \label{line:subsample} \\
                             $T_k \leftarrow \text{\vitree}(P_\psi, 0, l, b)$ \label{line:tree_construction} \\
                             $F \leftarrow F \cup T_k$ \label{line:end_forest_construction}}
                        \Return $F$
                \end{algorithm}
                
                \vitree is the  starting point of our framework for \pisolation. It is designed to be the most \emph{general} algorithm, as it can seamlessly adapt to any distance measure defined in the preference space -- such as those discussed in the previous section. In fact, \vitree relies on Voronoi tessellations, and can be extended to operate directly in the ambient space when appropriate. As such, it serves as a consistent reference for  evaluating detection accuracy varying the distances adopted, and illustrates the benefits of  \pisolation within \pspace.
                
                We employ the previously defined distances within a tree-based recursive partitioning algorithm, where points are grouped in each level based on their similarity according to the selected distance, to identify anomalies as the most isolated points in the \pspace $\mathcal{P}$. Our approach, \vitree, consists in a nested configuration of Voronoi tessellations, where each cell is recursively split into $b$ sub-regions, as illustrated in~\cref{fig:voronoi_itree} where $b = 3$.
                The construction of a \vitree, described in~\cref{alg:vitree}, starts by randomly sampling $b$ Voronoi seeds $\{\vect{s}_i\}_{i=1,\ldots,b} \subset P$ (\cref{line:subsample_seeds}) to partition the set of preferences $P$ (\cref{line:vsplit}). Specifically, $P$ is split into $b$ subsets $\{P_i\}_{i=1,\ldots, b}$, where each $P_i \subset P$ collects all the points having $\vect{s}_i$ as the closest seed according to the selected distance. Each $P_i$ is then used to build sub-trees recursively (\cref{line:begin_recursive,line:end_recursive}). The recursive partitioning process stops when the number of points in the region $P_i$ is less than $b$, or the tree reaches a maximum depth (\cref{line:stopping_criteria,line:node_creation}) set to $l = \log_{b}\psi$ as in~\cite{Knuth98}, where $\psi$ is the number of points used to build the tree.
                To decrease the randomness in \vitree construction, we build an ensemble of $t$ \vitrees, termed \viforest (\cref{alg:viforest}), where each \vitree is built on a random subset $P_{\psi} \subset P$.

            \subsubsection{Anomaly score computation}
                \label{subsubsec:anomaly_score}


                We compute the anomaly score of a point $\vect{x} \in \mathcal{X}$ as a function of its \emph{average} depth within the \viforest. The rationale is that \emph{isolated} points fall in leaves at lower depth as in~\cite{LiuTingAl12}. We then collect the depths reached by each preference $\vect{p}_{\vect{x}} = \mathcal{E}(\vect{x}) \in P$, in each of the $t$ trees of the forest, into a set $D_{\vect{x}}$ (\cref{alg:main}, \cref{line:set,line:height}).
                Then (\cref{line:anomaly_score}), we compute the anomaly score as:
                \begin{equation}
                    \label{eq:anomaly_score}
                    \alpha_\psi(\vect{x}) = 2^{-\frac{E(D_{\vect{x}})}{c(\psi)}},
                \end{equation}
                where $E(D_{\vect{x}})$ is the mean value over the elements of $D_{\vect{x}}$, and $c(\psi)$ is an adjustment factor.
                The depth of each $\vect{p}_{\vect{x}}$ is computed via the \textsc{PathLength} function, following a recursive tree traversal to the leaf containing $\vect{p}_{\vect{x}}$, similar to~\cite{LiuTingAl12}. As in~\cite{LiuTingAl12}, the depth is computed as $d(\vect{p}_{\vect{x}}) = e + c(exNode.Size)$, where $e$ represents the leaf node depth, and $c(\cdot)$ is an adjustment factor to account for samples that fell in the leaf during the tree construction process, but did not contribute to the depth due to the depth limit $l$ (\cref{alg:vitree}, \cref{line:stopping_criteria,line:node_creation}).

        \subsection{Dealing with efficiency: hashing to speed up distance computation}
            \label{subsec:efficiency}

            \begin{figure}
                \centering
                \begin{subfigure}[t]{.5\linewidth}
                    \centering
                    \includegraphics[width=0.7\linewidth]{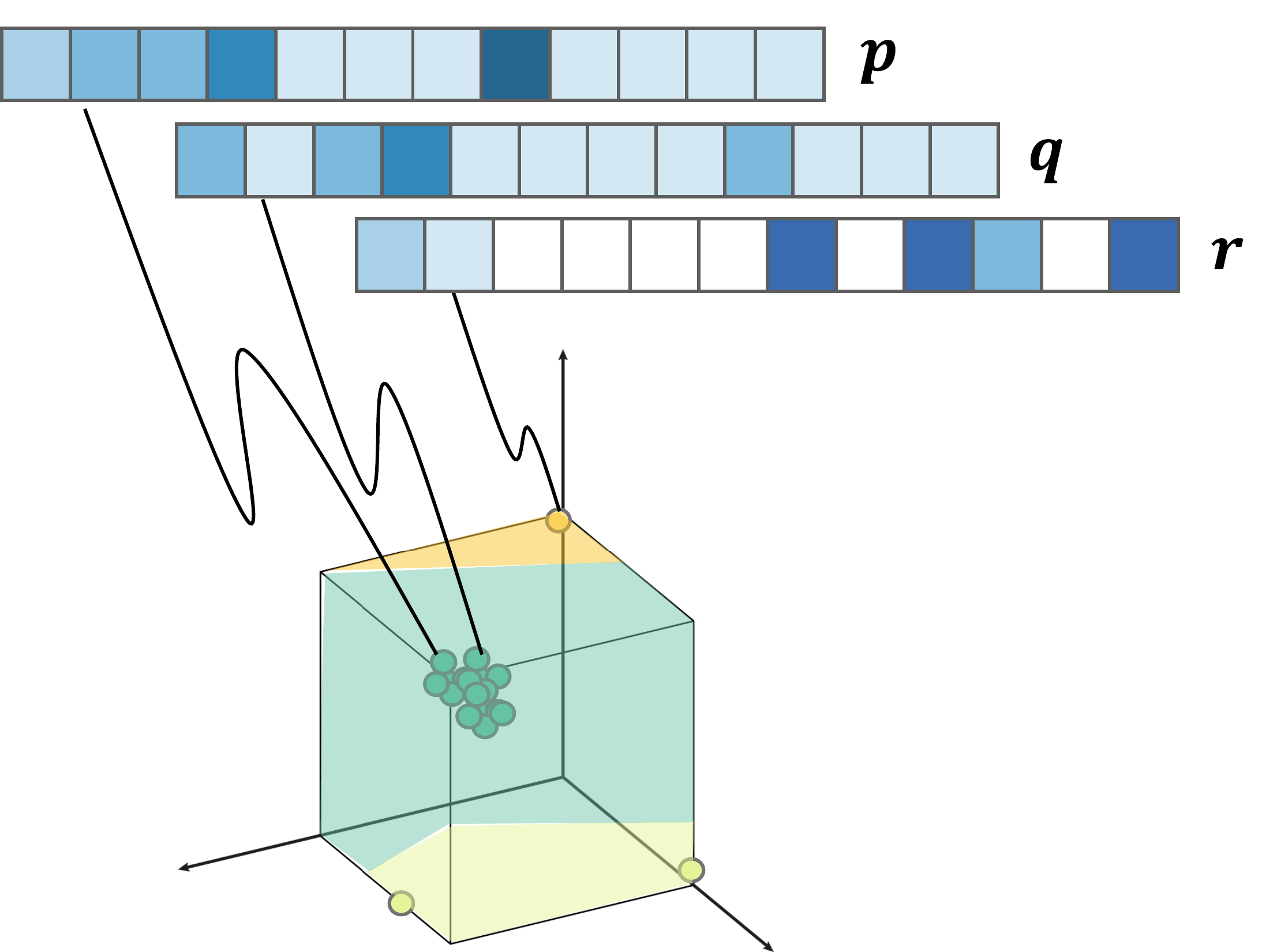}
                    \caption{}
                    \label{subfig:rzhash_partition}
                \end{subfigure}
                \hfill
                \hspace{-8cm}
                \begin{subfigure}[t]{.5\linewidth}
                    \centering
                    \includegraphics[width=\linewidth]{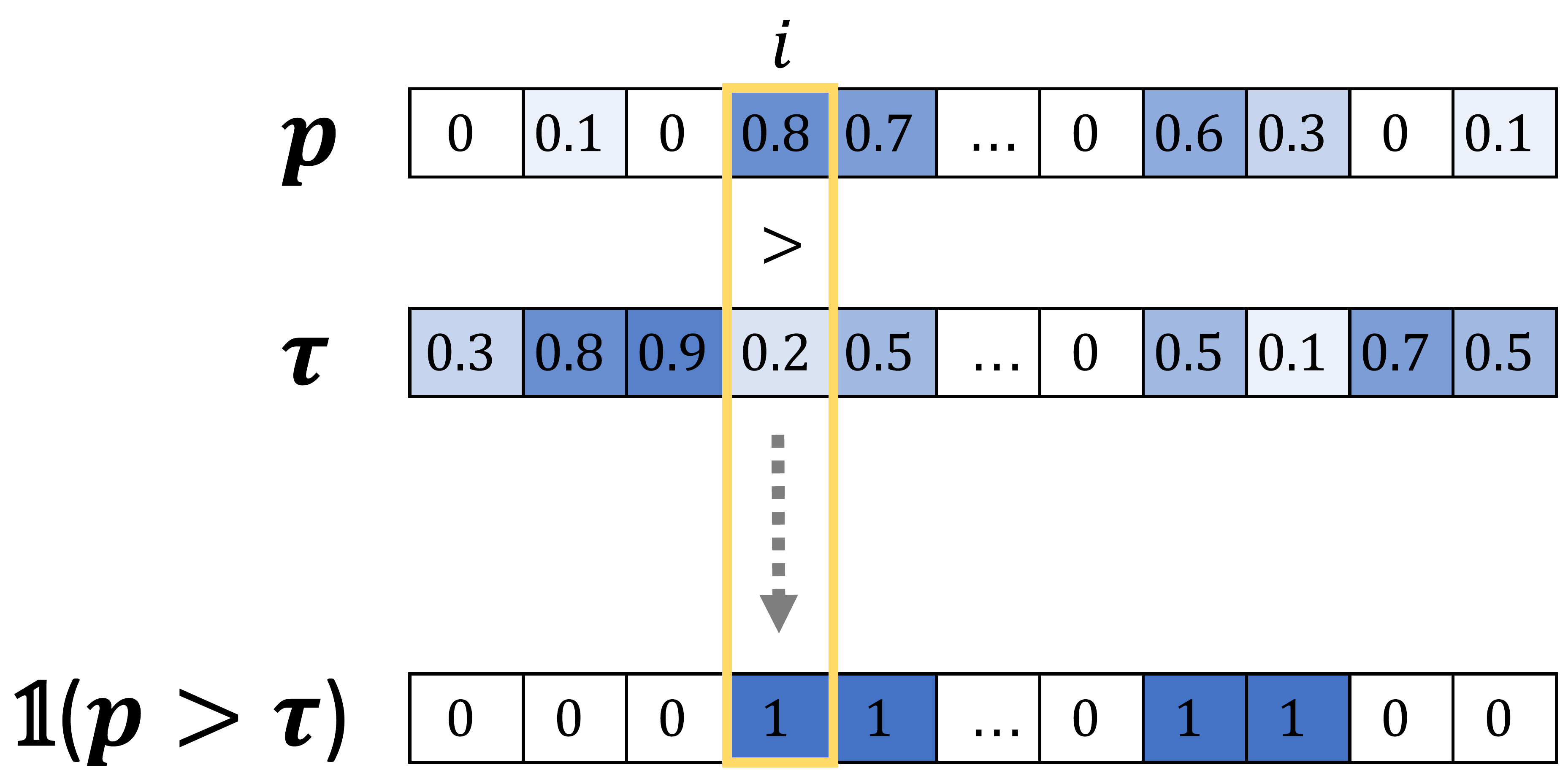}
                    \caption{}
                    \label{subfig:thresholding}
                \end{subfigure}
                \caption{(a) \rzhash induces a partitioning of the preference space $\mathcal{P}$ (color-coded here), where close points with respect to the Ruzicka distance ($\vect{p}, \vect{q}$) are more likely mapped to the same region, unlike isolate points ($\vect{r}$). We use \rzhash as the splitting criterion within each node of \rzhitree. (b) An example of binarization function used in~\cref{en:map}.}
                \label{fig:rzhash}
            \end{figure}

            Voronoi tessellations are effective in isolating preference vectors corresponding to anomalies, but they are inefficient when the number of points $n$ is large, as they require to compute distances between each of the $n$ points $\vect{p}_{\vect{x}} \in P$ and seeds $\{\vect{s}_i\}_{i=1,\ldots,b}$, leading to a computational complexity of $O(n \, b)$ for each Voronoi tessellation. To overcome this bottleneck, we revisit the way in which the distances between preferences are computed and introduce \rzhash, a novel Locality Sensitive Hashing (LSH) scheme that splits points in the \pspace with complexity $O(n)$, and ensures that nearby points are more likely mapped to the same bin without computing point-to-point distances. At high level, the bins obtained via LSH, play the same role of the Voronoi cells used in the \vitree.

            \subsubsection{\rzhash}
                \label{subsubsec:rzhash}
                
                \rzhash is a mapping $h^{\vect{\pi}, \vect{\tau}}_{ruz}: [0, 1]^m \rightarrow \{1, \dots, m\}$ from the \pspace to a collection of $m$ bins. The main idea, depicted in~\cref{subfig:rzhash_partition}, is that when two points $\vect{p}, \vect{q} \in \mathcal{P}$ are close with respect to Ruzicka distance, they are likely mapped to the same bin with high probability, \emph{i.e.}, $d_R(\vect{p}, \vect{q}) \approx 0 \iff Pr[h^{\vect{\pi}, \vect{\tau}}_{ruz}(\vect{p}) = h^{\vect{\pi}, \vect{\tau}}_{ruz}(\vect{q})] \approx 1$, while a point $\vect{r}$ far from $\vect{p}$ is likely mapped to a different bin, \emph{i.e.}, $d_R(\vect{p}, \vect{r}) \approx 1 \iff Pr[h^{\vect{\pi}, \vect{\tau}}_{ruz}(\vect{p}) = h^{\vect{\pi}, \vect{\tau}}_{ruz}(\vect{r})] \approx 0$.
                In practice, given a point $\vect{p} \in [0, 1]^m$, the \rzhash mapping is computed as follows:
                \begin{enumerate}
                    \item sample a vector of thresholds $\vect{\tau} = [\tau_1, \dots, \tau_m]$, where $\tau_i \sim \mathcal{U}_{[0, 1)}$,\label{en:thresholds}
                    \item sample a random permutation $\vect{\pi} = [\pi_1, \dots, \pi_m]$ of the indices $\{1, \dots, m\}$,\label{en:permutation}
                    \item map $\vect{p}$ to the bin having index $h^{\vect{\pi}, \vect{\tau}}_{ruz}(\vect{p}) = \min(\vect{\pi} \cdot \mathds{1}(\vect{p} > \vect{\tau}))$, where $\mathds{1}(\vect{p} > \vect{\tau}) = [\mathds{1}(p_1 > \tau_1), \dots, \mathds{1}(p_m > \tau_m)] \in \{0, 1\}^m$ is a binarization function (\cref{subfig:thresholding}), and the minimum is computed over the indices.\label{en:map}
                \end{enumerate}
                The binarization function uses the thresholds vector $\vect{\tau}$ to map points from the continuous \pspace $[0, 1]^m$ into the binary \pspace $\{0, 1\}^m$, while the permutation $\vect{\pi}$ and $\min(\cdot)$ operator are used as in \mhash~\cite{BroderCharikarAl00}, a Locality Sensitive Hashing (LSH) scheme for the Jaccard distance on binary vectors. We prove that the \rzhash mapping is an LSH of the Ruziska distance, ensuring that similar points are more likely to be mapped to the same bin, as stated in the following
                \begin{theorem}
                    \label{thm:ruzhash_short}
                    Given $\vect{p}, \vect{q} \in [0, 1]^m$, then:
                    \begin{equation*}
                        Pr[h^{\vect{\pi}, \vect{\tau}}_{ruz}(\vect{p}) = h^{\vect{\pi}, \vect{\tau}}_{ruz}(\vect{q})] = 1 - d_R(\vect{p}, \vect{q}).
                    \end{equation*}
                \end{theorem}
                The proof is in~\ref{apx:rzhash}, where~\cref{fig:correlation} also shows the empirical correlation between Ruzicka distance and its approximation via \rzhash.
                
            \subsubsection{\rzhitree}
                \label{subsubsec:rzhitree}
                
                \begin{figure}[t]
                    \centering
                    \includegraphics[width=0.6\linewidth]{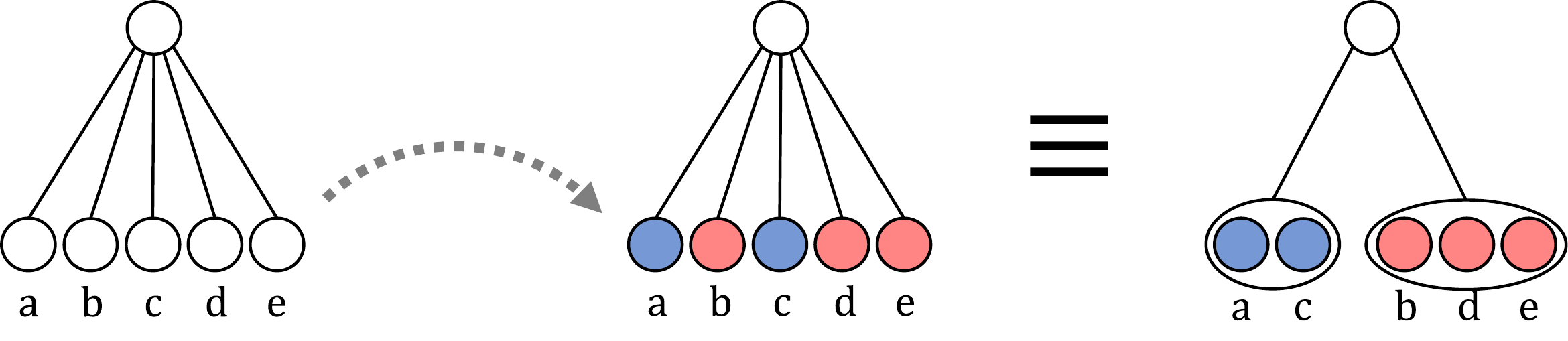}
                    \caption{Left: split performed by \rzhash when $m = 5$. Middle: nodes aggregation where the groups are color coded. Right: resulting tree with branching factor $b = 2$.}
                    \label{fig:partitioning}
                \end{figure}

                \begin{algorithm}[tb]
                    \footnotesize
                    \setstretch{1.35}
                    \caption{\rzhitree \label{alg:rzhitree}}
                    \KwIn{$P$ - preferences, $e$ - tree depth, $l$ - depth limit, $b$ - branching factor}
                    \KwOut{$T$ - a \rzhitree}
                        \eIf{$e \geq l$  \normalfont{or} $|P| < b$}
                            {\Return $exNode\{Size \leftarrow |P|\}$}
                            {$\vect{\tau}, \vect{\pi} \leftarrow \text{\textsc{RuzHashSample}}(m)$\label{line:rzhsample}\\
                             $\vect{\beta} \leftarrow \text{\textsc{AggregationSample}}(m, b)$\label{line:aggrsample}\\
                             $\{P_i\}_{i=1,\ldots, b} \leftarrow \text{\textsc{RuzHashPartition}}(P, \vect{\tau}, \vect{\pi}, \vect{\beta})$\label{line:rzhsplit}\\
                             $chNodes \leftarrow \emptyset$\\
                             \For{$i = 1$  \normalfont{to} $b$}
                                 {$chNodes \leftarrow chNodes \cup \text{\rzhitree}(P_i, e + 1, l, b)$}
                            \Return $inNode\{ChildNodes \leftarrow chNodes, Thresholds \leftarrow \vect{\tau}$\\
                            \hspace{2.5cm} $Permutation \leftarrow \vect{\pi}, Aggregation \leftarrow \vect{\beta}\}$}
                \end{algorithm}

                \begin{table}[t]
                    \footnotesize
                    \centering
                    \caption{Differences between \iforest, \viforest and \rzhiforest.}
                    \resizebox{0.8\textwidth}{!}{
                    \begin{tabular}{c||c|c|c|c}
                                    & \multirow{2}{*}{Splitting scheme} & \multirow{2}{*}{Distance} & \multicolumn{2}{c}{Computational complexity}                                               \\ \cline{4-5}
                                    &                                   &                           & Training                                      & Testing                                    \\ \hline
                        \viforest   & Voronoi                           & Tanimoto                  & $O(\psi \: t \: b \: \log_{b} \psi)$ & $O(n \: t \: b \: \log_{b} \psi)$ \\
                        \iforest    & LSH                               & $\ell_1$                  & $O(\psi \: t \: \log_{2} \psi)$         & $O(n \: t \: \log_{2}\psi)$          \\
                        \rzhiforest & LSH                               & Ruzicka                   & $O(\psi \: t \: \log_{b} \psi)$         & $O(n \: t \: \log_{b} \psi)$         \\
                    \end{tabular}
                    }
                    \label{tab:differences}
                \end{table}

                At each step of the recursive tree building process, \rzhash splits the set of preferences $P$ into exactly $m$ bins, where $m$ is the dimension of \pspace $\mathcal{P} = [0, 1]^m$, as illustrated in~\cref{fig:partitioning} (left). Since $m$ is often larger than the number of points $n$, most of the bins remain empty, causing unnecessary memory usage. Additionally, a high branching factor $m$ results in few tree levels, leading to less fine grained anomaly scores. To address this, we reduce the branching factor $b < m$ by randomly aggregating the $m$ bins produced by \rzhash, as shown in~\cref{fig:partitioning} (right).
                We perform this aggregation in two steps: \emph{i}) we sample a vector $\vect{\beta} = [\beta_1, \dots, \beta_m]$, where $\beta_i \sim \mathcal{U}_{\{1, \dots, b\}}$, to assign a random value from $\{1, \dots, b\}$ to each bin produced by \rzhash (represented by different colors in~\cref{fig:partitioning}, middle), and \emph{ii}) we define a new \rzhash mapping $h^{\vect{\pi}, \vect{\tau}, \vect{\beta}}_{ruz} : [0, 1]^m \rightarrow \{1, \dots, b\}$ such that $h^{\vect{\pi}, \vect{\tau}, \vect{\beta}}_{ruz}(\vect{p}) = \beta_{h^{\vect{\pi}, \vect{\tau}}_{ruz}}$. This aggregation remaps each bin $\{1, \dots, m\}$ produced by \rzhash to a new bin $\{1, \dots, b\}$, reducing the branching factor from $m$ to $b$.
                
                \cref{alg:rzhitree} details the construction of \rzhitree, where in~\cref{line:rzhsample,line:aggrsample} we sample the random vectors, and in~\cref{line:rzhsplit} we split $P$ via \rzhash.
                In~\ref{apx:variable_split}, we theoretically study the impact of the branching factor $b$ on \rzhash, showing that it influences the correlation slope between \rzhash and Ruzicka. \cref{tab:differences} summarizes key differences between \iforest, \viforest and \rzhiforest.
                
        \subsection{Dealing with locality: a sliding window approach}
            \label{subsec:sliding_pif}
            
            Here we present \spif, an efficient variation of \pif that exploits a \emph{locality principle}: points close in the ambient space $\mathcal{X}$ are likely to belong to the same genuine structure. This turns to be useful when the family $\mathcal{F}$ approximates the genuine data only locally (see~\cref{subfig:models_manifold_local_family}), and anomalies deviate from the regularity of smooth genuine structures.
            
            \begin{figure}[t]
                \centering
                \begin{subfigure}[t]{.15\linewidth}
                    \centering
                    \hspace*{-1cm}
                    \includegraphics[height=0.85\linewidth]{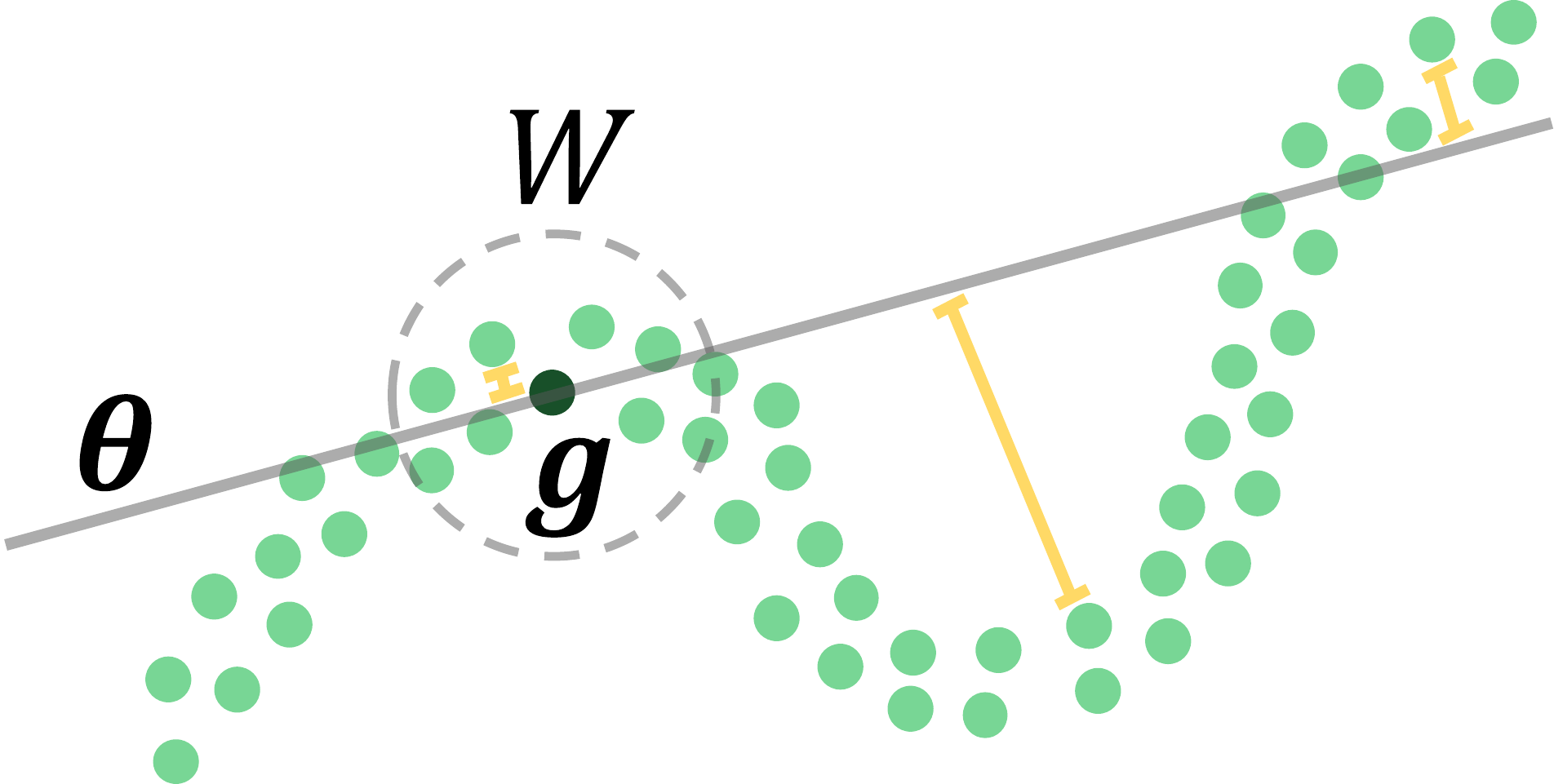}
                    \caption{}
                    \label{subfig:locality}
                \end{subfigure}
                \hspace{3cm}
                \begin{subfigure}[t]{.2\linewidth}
                    \centering
                    \hspace*{-0.5cm}
                    \includegraphics[height=\linewidth]{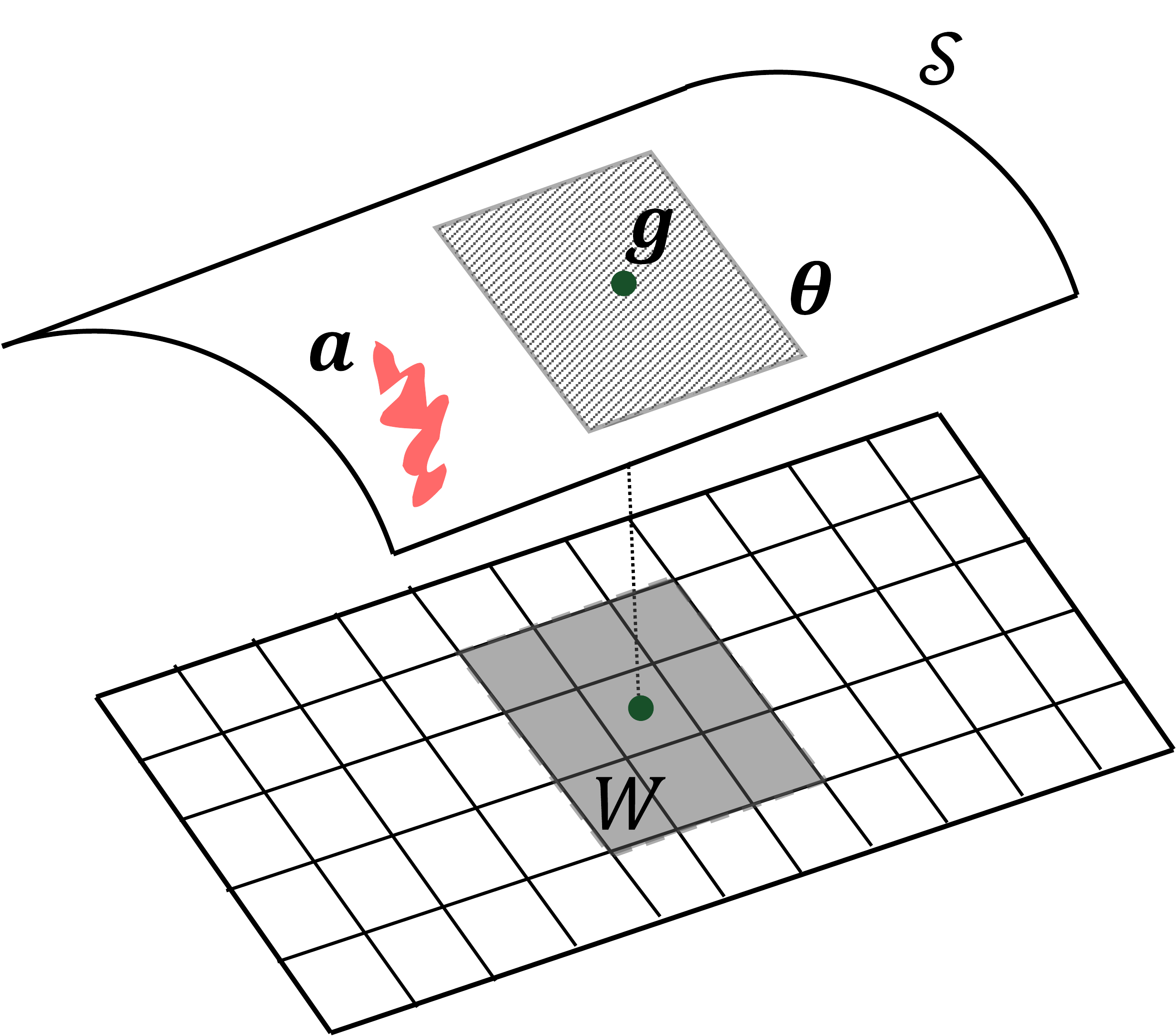}
                    \caption{}
                    \label{subfig:depthmap}
                \end{subfigure}
                \caption{(a) Preferences are meaningful within the approximated region. (b) Range images connectivity enables efficient neighborhood definition.}
                \label{fig:locality}
            \end{figure}
            
            \subsubsection{\spif}
              
                In many real-world problems, particularly in geometric settings—a global structure is either absent or too complex to capture with a single model, but local geometric regularities remain clearly identifiable. A representative example is that of 3D surface scans, where small neighborhoods can be approximated by simple geometric primitives (e.g., planes), even though the full surface may exhibit significant curvatures.
                In these scenarios, adopting a global preference embedding may be impractical or inefficient, as it would require an excessive number of model instances to capture the diverse local phenomena, resulting in increased computational and memory demands. \spif addresses this challenge by applying \pif locally, within spatially overlapping regions, allowing adaptation to spatial variations in structure. This approach preserves the strengths of the \pif framework while improving both scalability and accuracy, and illustrates its modular and flexible nature when applied to complex geometric data.
                
                Specifically, \spif leverages the \emph{locality} principle by fitting models on points sampled from windows $W$ and performing structure-based AD independently within each window. To present \spif, we consider range images, which can be represented as 3D surfaces as illustrated in~\cref{subfig:depthmap}, and where we can efficiently exploit the locality principle since data are structured in a grid format.
                We assume that genuine data $G$ lie on a smooth surface $\mathcal{S}$, thus a genuine point $\vect{g}$ can be locally described by its tangent plane $\vect{\theta}$, while anomalies $A$ (defects, such as cracks) cannot.
                Rather than considering the whole dataset, we operate in a window-wise fashion by first selecting overlapping windows, and then performing \pif on the corresponding set of points $X_W$, producing an anomaly score $\alpha^W_{\psi}(\vect{x})$~\eqref{eq:anomaly_score} for each point $\vect{x} \in X_W$ and for each window $W$ it belongs to. The window-wise approach has two advantages: \emph{i)} we draw minimal sample sets from nearby points, promoting the extraction of genuine local models as in localized sampling strategies~\cite{KanazawaKawakami04}, and \emph{ii)} we compute preferences only for points within the local region of interest, yielding significant computational gain (\cref{subfig:locality}).
                Finally, we compute the overall anomaly score for each point $\vect{x}$ by averaging scores $\alpha^W_{\psi}(\vect{x})$ across all windows.
                
                The choice of the window size $\omega$ is closely tied to the size of the anomalous region, and it must be large enough to ensure anomalies remain a minority within the window. We found that windows overlapping by half of their width strike a good balance between efficiency and effectiveness.
                
                \begin{figure}[t]
                    \centering
                    \begin{subfigure}[t]{.2\linewidth}
                        \centering
                        \hspace*{-0.75cm}
                        \includegraphics[height=\linewidth]{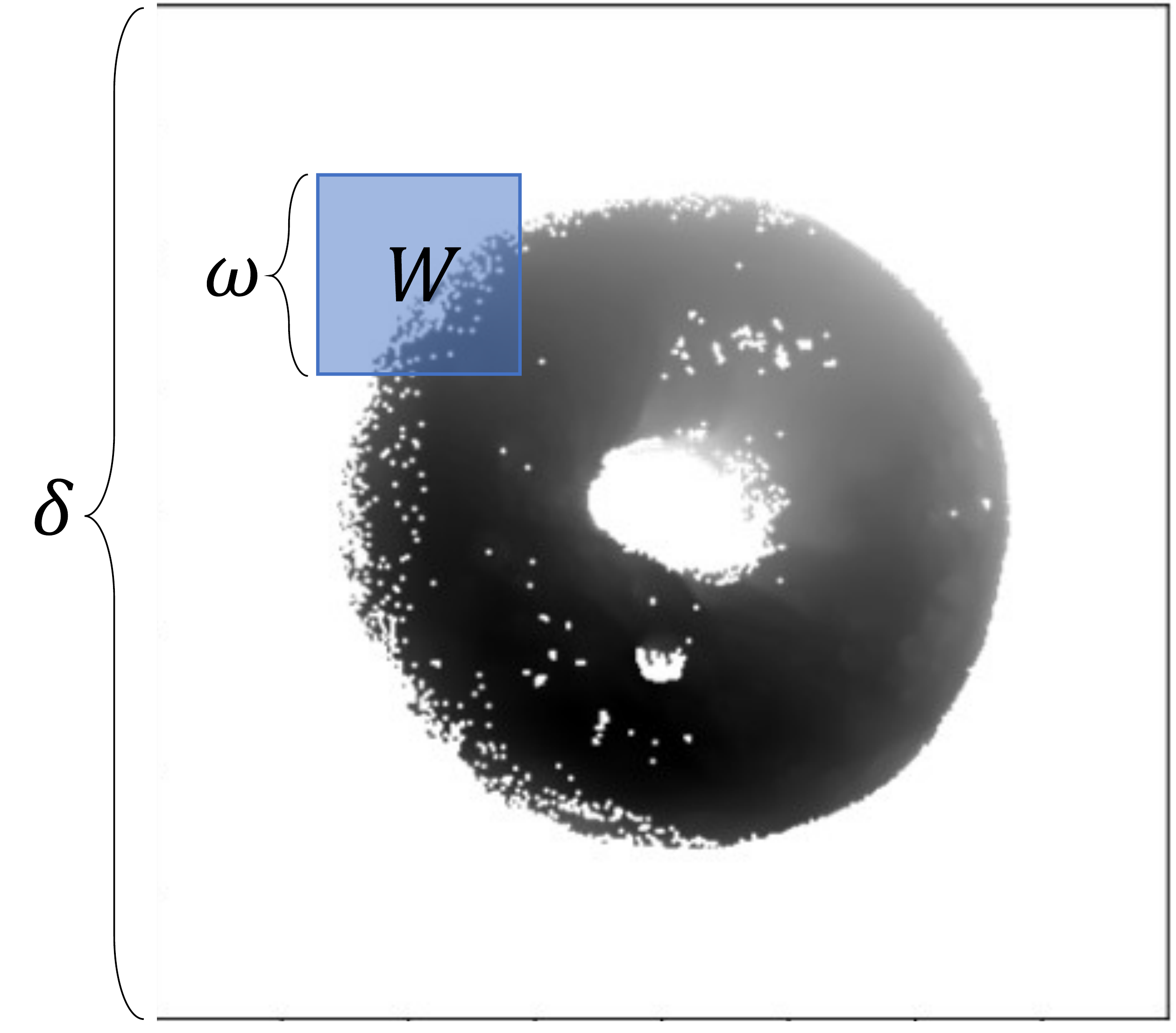}
                        \caption{}
                        \label{subfig:image_window}
                    \end{subfigure}
                    \hspace{2cm}
                    \begin{subfigure}[t]{.2\linewidth}
                        \centering
                        \hspace*{-0.9cm}
                        \includegraphics[height=\linewidth]{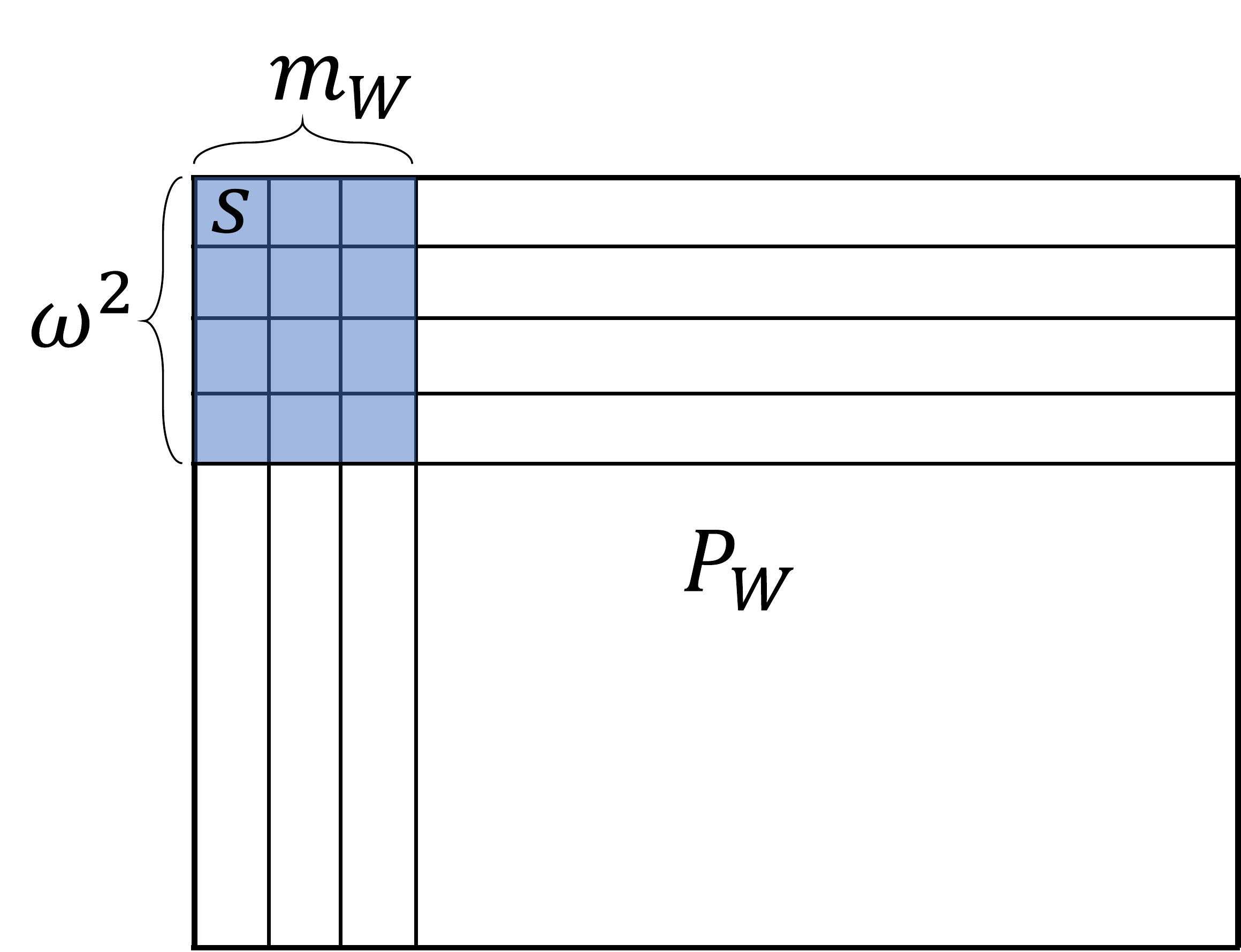}
                        \caption{}
                        \label{subfig:preference_matrix_window}
                    \end{subfigure}
                    \caption{(a) Bagel image with side length $\delta$ and superimposed window $W$ of size $\omega$, and (b) the corresponding local \pembedding matrix $P_W$.}
                    \label{fig:window}
                \end{figure}
                
                \spif benefits over \pif in terms of memory usage. \cref{subfig:image_window} shows an input image of size $\delta \times \delta$, a window $W$ of size $\omega = \frac{\delta}{k}$, and~\cref{subfig:preference_matrix_window} shows the \pembedding $P_W$ corresponding to $W$. The size of $P_W$ is proportional to the $\omega^2 = \frac{\delta^2}{k^2}$ points in $W$ and to the $m_W$ models fitted. Thus, the memory needed is $s \frac{\delta^2}{k^2} m_W$, where $s$ is the memory size of a preference value $p$.
                The total number of half-overlapping windows $W$ of size $\omega = \frac{\delta}{k}$ is $(2 \frac{\delta}{\omega} - 1)^2 = (2k - 1)^2$, resulting in a total memory need of
                \begin{equation*}
                    P_{size} = s \frac{d^2}{k^2} m_W (2k - 1)^2.
                \end{equation*}
                Thus, if we fix the total memory size to $P_{size}$, we can fit
                \begin{equation*}
                    m_W = \left\lfloor\frac{P_{size}}{s \frac{d^2}{k^2} (2k -1)^2}\right\rfloor
                \end{equation*}
                models.
                As instance, considering $s = 32$ bytes, $\delta = 800$ and $P_{size} = 1$ GB, \emph{each window} $W$ can accommodate $m_W = 419$ models for $\omega = \delta$ and $m_W = 110$ models for $\omega = \frac{\delta}{20}$, where \pif is \spif with window size $\omega = \delta$, resulting in a total number of models $m$ of $419$ and $167310$ respectively.

    \section{Experiments}
        \label{sec:experiments}
        
        In this section, we evaluate the benefits of \pif on synthetic and real datasets, showing its superior performance over state-of-the-art AD methods, and we demonstrate both the importance of leveraging our structure-based prior on genuine data and the need of employing the suitable distances in the \pspace (\cref{subsec:importance_pembedding}). We then evaluate the efficiency gain of \rzhiforest over \viforest (\cref{subsec:impact_hashing}) and assess the benefits of \spif on data satisfying the locality principle (\cref{subsec:impact_sliding}).
        
        \subsection{Datasets}
            \label{subsec:datasets}

            As regard synthetic datasets, we consider the 2D primitive fitting~\cite{LeveniMagriAl21,LeveniMagriAl23} comprising points $X \in \mathbb{R}^2$ where genuine data $G$ lie on lines or circles, and anomalies are uniformly sampled within the range of $G$ with $\frac{|A|}{|X|} = 0.5$.
            As far as real data are concerned, we consider AdelaideRMF dataset, comprising stereo images with annotated matching points, where anomalies are mismatches. The first $19$ images are static scenes with many planes, giving rise to genuine matches described by homographies. The last $19$ images are dynamic scenes with moving objects, genuine matches here are described by fundamental matrices.
            
            To evaluate the impact of local priors, we also considered the MVTec-3D AD~\cite{BergmannXinAl22} dataset, containing range images of $10$ different objects with various defects, such as holes and cracks. Scans were acquired using structured light, with positions $x$, $y$, and $z$ stored in three-channel tensors. We discard the color information and focus solely on the geometric data. We define and evaluate two distinct scenarios.
            i) \emph{Smooth objects scenario}: this includes objects that are locally smooth and thus conform well to the locality and smoothness assumptions underlying \spif. Specifically, we select \emph{cookie}, \emph{peach}, \emph{potato}, and \emph{carrot}. For these objects, the genuine structures are approximately smooth in local neighborhoods, and anomalies such as holes manifest as localized geometric irregularities. We also restrict our evaluation to the hole defect type, which directly violates the local smoothness assumption. ii) \emph{Mechanical objects scenario}: to assess the limits of our method, we also consider a set of mechanical  objects, namely \emph{cable gland}, \emph{dowel}, \emph{foam}, \emph{rope}, and \emph{tire}, which are characterized by sharp edges, corners, and discontinuities. In these cases, the assumption of local smoothness is frequently violated even in non-defective regions, making the detection problem more challenging and testing the robustness of our approach in settings that deviate from its ideal operating assumptions.
            
        \subsection{The effectiveness of structure-based anomaly detection}
            \label{subsec:importance_pembedding}

            \begin{table}[t]
                \footnotesize
                \centering
                \caption{Optimal $k$ parameters of \lof}
                \resizebox{.65\textwidth}{!}{
                \begin{tabular}{l@{\hskip 1.6cm}c@{\hskip 1.6cm}c@{\hskip 1.6cm}c}
                \toprule
                             & Euclidean & Preference binary & Preference \\
                \midrule
                circle       & $75$      & $25$              & $25$       \\
                line         & $25$      & $150$             & $75$       \\
                homography   & $-$       & $100$             & $100$      \\
                fundamental  & $-$       & $75$              & $80$       \\
                \bottomrule
                \end{tabular}
                }
                \label{tab:neighborhood}
            \end{table}
            
            Regarding synthetic datasets, we evaluated performance across multiple structured and controlled scenarios to assess the utility of the preference embedding.
            We compare \pif to \ifor~\cite{LiuTingAl12}, \eifor~\cite{HaririKindAl21} and \lof~\cite{BreunigKriegel00} either in the ambient space $\mathbb{R}^2$, the binary \pspace $\{0, 1\}^d$ and the continuous \pspace $[0, 1]^d$. We compute preferences with respect to $m = 10|X|$ model instances, and  we tested \pisolation respectively  with \emph{i}) \viforest Euclidean (\vifor $\ell_2$) in the ambient space $\mathbb{R}^2$, \emph{ii}) \viforest Jaccard (\vifor jac) in the binary \pspace $\{0, 1\}^d$, and \emph{iii}) \viforest Tanimoto (\vifor tani), \viforest Ruzicka (\vifor ruz) and \rzhiforest (\rzhifor) in the continuous \pspace $[0, 1]^d$.
            
            In this synthetic scenario, we also considered as a baseline the Euclidean distances computed directly in the ambient space, which is meaningful in this setting due to the low dimensionality of the data ($\mathbb{R}^2$) and its geometric interpretability. Moreover, inliers tend to be spatially denser and more structured, whereas outliers are often uniformly distributed over the plane.
            In contrast, for real-world datasets the situation is significantly different: the ambient space is higher-dimensional, as data represent point correspondences between two images, which naturally reside in the product of two projective planes and can be embedded in $\mathbb{R}^6$. In such a high-dimensional and inherently non-Euclidean space, standard Euclidean distances become much less informative for detecting anomalies. For this reason, we perform detection exclusively in the continuous \pspace, where geometric priors (\emph{e.g.}, homographies or fundamental matrices) can be meaningfully exploited. In this setup we  compute preferences on $m = 6|X|$ models.
            
            Parameters of \ifor, \eifor, \vifor and \rzhifor are fixed to $t = 100, \psi = 256$ and $b = 2$ in all experiments following standard practice  \ifor~\cite{LiuTingAl12}. Regarding \lof, we vary $k$, ranging from $k = 10$ to $k = 500$, and keep the best result.
            
            \begin{table*}[t]
                \footnotesize
                \setstretch{1.35}
                \centering
                \caption{Synthetic datasets AUCs}
                \resizebox{\textwidth}{!}{
                \begin{tabular}{lllllllllllllll}
                    \toprule
                            & \multicolumn{4}{l}{Euclidean} & \multicolumn{4}{l}{Preference binary} & \multicolumn{4}{l}{Preference}  \\
                            \midrule
                            & \lof $\ell_2$  & \ifor   & \eifor  & \vifor $\ell_2$   & \lof jac & \ifor   & \eifor  & \vifor jac   & \lof tani                    & \ifor   & \eifor  & \vifor tani                  & \vifor ruz & \rzhifor              \\
                            \midrule
                    stair3  & $0.737$        & $0.925$ & $0.920$ & $0.918$           & $0.904$  & $0.885$ & $0.864$ & $0.958$      & $0.815$                      & $0.923$ & $0.925$ & \underline{$0.971$}          & $0.969$    & $0.959$                \\
                    stair4  & $0.814$        & $0.889$ & $0.874$ & $0.871$           & $0.849$  & $0.855$ & $0.860$ & $0.941$      & $0.881$                      & $0.912$ & $0.908$ & $0.952$                      & $0.950$    & \underline{$\mathbf{0.965}$}    \\
                    star5   & $0.771$        & $0.722$ & $0.738$ & $0.788$           & $0.875$  & $0.745$ & $0.769$ & $0.872$      & \underline{$\mathbf{0.929}$} & $0.761$ & $0.822$ & $0.910$                      & $0.869$    & $0.841$                \\
                    star11  & $0.671$        & $0.728$ & $0.727$ & $0.738$           & $0.830$  & $0.739$ & $0.741$ & $0.771$      & \underline{$\mathbf{0.900}$} & $0.738$ & $0.774$ & $0.796$                      & $0.817$    & $0.714$                \\
                    circle3 & $0.761$        & $0.698$ & $0.732$ & $0.779$           & $0.719$  & $0.842$ & $0.854$ & $0.900$      & $0.731$                      & $0.854$ & $0.891$ & \underline{$\mathbf{0.930}$} & $0.851$    & $0.871$                \\
                    circle4 & $0.640$        & $0.641$ & $0.665$ & $0.679$           & $0.827$  & $0.686$ & $0.699$ & $0.860$      & \underline{$0.906$}          & $0.667$ & $0.720$ & $0.897$                      & $0.777$    & $0.722$                \\
                    circle5 & $0.543$        & $0.569$ & $0.570$ & $0.633$           & $0.699$  & $0.597$ & $0.617$ & $0.672$      & \underline{$\mathbf{0.823}$} & $0.573$ & $0.593$ & $0.780$                      & $0.565$    & $0.638$                \\
                    \midrule
                    Mean    & $0.705$        & $0.739$ & $0.747$ & $0.772$           & $0.815$  & $0.764$ & $0.772$ & $0.853$      & $0.855$                      & $0.775$ & $0.805$ & \underline{$\mathbf{0.891}$} & $0.828$    & $0.816$                \\
                \bottomrule
                \end{tabular}
                }
                \label{tab:lines_and_circles}
            \end{table*}
            
            \begin{table*}[t]
                \setstretch{1.35}
                \caption{Real datasets AUCs}
                \begin{subtable}{.49\textwidth}
                    \footnotesize
                    \centering
                    \caption{Homographies}
                    \resizebox{.96\textwidth}{!}{
                    \begin{tabular}{lllllll}
                            \toprule
                                        & \lof tani                    & \ifor               & \eifor   & \vifor tani                  & \vifor ruz                       & \rzhifor \\
                                        \midrule
                        barrsmith       & \underline{$\mathbf{0.969}$} & $0.708$             & $0.715$  & $0.944$                      & $0.698$                          & $0.692$  \\
                        bonhall         & $0.918$                      & \underline{$0.969$} & $0.967$  & $0.949$                      & $0.960$                          & $0.951$  \\
                        bonython        & \underline{$\mathbf{0.978}$} & $0.679$             & $0.691$  & $0.954$                      & $0.911$                          & $0.857$  \\
                        elderhalla      & \underline{$0.999$}          & $0.925$             & $0.909$  & \underline{$\mathbf{0.999}$} & $0.878$                          & $0.877$  \\
                        elderhallb      & $0.986$                      & $0.924$             & $0.943$  & \underline{$\mathbf{0.999}$} & $0.966$                          & $0.976$  \\
                        hartley         & $0.963$                      & $0.749$             & $0.793$  & \underline{$\mathbf{0.989}$} & $0.911$                          & $0.882$  \\
                        johnsona        & $0.993$                      & $0.993$             & $0.993$  & \underline{$\mathbf{0.998}$} & $0.962$                          & $0.993$  \\
                        johnsonb        & $0.776$                      & \underline{$0.999$} & $0.998$  & \underline{$0.999$}          & $0.929$                          & $0.980$  \\
                        ladysymon       & $0.847$                      & $0.944$             & $0.943$  & \underline{$\mathbf{0.997}$} & $0.942$                          & $0.983$  \\
                        library         & \underline{$\mathbf{1.000}$} & $0.764$             & $0.771$  & $0.998$                      & $0.968$                          & $0.936$  \\
                        napiera         & $0.975$                      & $0.869$             & $0.879$  & \underline{$\mathbf{0.983}$} & $0.756$                          & $0.791$  \\
                        napierb         & $0.888$                      & $0.931$             & $0.936$  & \underline{$\mathbf{0.953}$} & $0.938$                          & $0.948$  \\
                        neem            & $0.985$                      & $0.896$             & $0.906$  & \underline{$\mathbf{0.996}$} & $0.928$                          & $0.969$  \\
                        nese            & \underline{$\mathbf{0.996}$} & $0.888$             & $0.892$  & $0.980$                      & $0.920$                          & $0.958$  \\
                        oldclassicswing & $0.936$                      & $0.923$             & $0.943$  & $0.987$                      & \underline{$\mathbf{0.998}$}     & $0.993$  \\
                        physics         & $0.670$                      & $0.858$             & $0.787$  & \underline{$\mathbf{1.000}$} & $0.989$                          & $0.969$  \\
                        sene            & \underline{$\mathbf{0.997}$} & $0.698$             & $0.731$  & $0.988$                      & $0.936$                          & $0.763$  \\
                        unihouse        & $0.785$                      & $0.998$             & $0.998$  & \underline{$\mathbf{0.999}$} & $0.930$                          & $0.980$  \\
                        unionhouse      & \underline{$\mathbf{0.987}$} & $0.639$             & $0.664$  & $0.968$                      & $0.877$                          & $0.840$  \\
                        \midrule
                        Mean            & $0.929$                      & $0.861$             & $0.866$  & \underline{$\mathbf{0.983}$} & $0.916$                          & $0.913$  \\
                        \bottomrule
                    \end{tabular}
                    }
                    \label{tab:homographies}
                \end{subtable}
                \hfill
                \begin{subtable}{.49\textwidth}
                    \footnotesize
                    \centering
                    \caption{Fundamental matrices}
                    \resizebox{\textwidth}{!}{
                    \begin{tabular}{lllllll}
                            \toprule
                                          & \lof tani                    & \ifor               & \eifor              & \vifor tani                  & \vifor ruz   & \rzhifor                     \\
                                          \midrule
                        biscuit           & $0.976$                      & $0.994$             & $0.996$             & \underline{$\mathbf{1.000}$} & $0.922$      & $0.967$                      \\
                        biscuitbook       & \underline{$1.000$}          & $0.987$             & $0.988$             & \underline{$1.000$}          & $0.945$      & $0.986$                      \\
                        biscuitbookbox    & \underline{$\mathbf{1.000}$} & $0.990$             & $0.989$             & $0.996$                      & $0.895$      & $0.982$                      \\
                        boardgame         & \underline{$\mathbf{0.962}$} & $0.400$             & $0.304$             & $0.949$                      & $0.750$      & $0.883$                      \\
                        book              & $0.996$                      & \underline{$1.000$} & \underline{$1.000$} & \underline{$1.000$}          & $0.925$      & $0.982$                      \\
                        breadcartoychips  & \underline{$\mathbf{0.989}$} & $0.978$             & $0.971$             & $0.976$                      & $0.827$      & $0.975$                      \\
                        breadcube         & \underline{$\mathbf{1.000}$} & $0.998$             & $0.998$             & $0.999$                      & $0.777$      & $0.979$                      \\
                        breadcubechips    & \underline{$\mathbf{0.999}$} & $0.985$             & $0.985$             & $0.998$                      & $0.861$      & $0.976$                      \\
                        breadtoy          & $0.984$                      & \underline{$0.999$} & $0.998$             & \underline{$0.999$}          & $0.920$      & $0.984$                      \\
                        breadtoycar       & \underline{$\mathbf{0.998}$} & $0.933$             & $0.883$             & $0.991$                      & $0.827$      & $0.957$                      \\
                        carchipscube      & \underline{$\mathbf{0.993}$} & $0.981$             & $0.966$             & $0.987$                      & $0.829$      & $0.975$                      \\
                        cubetoy           & \underline{$\mathbf{1.000}$} & $0.997$             & $0.995$             & \underline{$1.000$}          & $0.960$      & $0.987$                      \\
                        cube              & \underline{$0.999$}          & $0.970$             & $0.982$             & \underline{$0.999$}          & $0.895$      & $0.952$                      \\
                        cubebreadtoychips & \underline{$0.990$}          & $0.962$             & $0.958$             & $0.989$                      & $0.752$      & $0.930$                      \\
                        cubechips         & \underline{$\mathbf{1.000}$} & $0.995$             & $0.994$             & \underline{$1.000$}          & $0.980$      & $0.996$                      \\
                        dinobooks         & $0.887$                      & $0.873$             & $0.857$             & $0.899$                      & $0.879$      & \underline{$\mathbf{0.946}$} \\
                        game              & \underline{$\mathbf{1.000}$} & $0.901$             & $0.895$             & $0.999$                      & $0.844$      & $0.898$                      \\
                        gamebiscuit       & \underline{$1.000$}          & $0.985$             & $0.988$             & \underline{$\mathbf{1.000}$} & $0.876$      & $0.981$                      \\
                        toycubecar        & \underline{$\mathbf{0.973}$} & $0.290$             & $0.192$             & $0.964$                      & $0.742$      & $0.935$                      \\
                        \midrule
                        Mean              & \underline{$0.987$}          & $0.906$             & $0.891$             & \underline{$0.987$}          & $0.864$      & $0.962$                      \\
                        \bottomrule
                    \end{tabular}
                    }
                    \label{tab:fundamentals}
                \end{subtable}
            \end{table*}
            
            The AD performance is evaluated with AUC averaged over $10$ runs (\cref{tab:lines_and_circles,tab:homographies,tab:fundamentals}).
            The highest value for each dataset is underlined, and boldface indicates statistically better methods, according to paired t-test ($\alpha = 0.05$).
            \lof results refer to the $k$ with highest AUC along datasets of each family $\mathcal{F}$, for all the embeddings ($k$ values reported in~\cref{tab:neighborhood}).
                
            \cref{tab:lines_and_circles} shows that all methods improve their performance when run in \pspace, thus confirming the benefits of using the prior $\mathcal{F}$ rather than working with density in the ambient space, with continuous preferences to be preferred. \lof performs well on \emph{star5}, \emph{star11}, \emph{circle4} and \emph{circle5}, where genuine data points are evenly distributed among structures, \emph{i.e.},  the different structures have similar cardinality. In such balanced settings, it is easier to choose an appropriate neighborhood size $k$, as each inlier has a sufficient number of nearby points belonging to the same structure, reducing the risk of misclassification due to uneven density. By contrast, \lof performs poorly when structures have different cardinalities sizes (\emph{stair3} and \emph{circle3}), while \vifor and \rzhifor (run with fixed parameters) achieve superior performance, with \vifor the best statistically significant method. \vifor and \rzhifor outperform \ifor and \eifor as they leverage the most suitable distances for \pspace.
            On real data, high AUC values in~\cref{tab:homographies} and~\cref{tab:fundamentals} show that AD is easier than the synthetic case, still the performance improvement \vifor, \rzhifor and \lof over \ifor and \eifor remains evident.
                
        \subsection{The impact of hashing on efficiency}
            \label{subsec:impact_hashing}

            \begin{figure}[t]
               \centering
               \includegraphics[width=.6\linewidth]{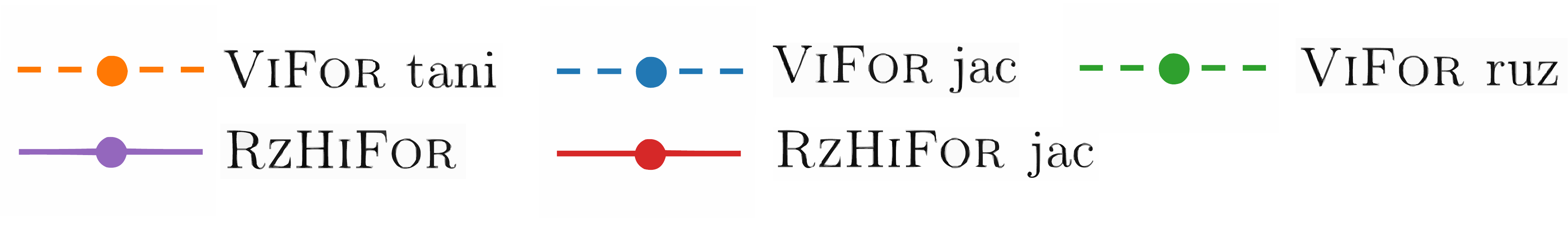}
                \\
                \hspace*{-1.2cm}
                \captionsetup[subfigure]{oneside, margin={0.75cm,0cm}}
                \subfloat[\label{fig:roc_auc_branching}]{
                    \includegraphics[width=.25\linewidth]{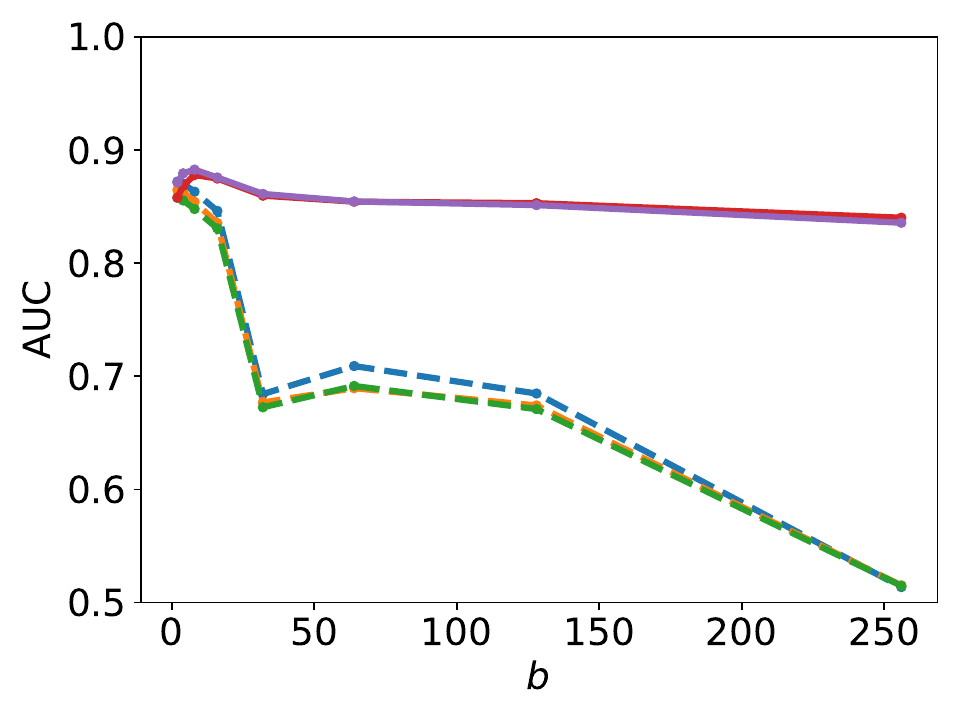}
                    \vspace{-0.25cm}
                }
                \subfloat[\label{fig:test_time_branching}]{
                    \includegraphics[width=.25\linewidth]{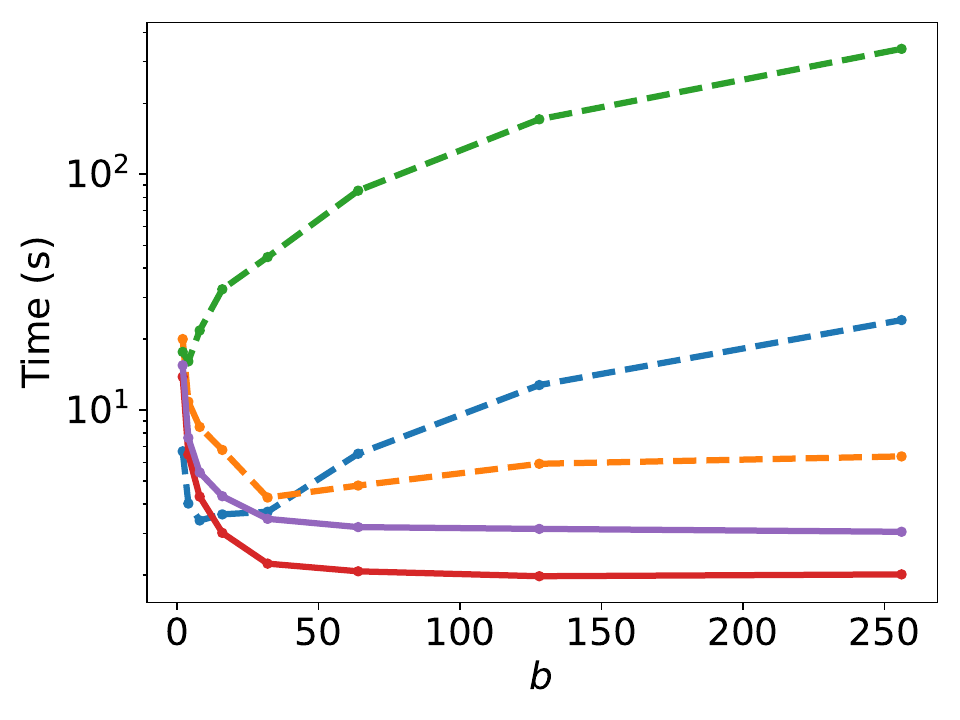}
                    \vspace{-0.25cm}
                }
                \subfloat[\label{fig:best_roc_auc_time_branching}]{
                    \includegraphics[width=.25\linewidth]{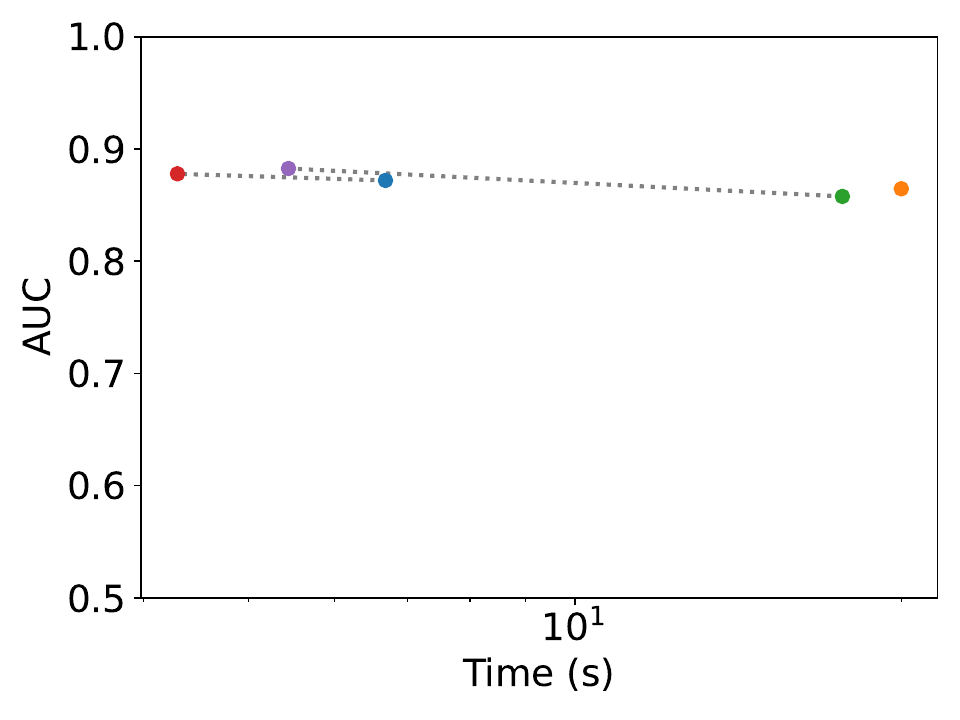}
                    \vspace{-0.25cm}
                }
                \caption{Results on synthetic and real datasets, averaged together, at various branching factors. (a) Average AUCs. (b) Average test times. (c) Best average AUCs and corresponding test time.}
                \label{fig:results_branching}
            \end{figure}

            \begin{figure}[t]
               \centering
               \hspace*{-1.2cm}
               \captionsetup[subfigure]{oneside, margin={0.75cm,0cm}}
                \subfloat[\label{fig:roc_auc_models}]{
                    \includegraphics[width=.25\linewidth]{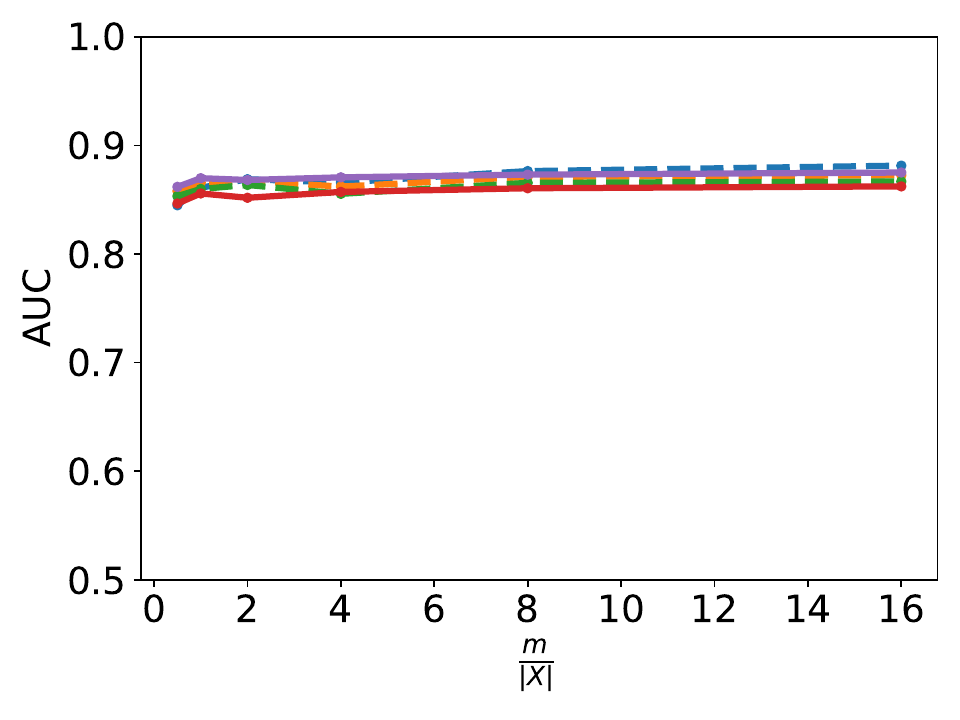}
                    \vspace{-0.25cm}
                }
                \subfloat[\label{fig:test_time_models}]{
                    \includegraphics[width=.25\linewidth]{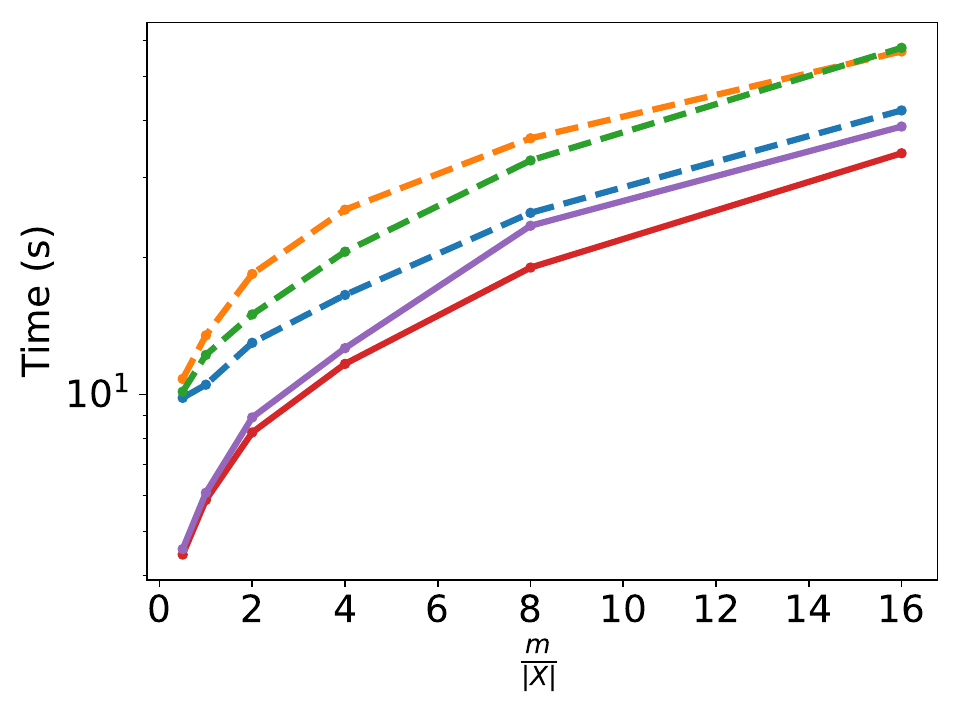}
                    \vspace{-0.25cm}
                }
                \subfloat[\label{fig:best_roc_auc_time_models}]{
                    \includegraphics[width=.25\linewidth]{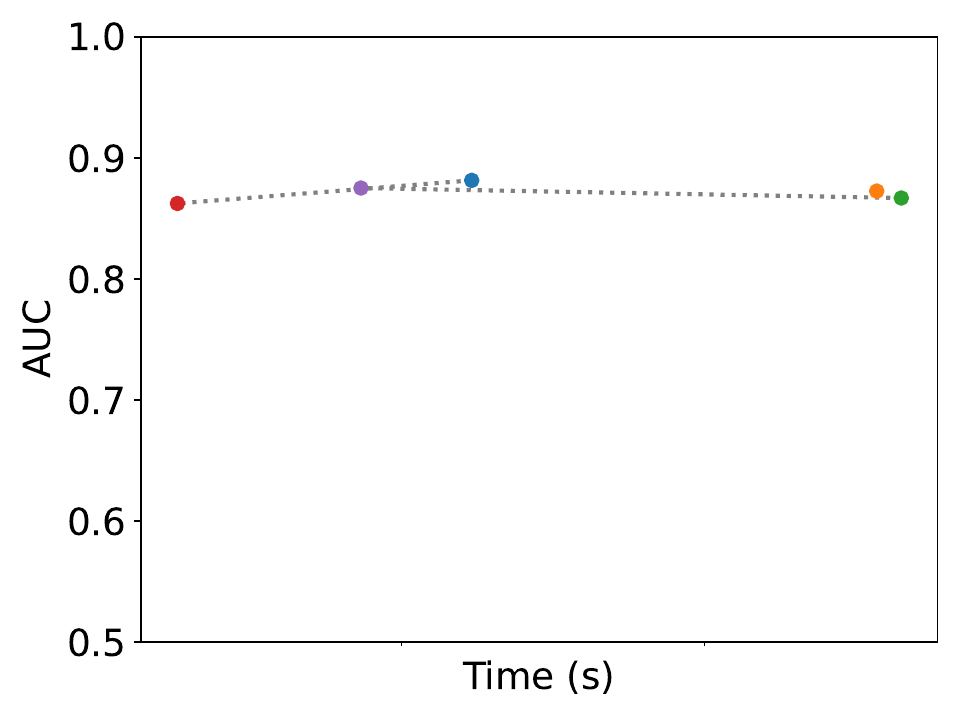}
                    \vspace{-0.1cm}
                }
                \caption{Results on synthetic and real datasets, averaged together, at various number of models. (a) Average AUCs. (b) Average test times. (c) Best average AUCs and corresponding test time.}
                \label{fig:results_models}
            \end{figure}

            We perform an extensive comparison between \rzhiforest and \viforest both in the binary and continuous \pspace on the same datasets used for~\cref{subsec:importance_pembedding}.
            We tested \viforest and \rzhiforest with same parameters: $t = 100$, $\psi = 256$, but varying branching factors in $b \in \{2, 4, 8, 16, 32, 64, 128, 256\}$ and number of models  $m \in \{0.5|X|, |X|, 2|X|, 4|X|, 8|X|, 16|X|\}$.
            
            \cref{fig:roc_auc_branching} and~\cref{fig:test_time_branching} show the average AUC and test time curves at various $b$ values, produced by averaging the results of $5$ executions on synthetic and real datasets separately, then averaging them together.
            \rzhiforest shows the most stable AUC with respect to $b$, and an evident efficiency gain consistent with the complexities in~\cref{tab:differences}.
            \cref{fig:best_roc_auc_time_branching} relates the best AUC and corresponding test time of methods referring to the same distance function. \rzhiforest is as accurate as \viforest, but with gain in time. \rzhifor jac is $\times 35\%$ faster than \vifor jac, and \rzhifor is $\times 70\%$ faster than \vifor ruz.

            Similarly, \cref{fig:roc_auc_models} and~\cref{fig:test_time_models} show the average AUC and test time curves for various values of $m$. All methods exhibit stable performance across different $m$ values, indicating that the sensitivity to this parameter is less critical. Increasing $m$ generally leads to longer execution times without yielding significant improvements in detection performance.

        \subsection{The impact of sliding window}
            \label{subsec:impact_sliding}
            
            We assess \spif on a real-world scenario, namely the MVTec-3D AD dataset, using families of local models $\mathcal{F}$ with increasing generality (planes, spheres, and quadrics). Notably, we do not rely on any ground-truth genuine or anomalous data, since \spif is a truly unsupervised method. We ran \spif with different window sizes $\omega$ to account for various defect sizes ($\omega = [\delta, \frac{\delta}{2}, \frac{\delta}{4}, \frac{\delta}{10}, \frac{\delta}{20}]$, where $\delta$ is the image side length). For \pisolation we used \rzhiforest in the continuous \pspace with $t = 100$, $\psi = 256$ and $b = 16$ for greater efficiency.
            
            We evaluate the impact of the window size $\omega$  on qualitative experiments. \cref{tab:sliding_pif_smooth_qualitative} compares \pif ($\omega = \delta$) to \spif with smallest window size ($\omega = \frac{\delta}{20}$), alongside ground truths. \cref{tab:sliding_pif_smooth_quantitative} shows average AUC on $5$ runs achieved by \pif and \spif at different $\omega$.

            \begin{table}[!t]
                \centering
                \caption{(a) Qualitative results of \pif ($\omega = \delta$) and \spif ($\omega = \frac{\delta}{20}$), alongside ground truths, when executed on smooth objects. Anomaly scores $\alpha(\cdot)$ color-coded (red indicates anomalies), and AUC values displayed below objects. (b) Average AUC values on $5$ runs achieved by \pif and \spif at different $\omega$.}
                
                \hspace*{-1.2cm}
                \resizebox{.73\textwidth}{!}{
                \begin{subtable}[b]{.8\textwidth}
                    \centering
                    \caption{}
                    \resizebox{\textwidth}{!}{
                    \begin{tabular}{ll|ccccc}
                        $\mathcal{F}$ & $\omega$                                       & bagel & cookie & peach & potato & carrot \\
                        \midrule
                        \multirow{2}{*}{Plane}
                                      & \raisebox{0.06\textwidth}{$\delta$}            & \includegraphics[width=0.1\textwidth]{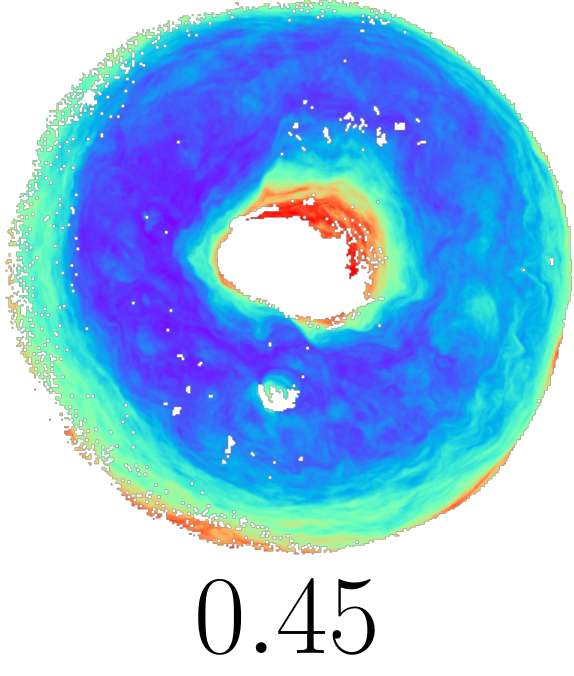}
                                                                                       & \includegraphics[width=0.1\textwidth]{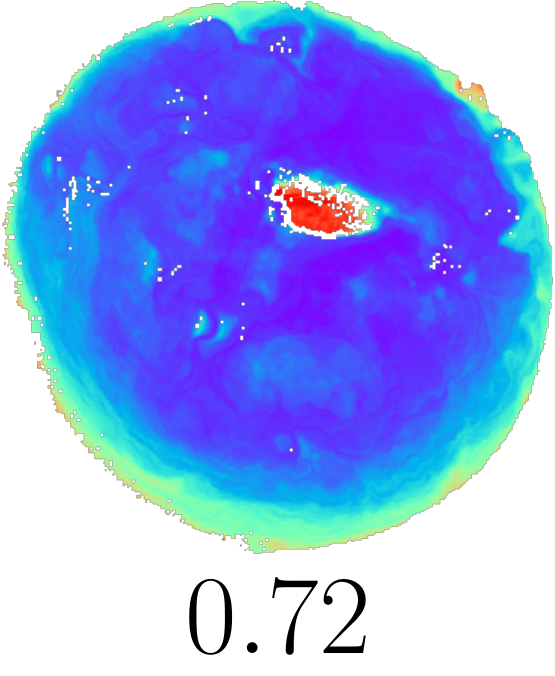}
                                                                                       & \includegraphics[width=0.1\textwidth]{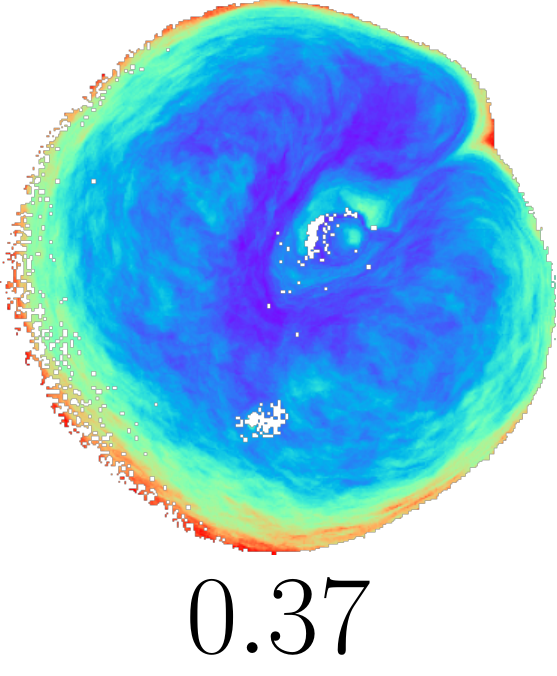}
                                                                                       & \includegraphics[width=0.16\textwidth]{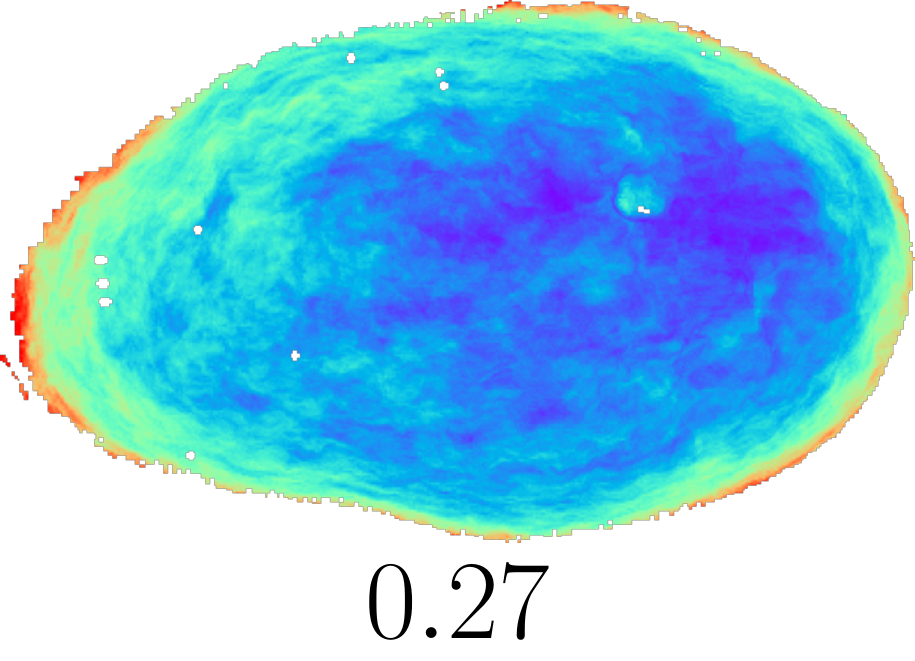}
                                                                                       & \includegraphics[width=0.16\textwidth]{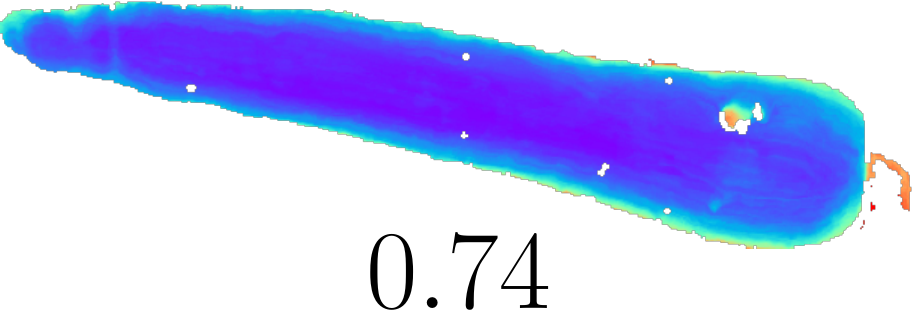} \\
                                      & \raisebox{0.06\textwidth}{$\frac{\delta}{20}$} & \includegraphics[width=0.1\textwidth]{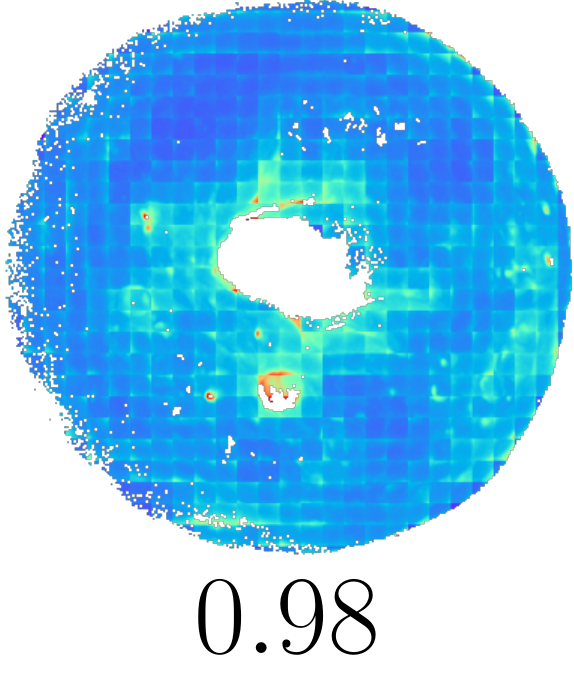}
                                                                                       & \includegraphics[width=0.1\textwidth]{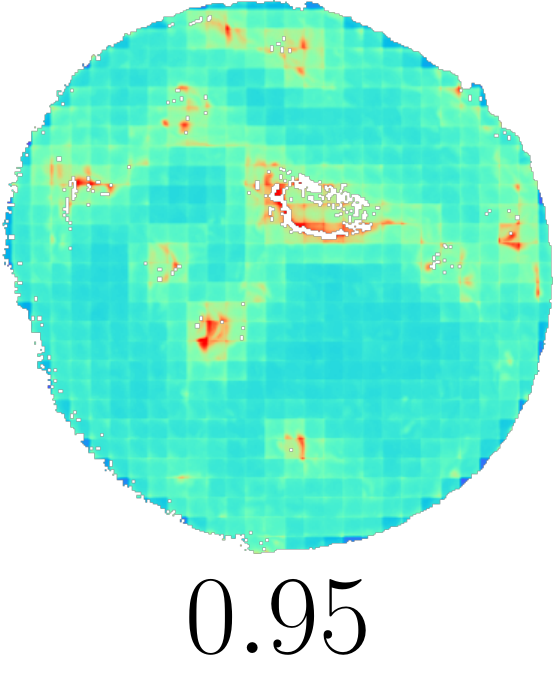}
                                                                                       & \includegraphics[width=0.1\textwidth]{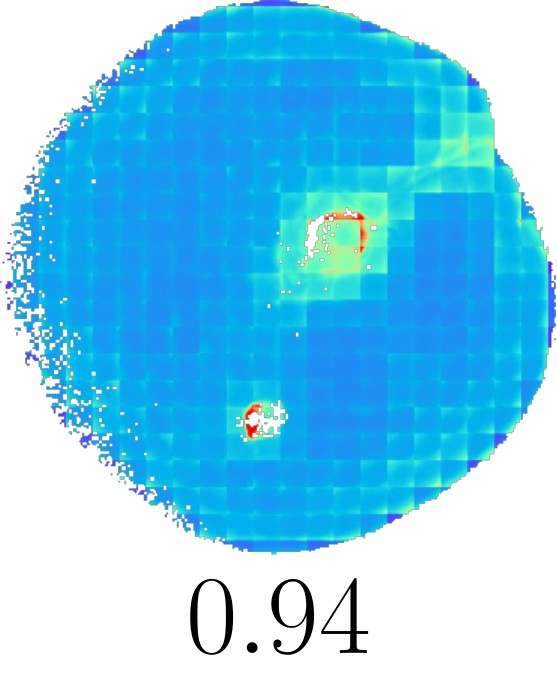}
                                                                                       & \includegraphics[width=0.16\textwidth]{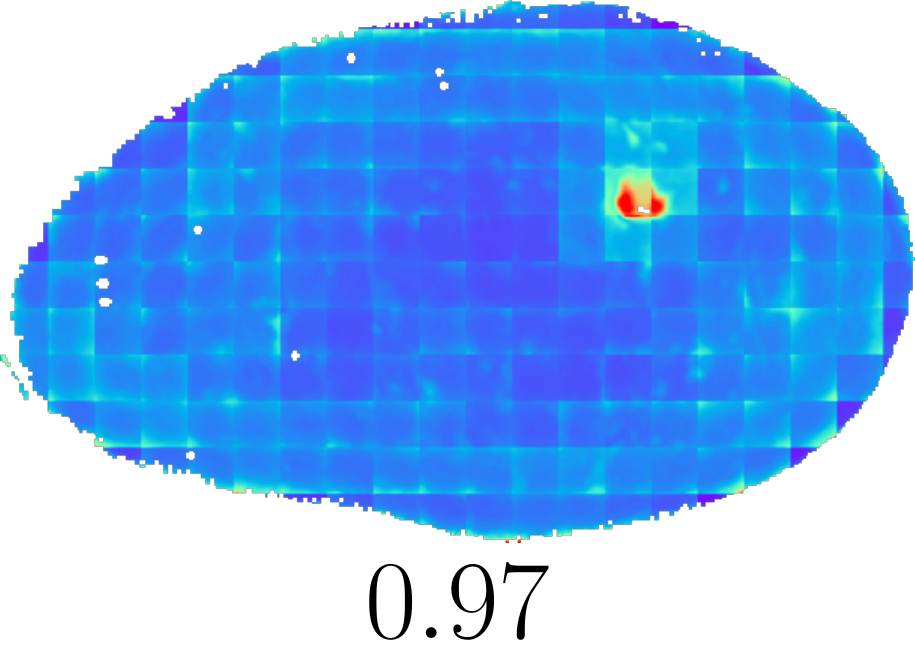}
                                                                                       & \includegraphics[width=0.16\textwidth]{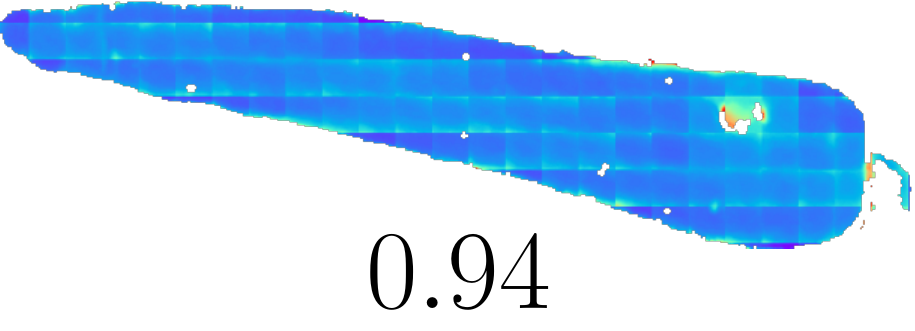} \\
                        \midrule
                        \multirow{2}{*}{Sphere}
                                      & \raisebox{0.06\textwidth}{$\delta$}            & \includegraphics[width=0.1\textwidth]{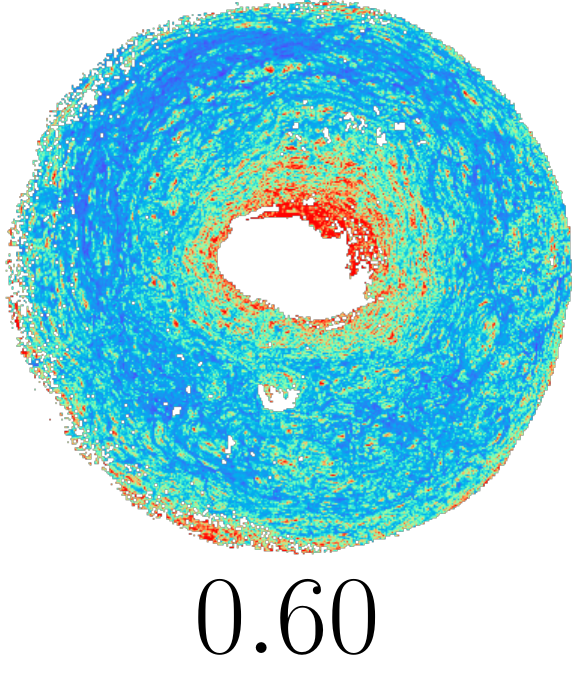}
                                                                                       & \includegraphics[width=0.1\textwidth]{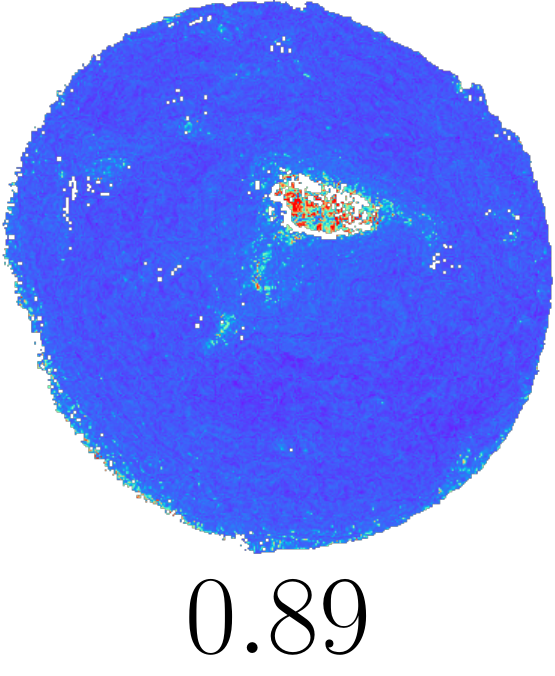}
                                                                                       & \includegraphics[width=0.1\textwidth]{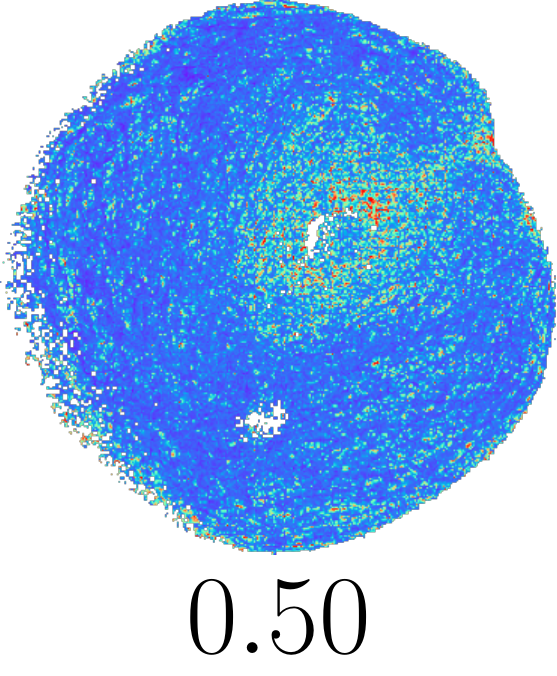}
                                                                                       & \includegraphics[width=0.16\textwidth]{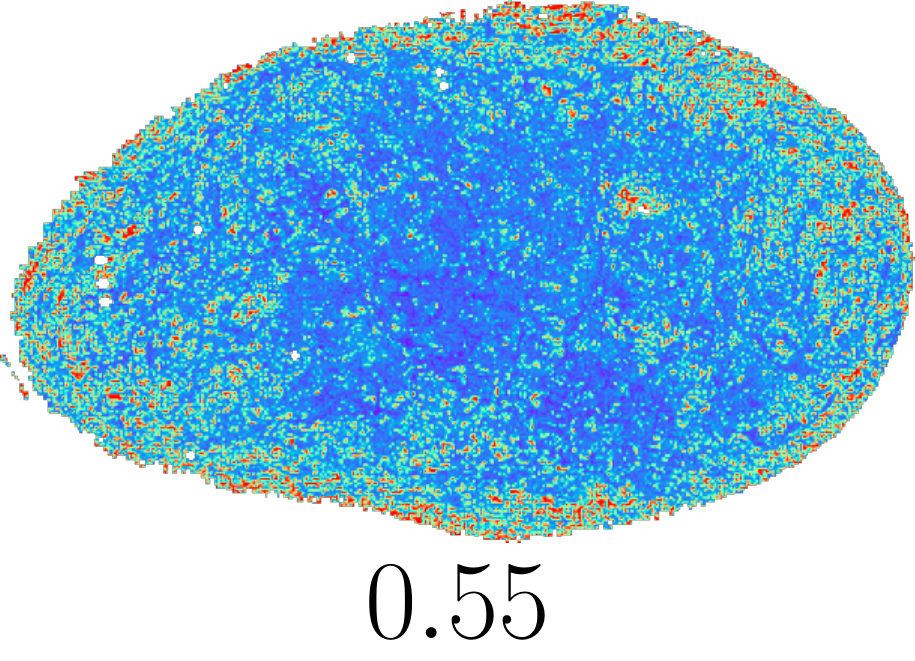}
                                                                                       & \includegraphics[width=0.16\textwidth]{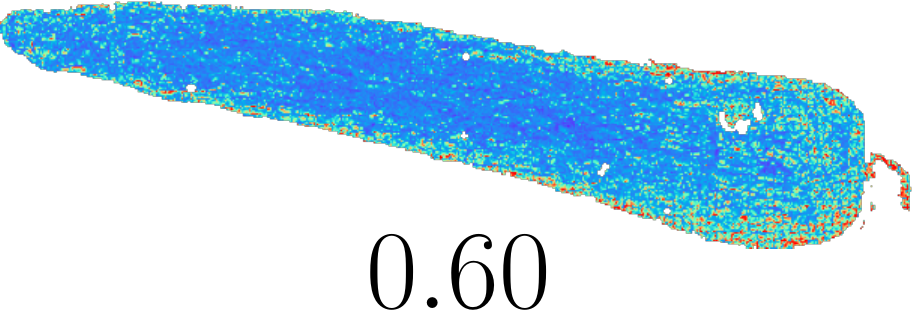} \\
                                      & \raisebox{0.06\textwidth}{$\frac{\delta}{20}$} & \includegraphics[width=0.1\textwidth]{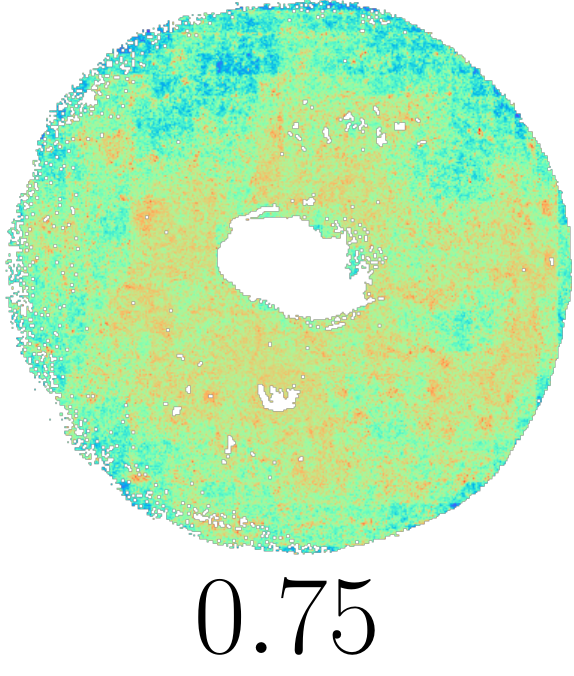}
                                                                                       & \includegraphics[width=0.1\textwidth]{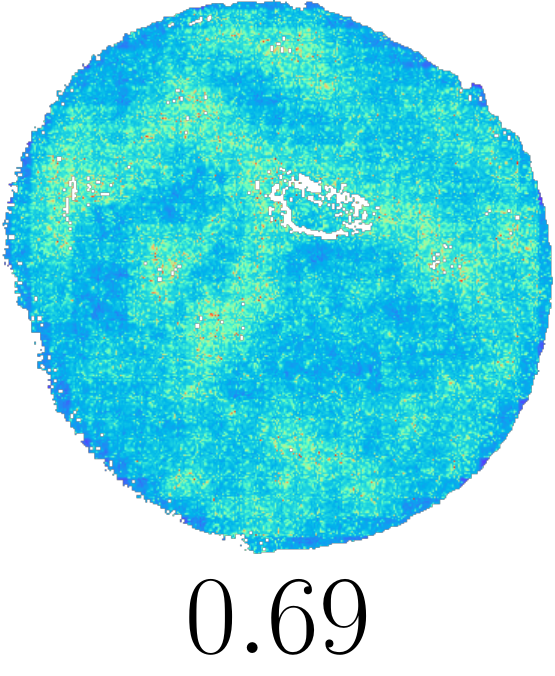}
                                                                                       & \includegraphics[width=0.1\textwidth]{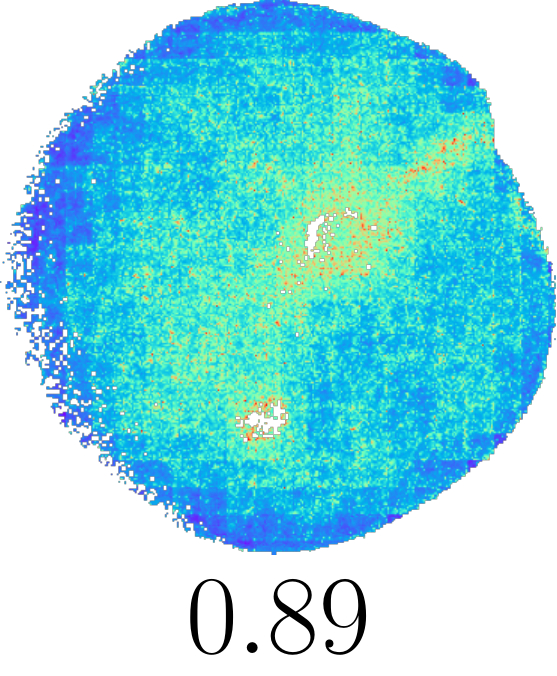}
                                                                                       & \includegraphics[width=0.16\textwidth]{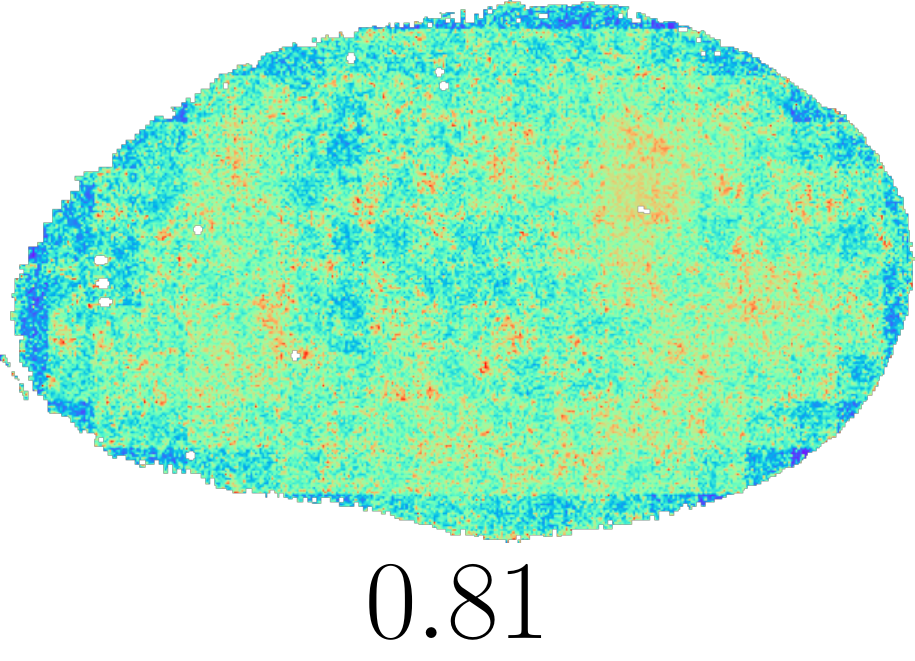}
                                                                                       & \includegraphics[width=0.16\textwidth]{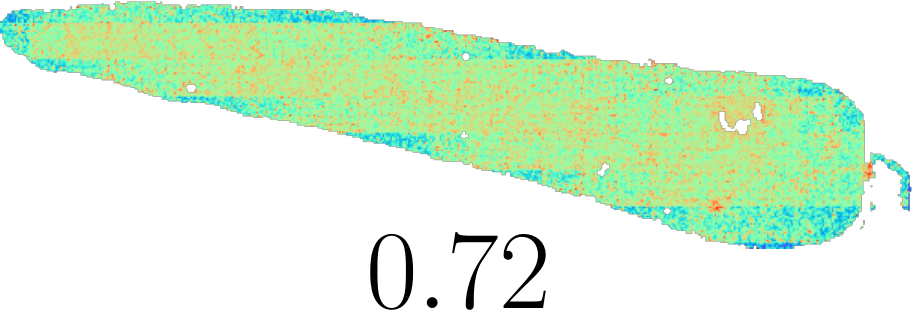} \\
                        \midrule
                        \multirow{2}{*}{Quadric}
                                      & \raisebox{0.06\textwidth}{$\delta$}            & \includegraphics[width=0.1\textwidth]{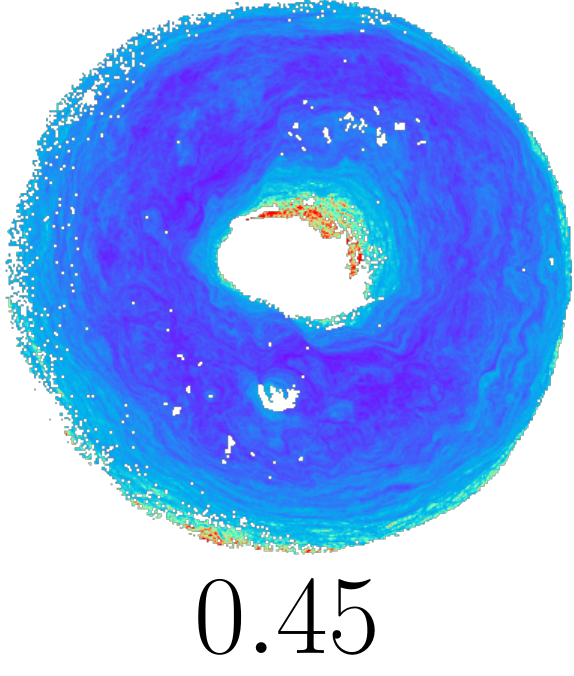}
                                                                                       & \includegraphics[width=0.1\textwidth]{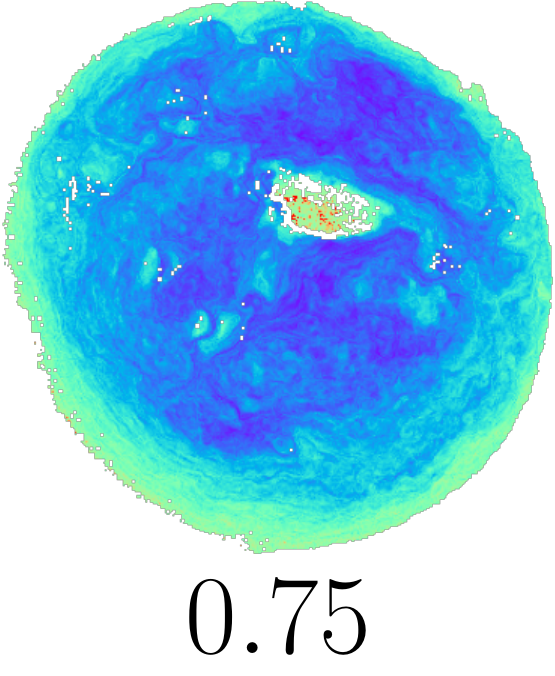}
                                                                                       & \includegraphics[width=0.1\textwidth]{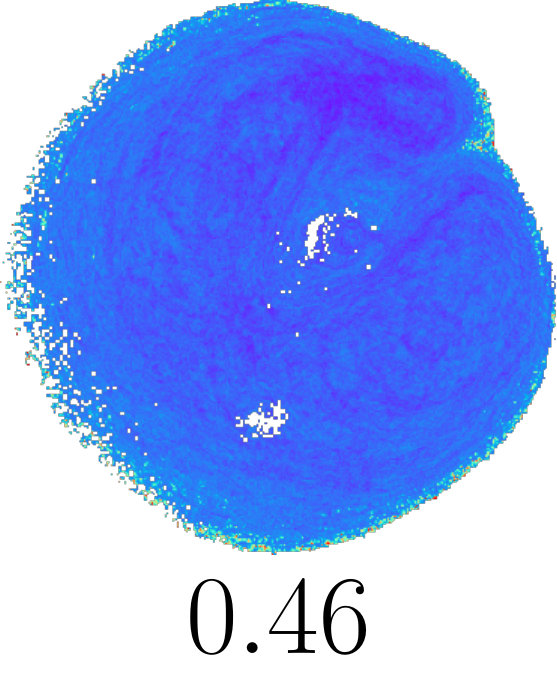}
                                                                                       & \includegraphics[width=0.16\textwidth]{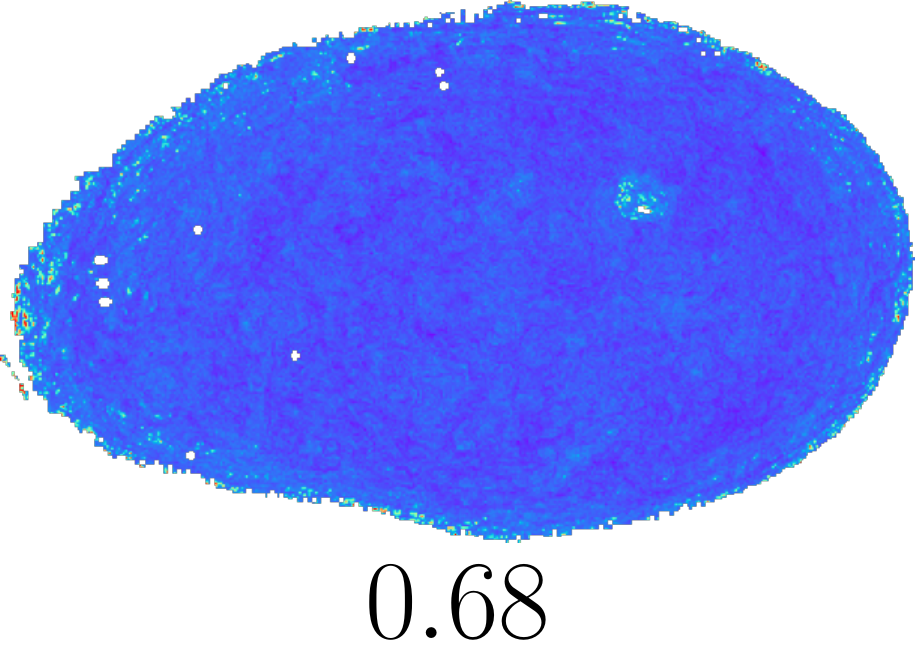}
                                                                                       & \includegraphics[width=0.16\textwidth]{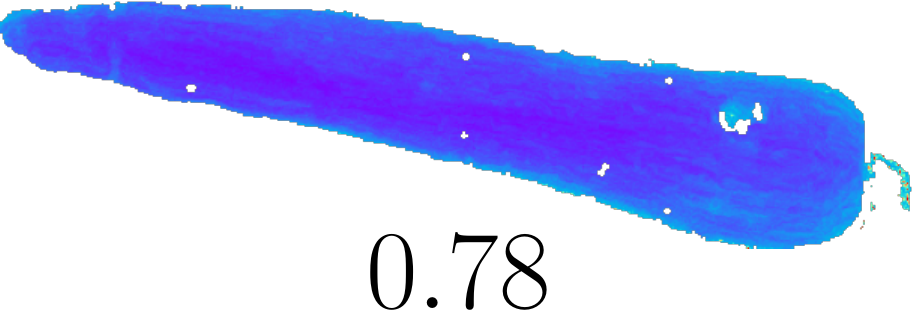} \\
                                      & \raisebox{0.06\textwidth}{$\frac{\delta}{20}$} & \includegraphics[width=0.1\textwidth]{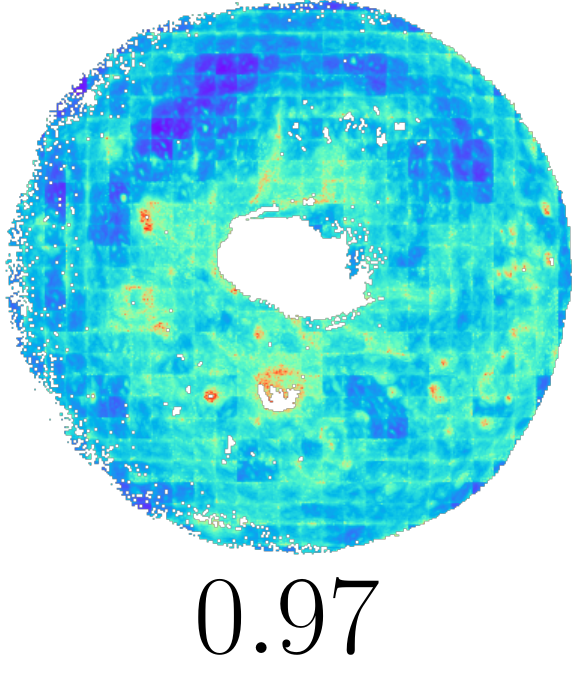}
                                                                                       & \includegraphics[width=0.1\textwidth]{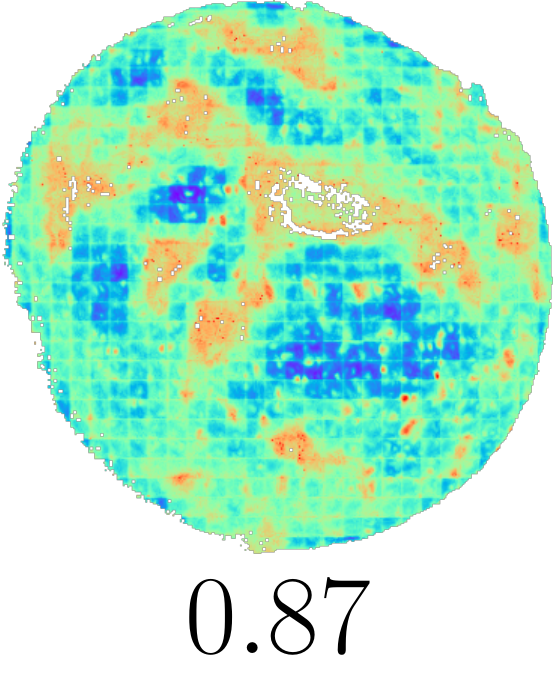}
                                                                                       & \includegraphics[width=0.1\textwidth]{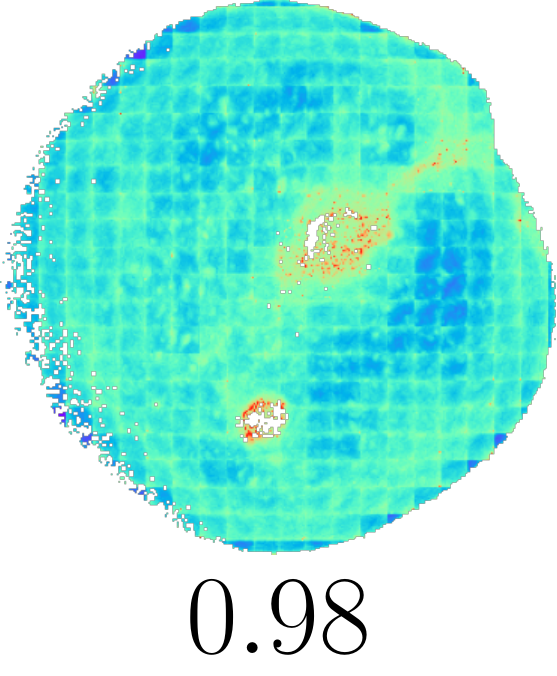}
                                                                                       & \includegraphics[width=0.16\textwidth]{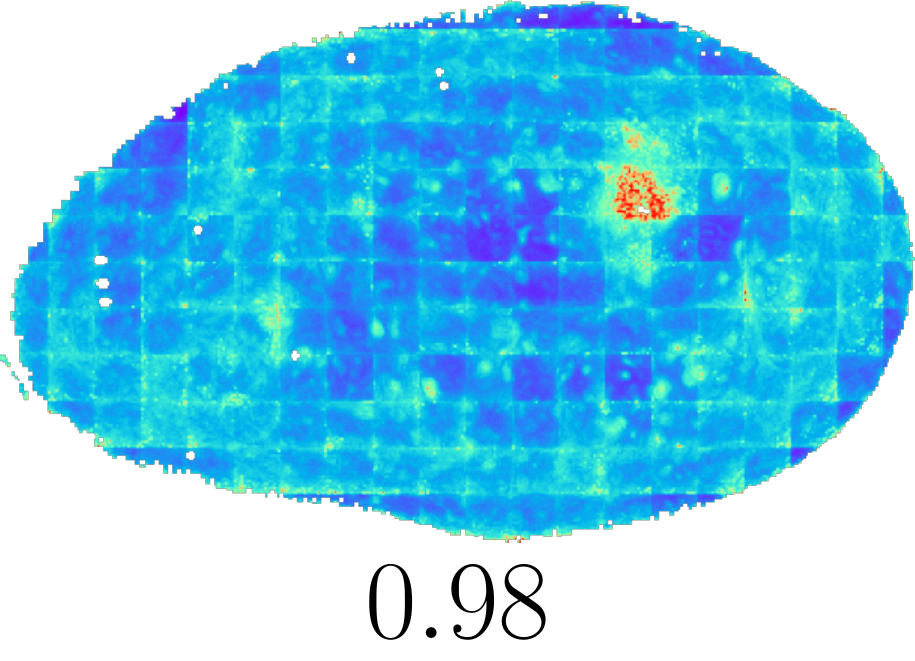}
                                                                                       & \includegraphics[width=0.16\textwidth]{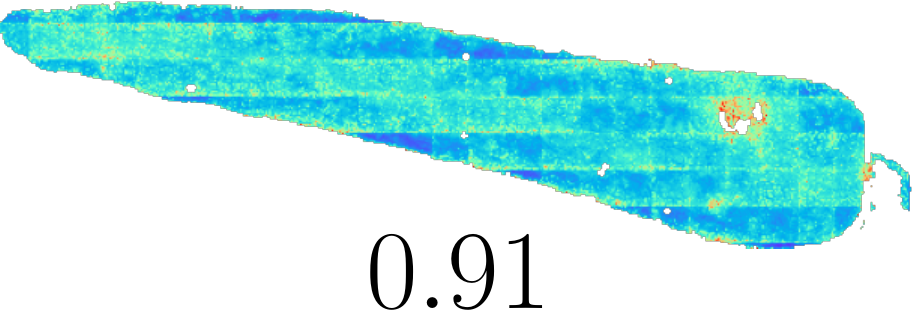} \\
                        \midrule
                        \midrule
                        \raisebox{0.05\textwidth}{\parbox{1cm}{\setlength{\baselineskip}{0.4cm}Ground\\truth}}
                                      &                                                & \includegraphics[width=0.1\textwidth]{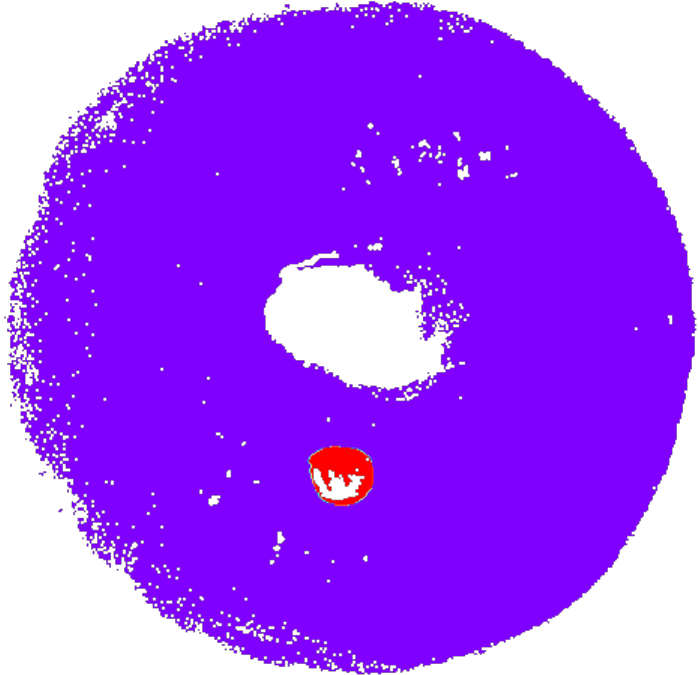}
                                                                                       & \includegraphics[width=0.1\textwidth]{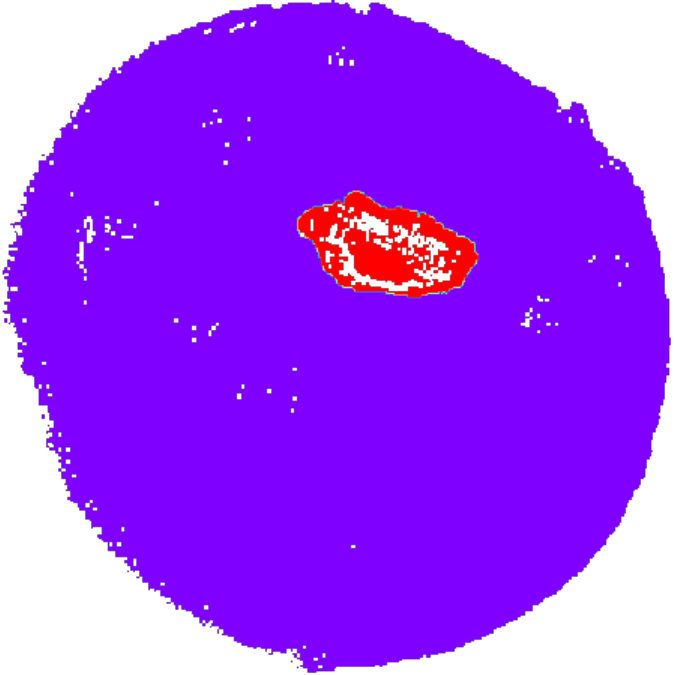}
                                                                                       & \includegraphics[width=0.1\textwidth]{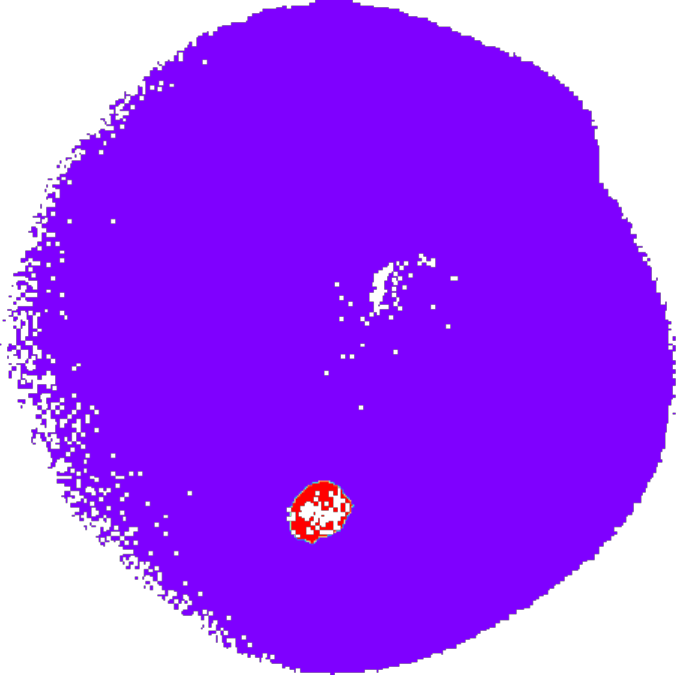}
                                                                                       & \includegraphics[width=0.16\textwidth]{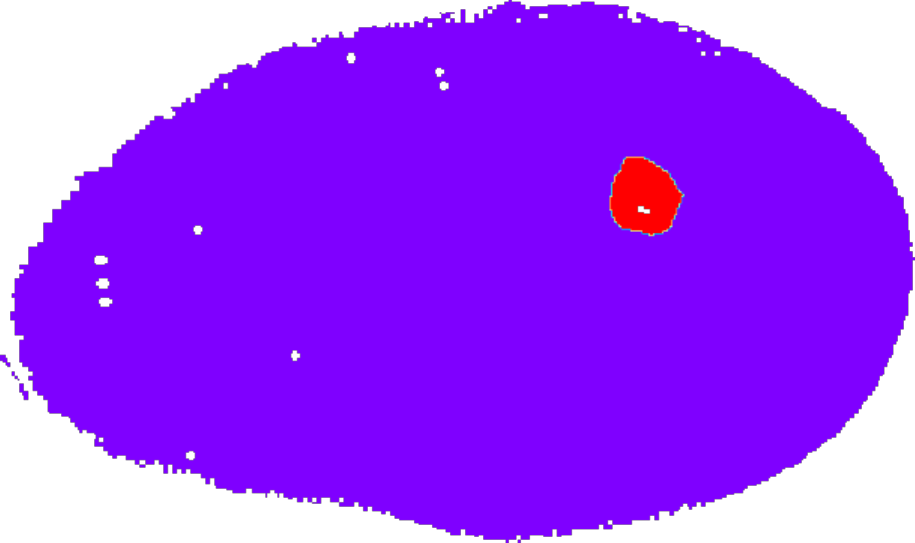}
                                                                                       & \includegraphics[width=0.16\textwidth]{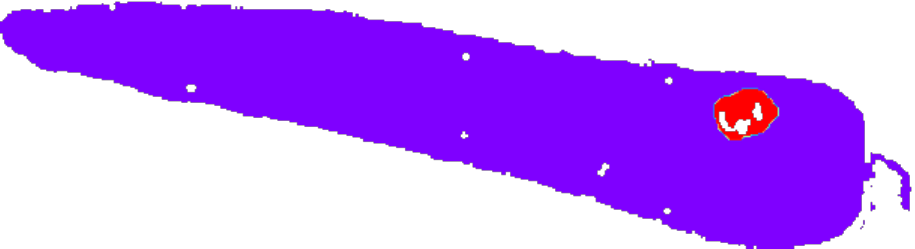} \\
                    \end{tabular}
                    }
                    \label{tab:sliding_pif_smooth_qualitative}
                \end{subtable}
                \hfill
                \begin{subtable}[b]{.275\textwidth}
                    \centering
                    \caption{}
                    \vspace{.9cm}
                    \begin{tabular}{l}
                        \includegraphics[width=\textwidth]{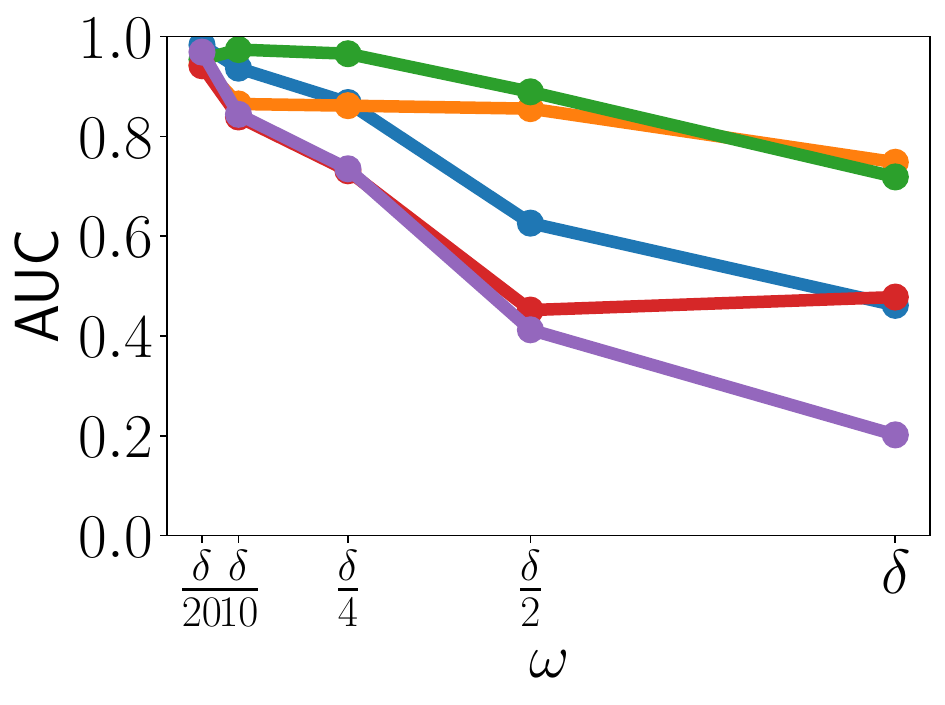}     \\
                        \vspace{.1cm}
                        \includegraphics[width=\textwidth]{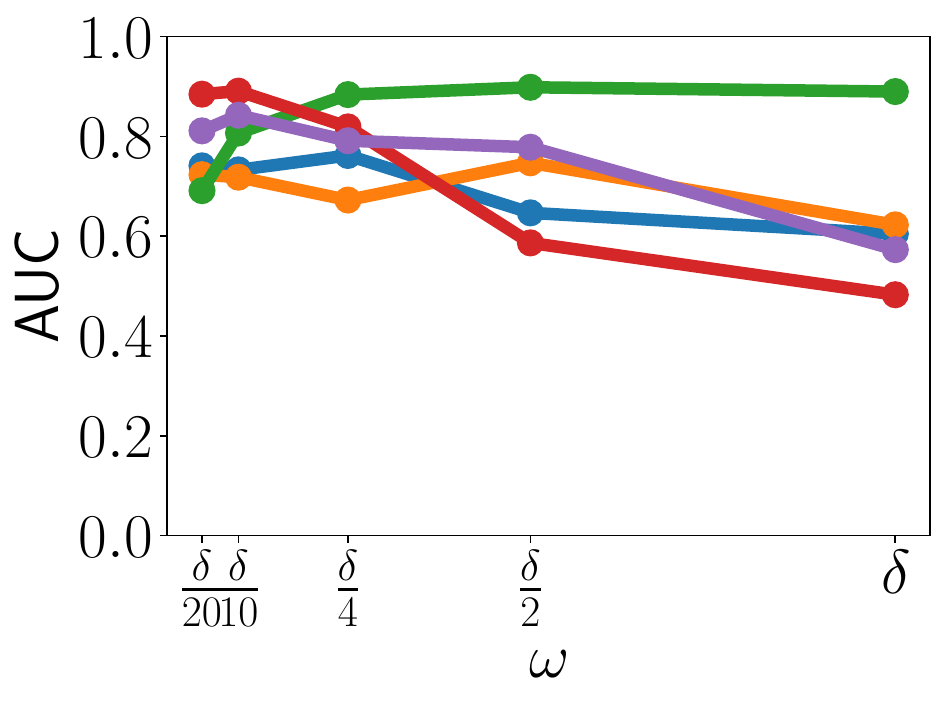}   \\
                        \vspace{.1cm}
                        \includegraphics[width=\textwidth]{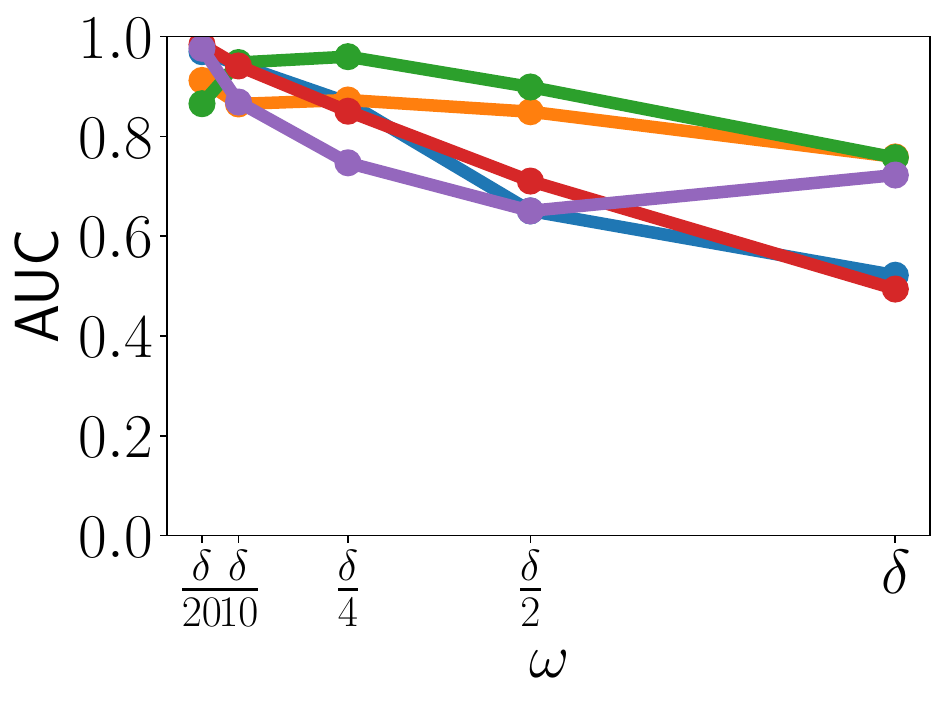} \\
                        \vspace{.1cm}
                        \parbox{\textwidth}{\vspace{-0.3cm} \includegraphics[width=1.2\textwidth]{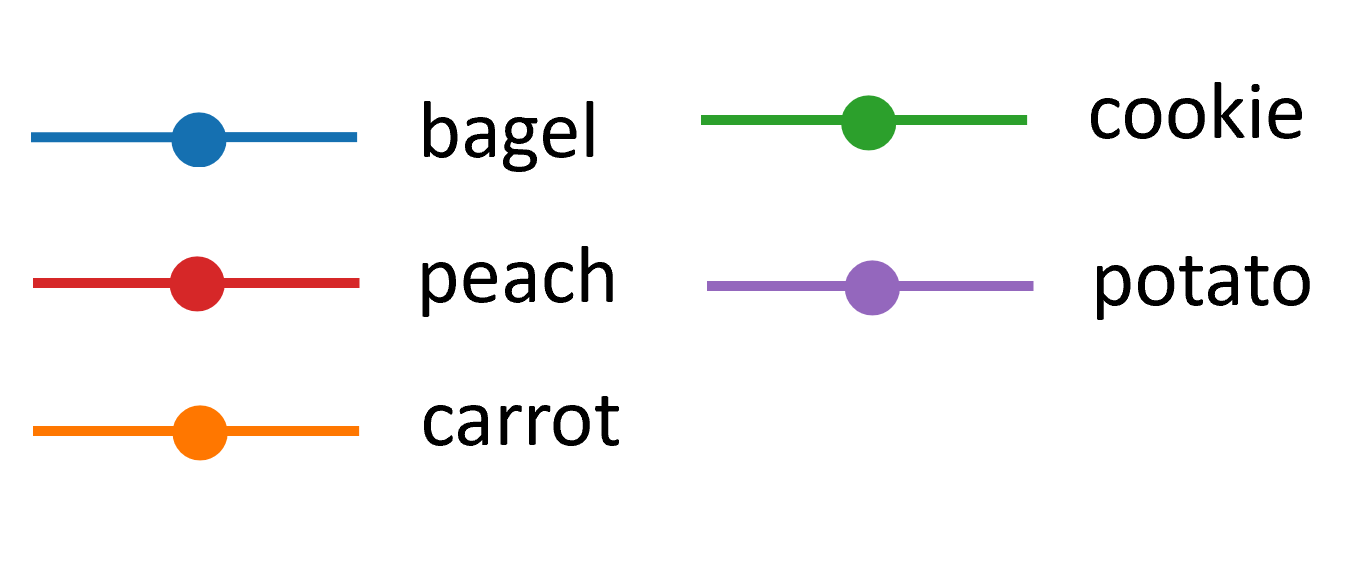}}                         \\
                        \vspace{-1.3cm}
                    \end{tabular}
                    \label{tab:sliding_pif_smooth_quantitative}
                \end{subtable}
                }
                \label{tab:sliding_pif_smooth}
            \end{table}

            \begin{table}[!t]
                \centering
                \caption{(a) Qualitative results of \pif ($\omega = \delta$) and \spif ($\omega = \frac{\delta}{20}$), alongside ground truths, when executed on non-smooth objects, alongside average AUC values (b).}
                
                \hspace*{-1.2cm}
                \resizebox{.73\textwidth}{!}{
                \begin{subtable}[b]{.8\textwidth}
                    \centering
                    \caption{}
                    \resizebox{\textwidth}{!}{
                    \begin{tabular}{ll|ccccc}
                        $\mathcal{F}$ & $\omega$                                       & cable\_gland & dowel & foam & rope & tire \\
                        \midrule
                        \multirow{2}{*}{Plane}
                                      & \raisebox{0.06\textwidth}{$\delta$}            & \includegraphics[width=0.1\textwidth]{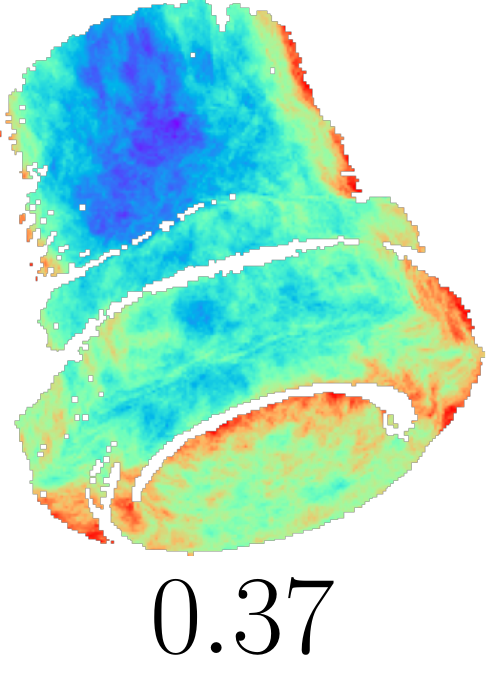}
                                                                                       & \includegraphics[width=0.1\textwidth]{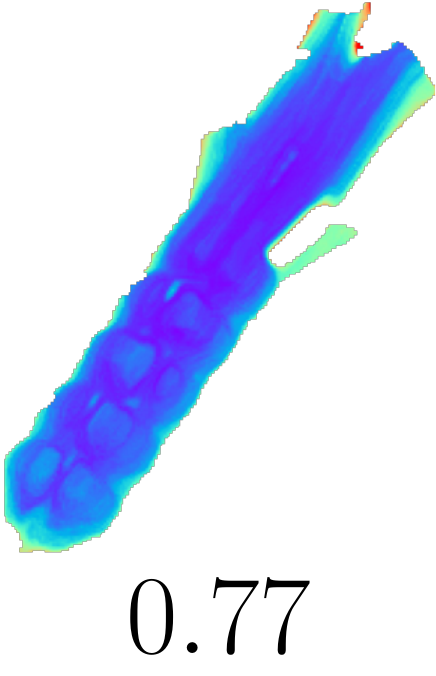}
                                                                                       & \includegraphics[width=0.1\textwidth]{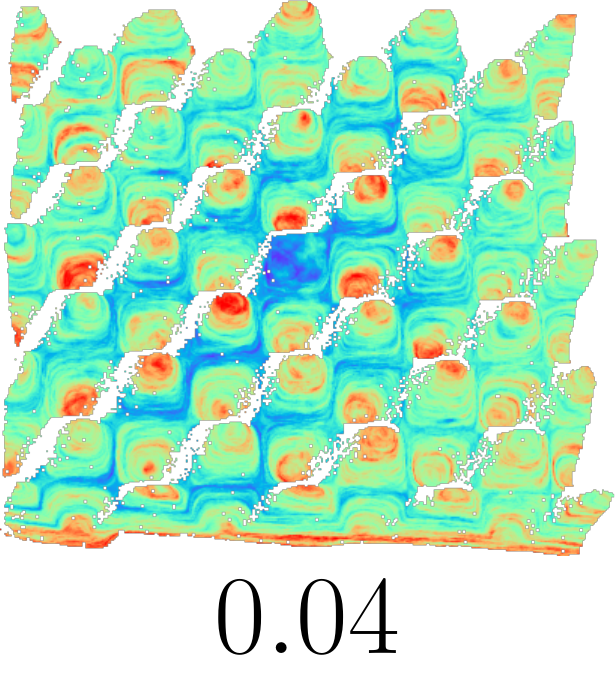}
                                                                                       & \includegraphics[width=0.16\textwidth]{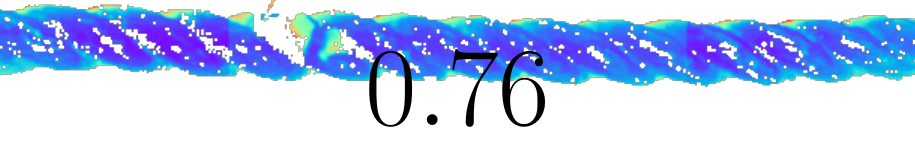}
                                                                                       & \includegraphics[width=0.16\textwidth]{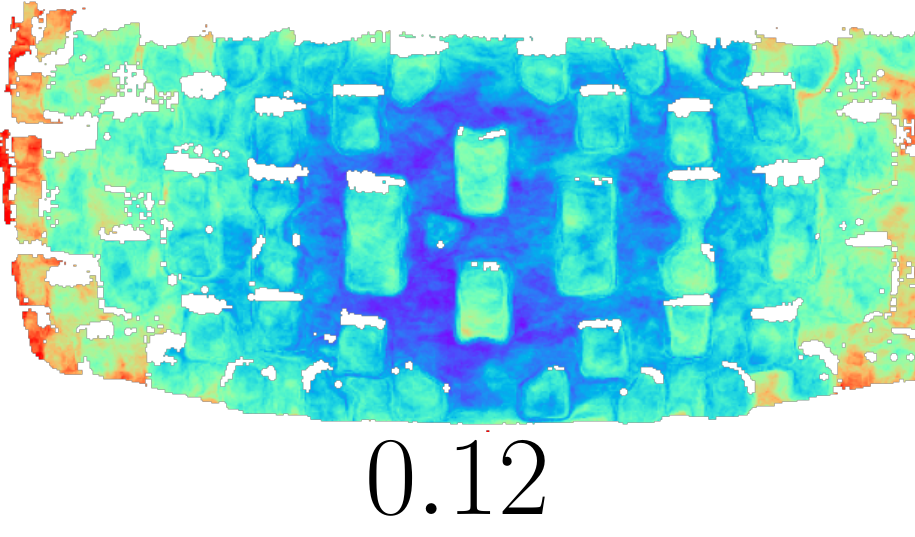} \\
                                      & \raisebox{0.06\textwidth}{$\frac{\delta}{20}$} & \includegraphics[width=0.1\textwidth]{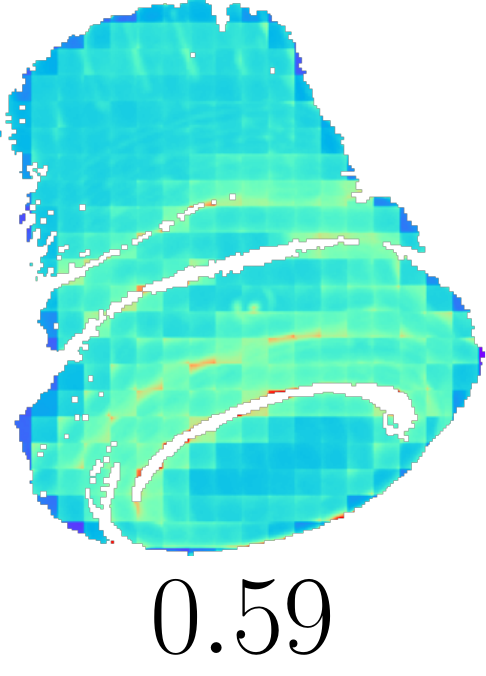}
                                                                                       & \includegraphics[width=0.1\textwidth]{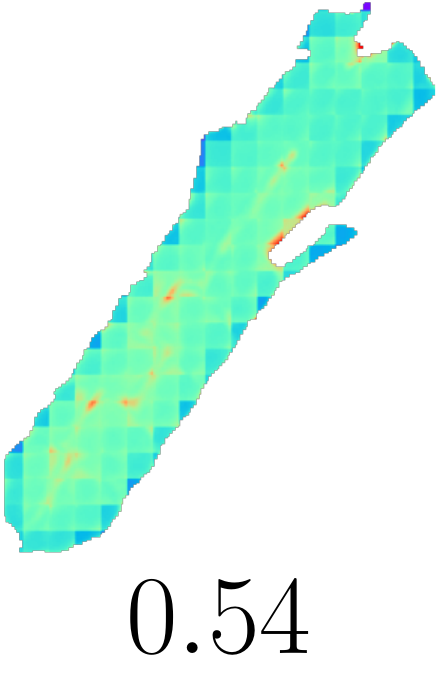}
                                                                                       & \includegraphics[width=0.1\textwidth]{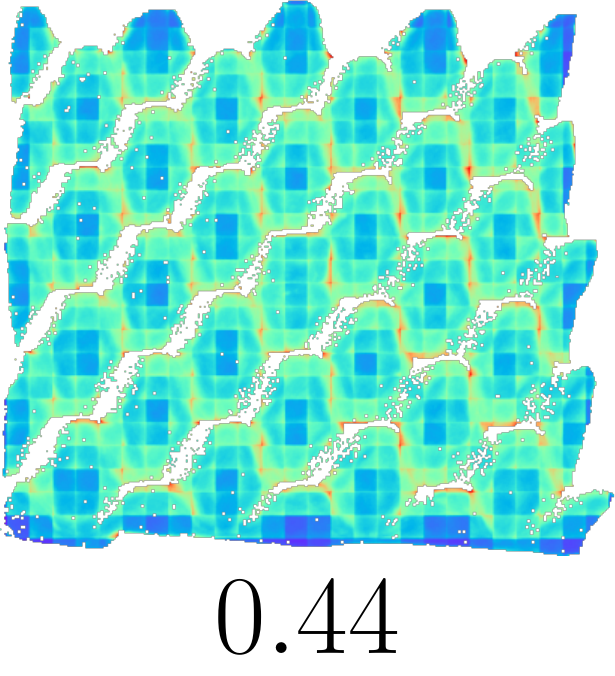}
                                                                                       & \includegraphics[width=0.16\textwidth]{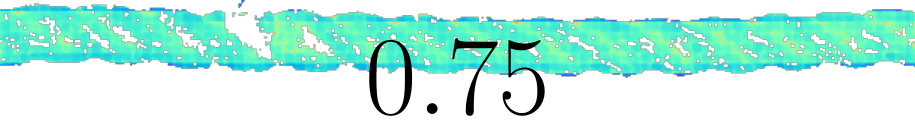}
                                                                                       & \includegraphics[width=0.16\textwidth]{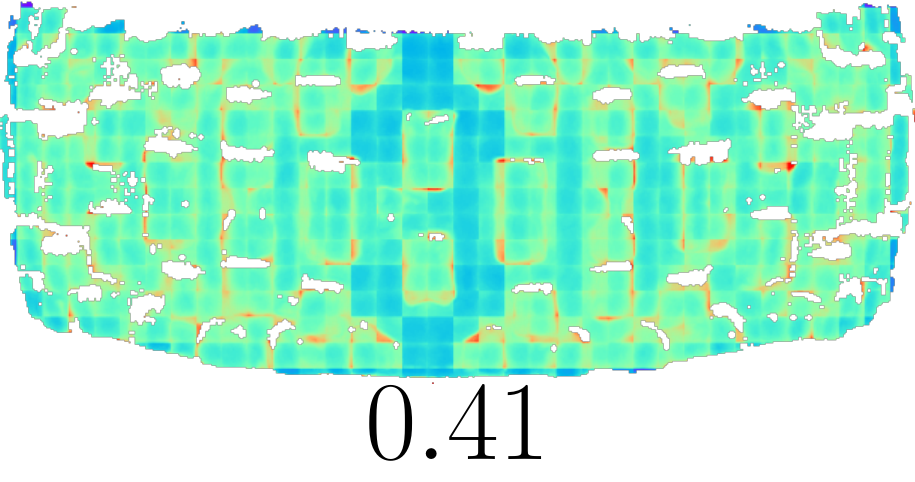} \\
                        \midrule
                        \multirow{2}{*}{Sphere}
                                      & \raisebox{0.06\textwidth}{$\delta$}            & \includegraphics[width=0.1\textwidth]{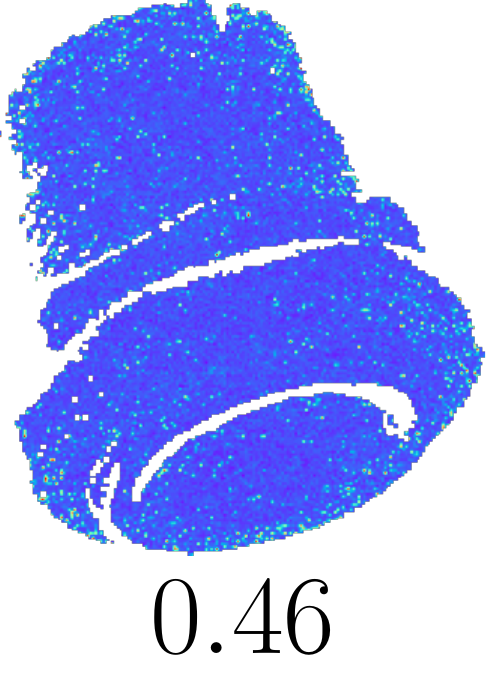}
                                                                                       & \includegraphics[width=0.1\textwidth]{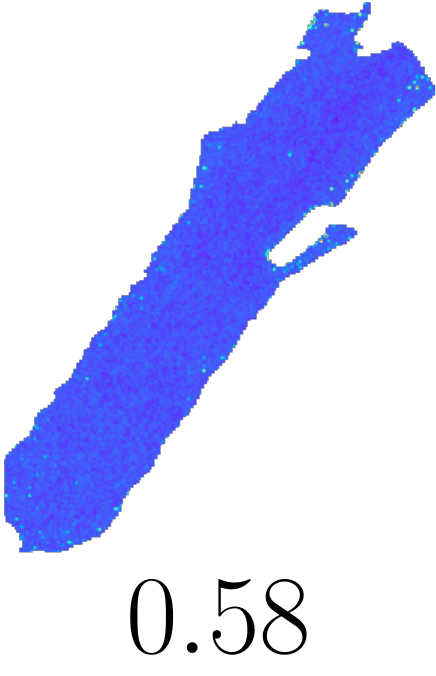}
                                                                                       & \includegraphics[width=0.1\textwidth]{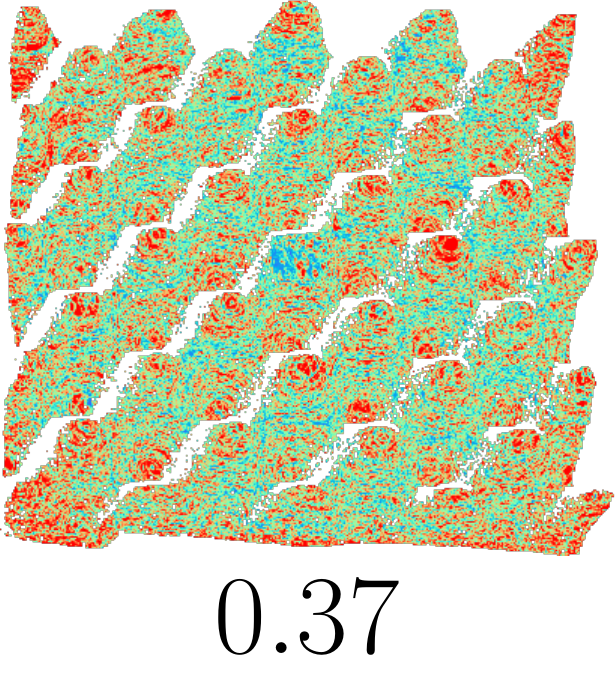}
                                                                                       & \includegraphics[width=0.16\textwidth]{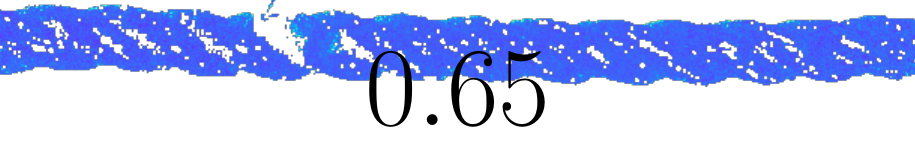}
                                                                                       & \includegraphics[width=0.16\textwidth]{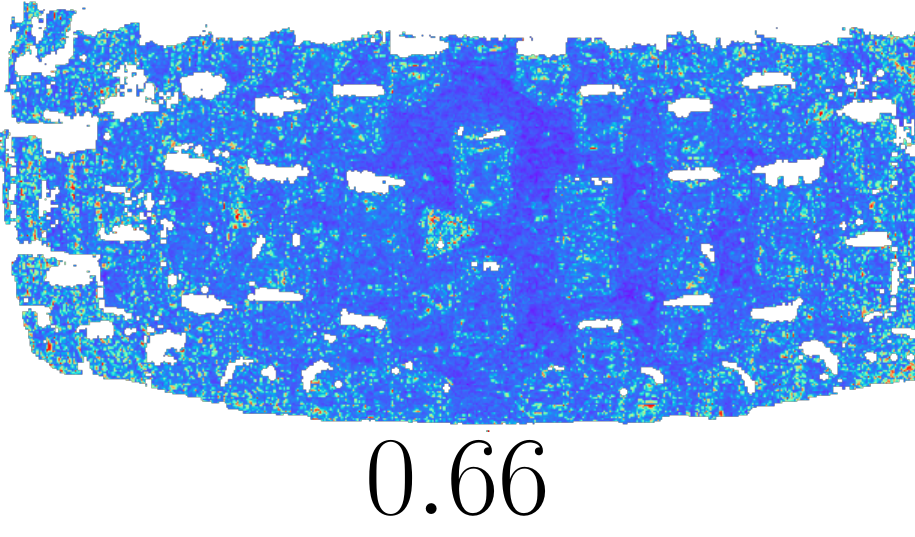} \\
                                      & \raisebox{0.06\textwidth}{$\frac{\delta}{20}$} & \includegraphics[width=0.1\textwidth]{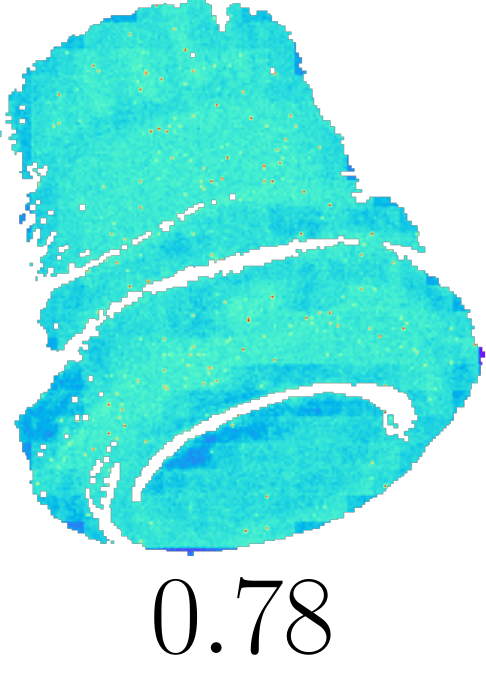}
                                                                                       & \includegraphics[width=0.1\textwidth]{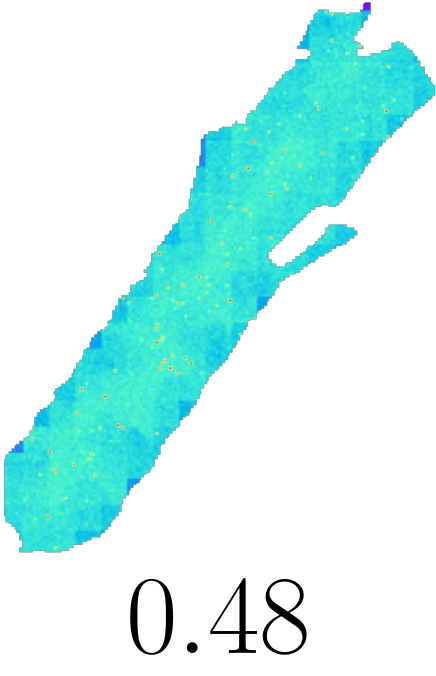}
                                                                                       & \includegraphics[width=0.1\textwidth]{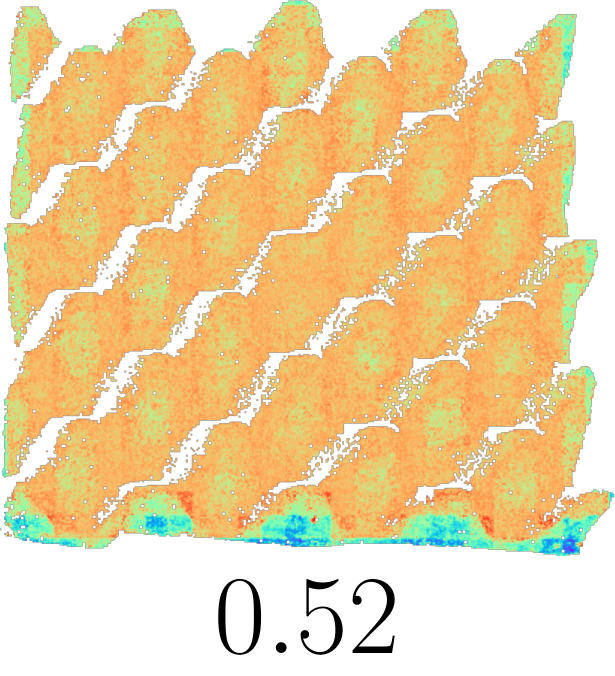}
                                                                                       & \includegraphics[width=0.16\textwidth]{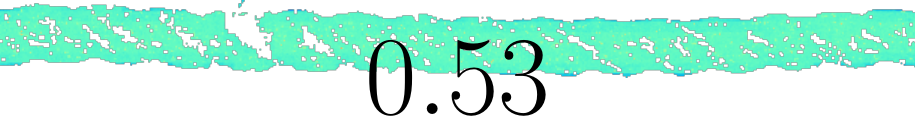}
                                                                                       & \includegraphics[width=0.16\textwidth]{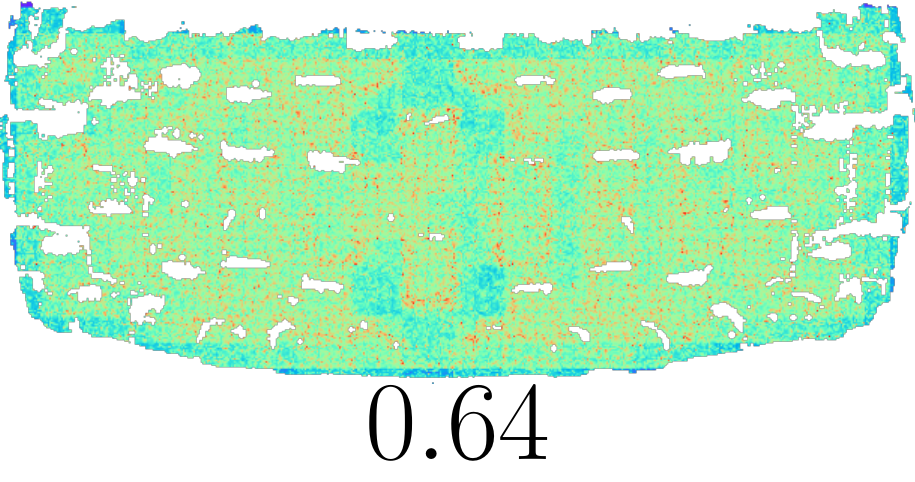} \\
                        \midrule
                        \multirow{2}{*}{Quadric}
                                      & \raisebox{0.06\textwidth}{$\delta$}            & \includegraphics[width=0.1\textwidth]{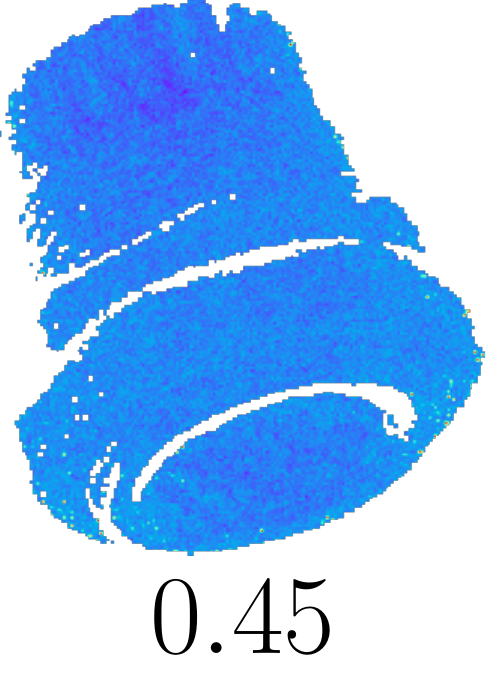}
                                                                                       & \includegraphics[width=0.1\textwidth]{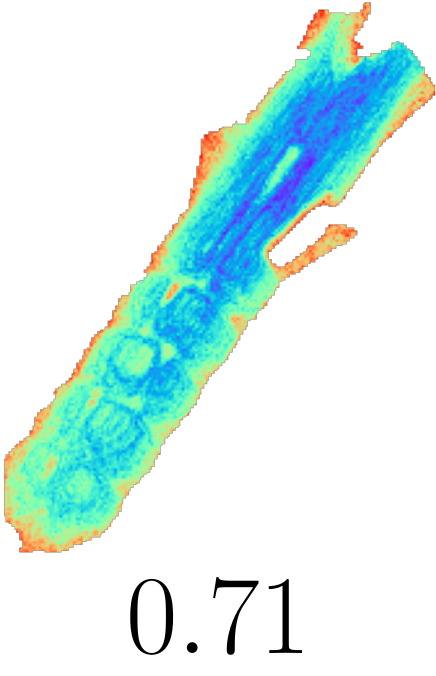}
                                                                                       & \includegraphics[width=0.1\textwidth]{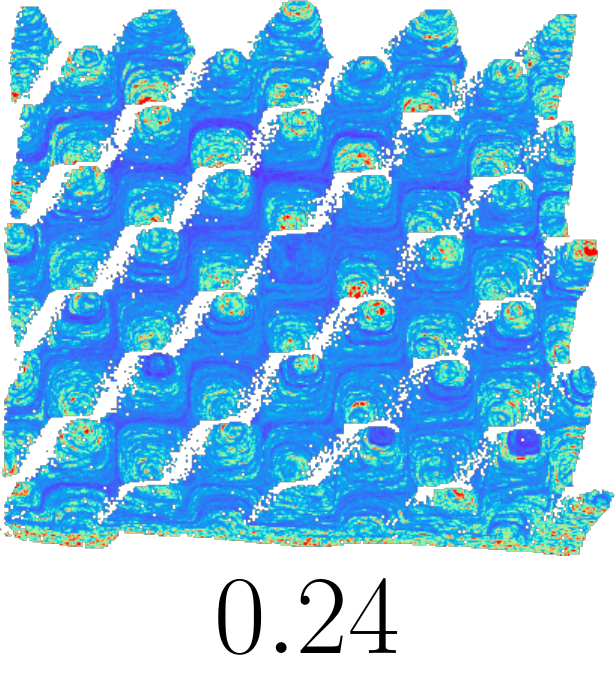}
                                                                                       & \includegraphics[width=0.16\textwidth]{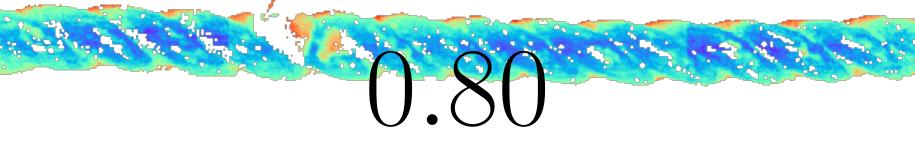}
                                                                                       & \includegraphics[width=0.16\textwidth]{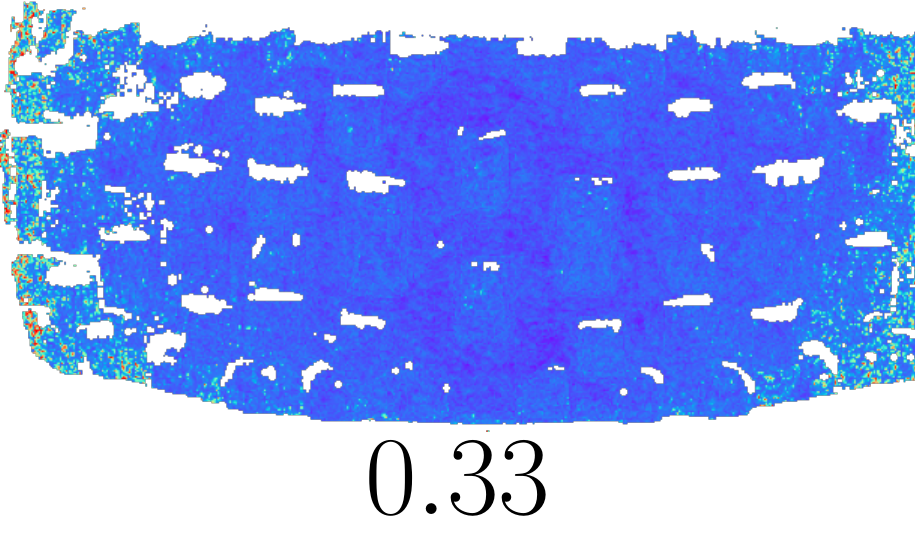} \\
                                      & \raisebox{0.06\textwidth}{$\frac{\delta}{20}$} & \includegraphics[width=0.1\textwidth]{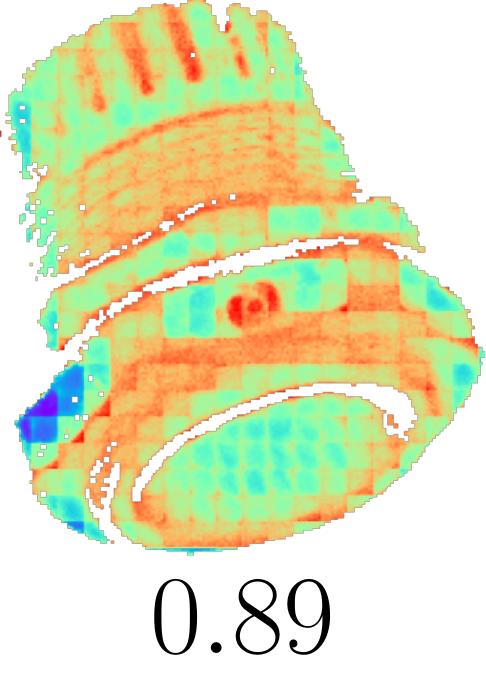}
                                                                                       & \includegraphics[width=0.1\textwidth]{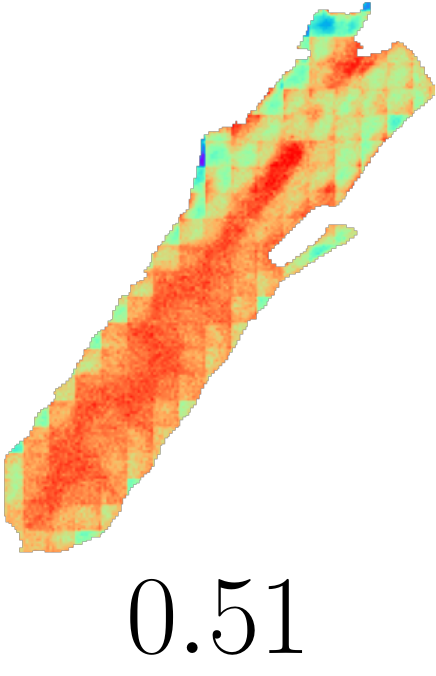}
                                                                                       & \includegraphics[width=0.1\textwidth]{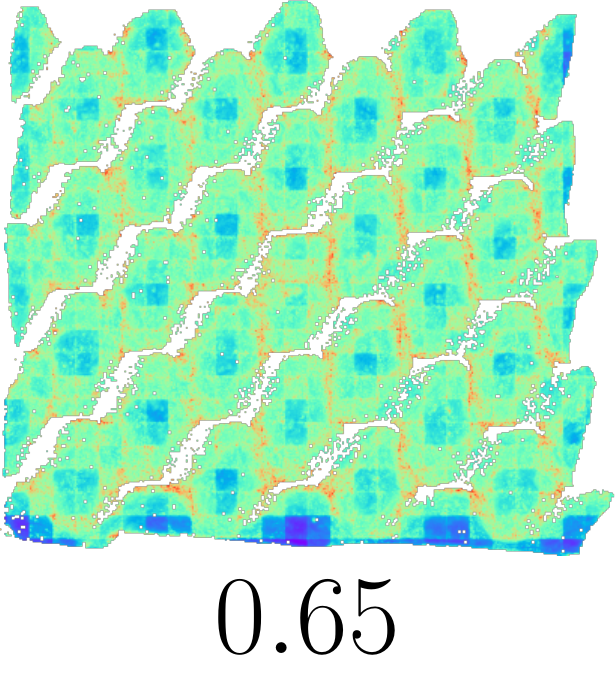}
                                                                                       & \includegraphics[width=0.16\textwidth]{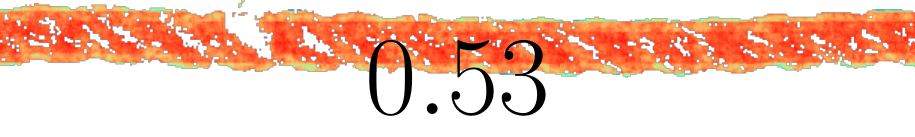}
                                                                                       & \includegraphics[width=0.16\textwidth]{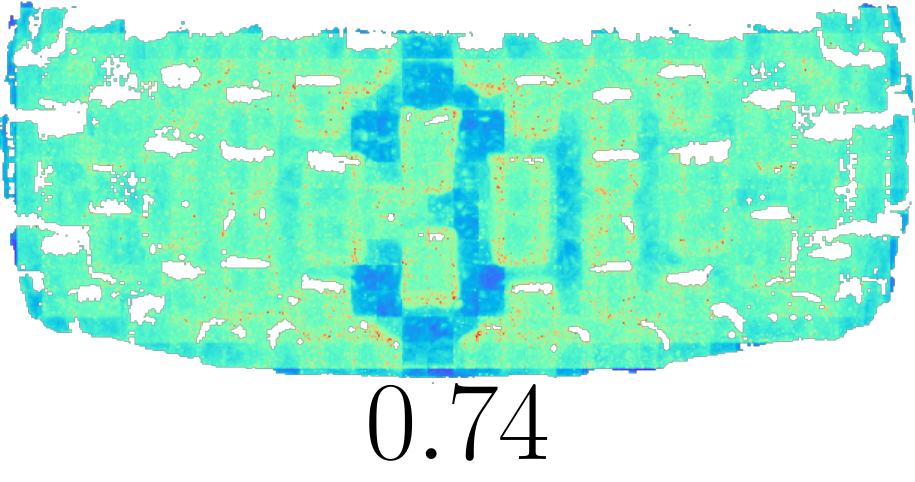} \\
                        \midrule
                        \midrule
                        \raisebox{0.05\textwidth}{\parbox{1cm}{\setlength{\baselineskip}{0.4cm}Ground\\truth}}
                                      &                                                & \includegraphics[width=0.1\textwidth]{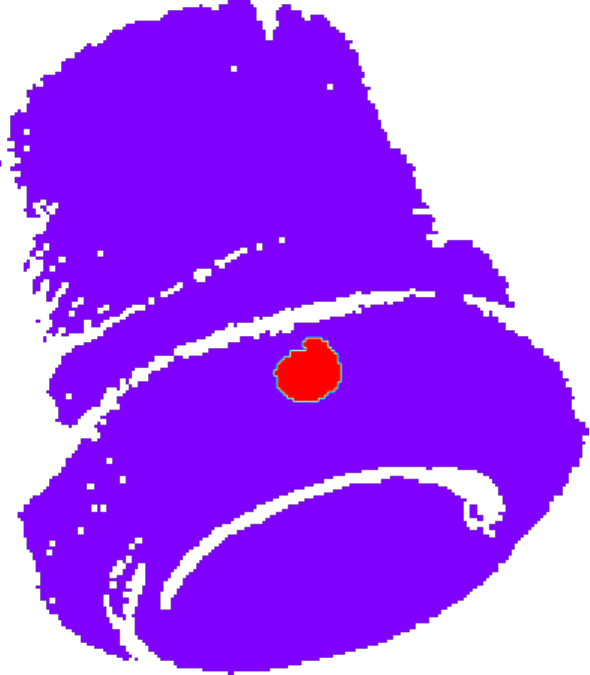}
                                                                                       & \includegraphics[width=0.1\textwidth]{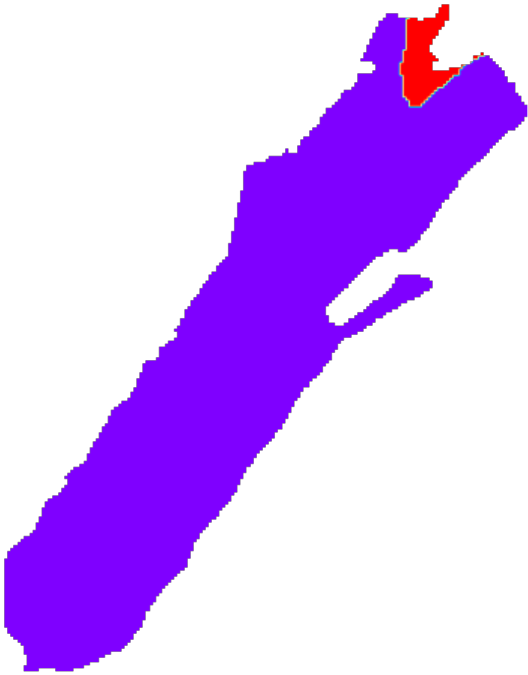}
                                                                                       & \includegraphics[width=0.1\textwidth]{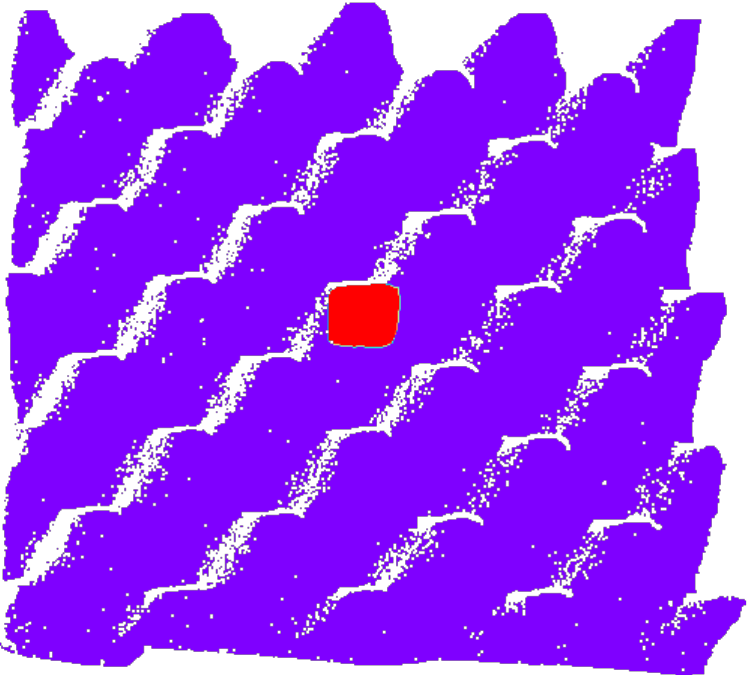}
                                                                                       & \includegraphics[width=0.16\textwidth]{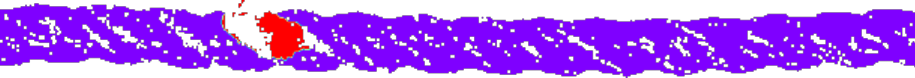}
                                                                                       & \includegraphics[width=0.16\textwidth]{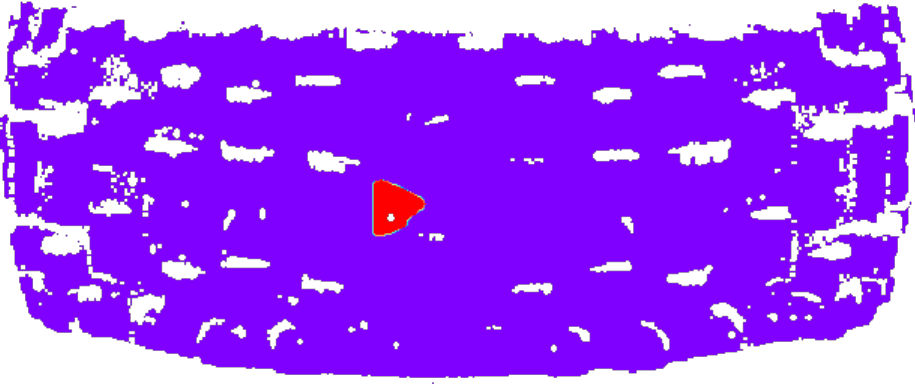} \\
                    \end{tabular}
                    }
                    \label{tab:sliding_pif_non-smooth_qualitative}
                \end{subtable}
                \hfill
                \begin{subtable}[b]{.275\textwidth}
                    \centering
                    \caption{}
                    \vspace{.9cm}
                    \begin{tabular}{l}
                        \includegraphics[width=\textwidth]{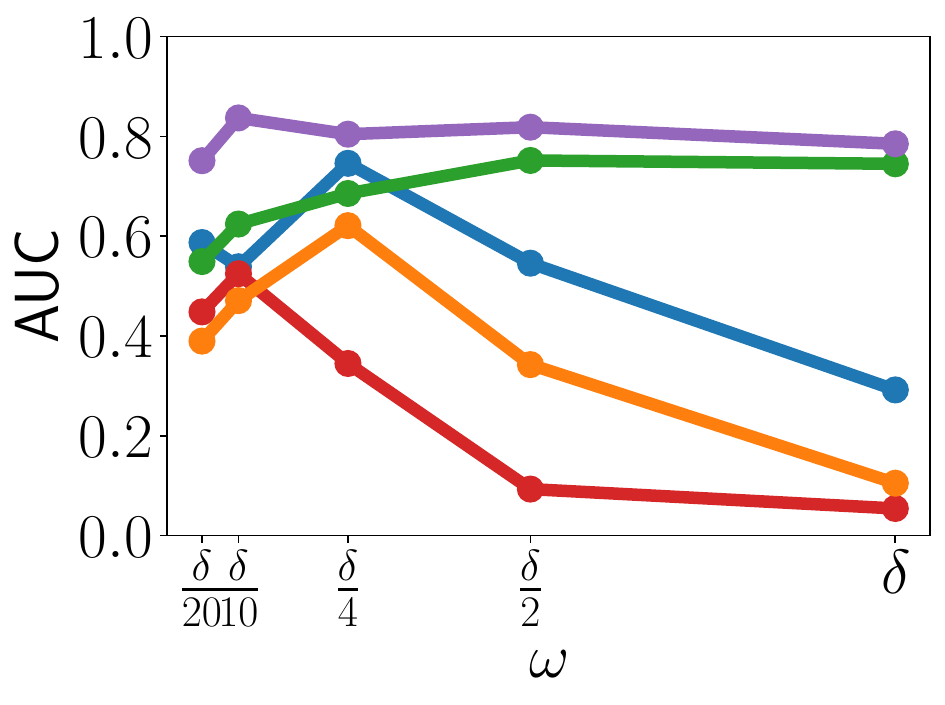}     \\
                        \vspace{.5cm}
                        \includegraphics[width=\textwidth]{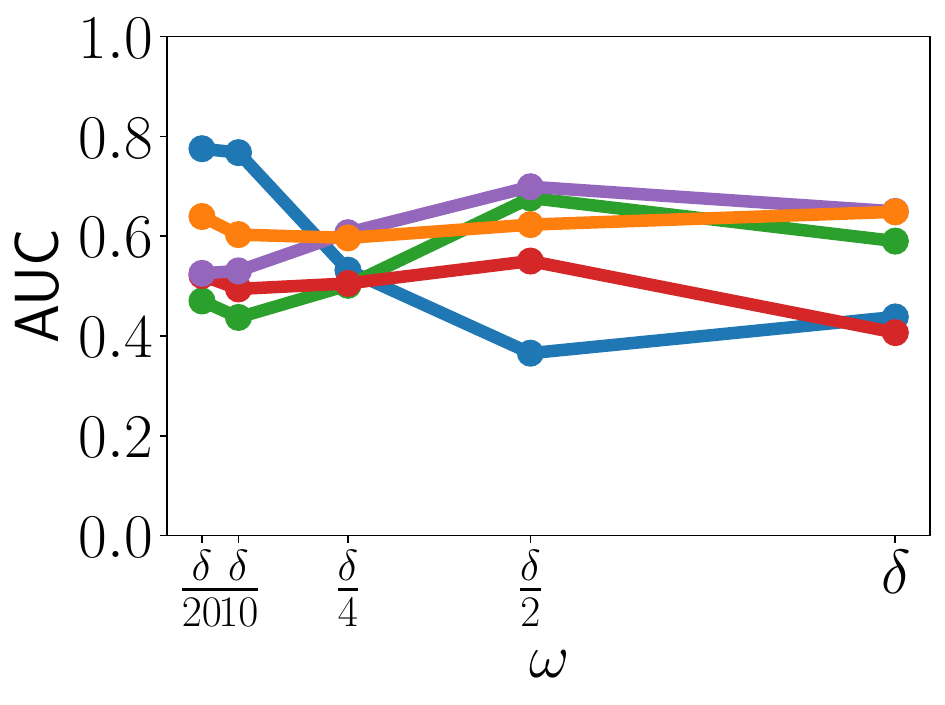}   \\
                        \vspace{.1cm}
                        \includegraphics[width=\textwidth]{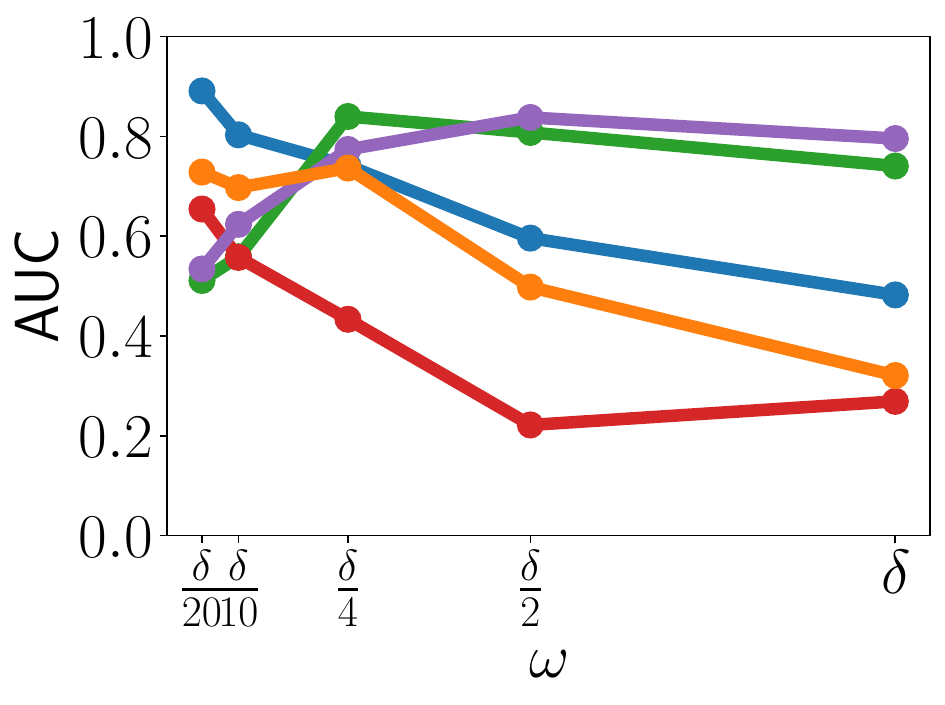} \\
                        \vspace{.5cm}
                        \parbox{\textwidth}{\vspace{-0.3cm} \includegraphics[width=1.2\textwidth]{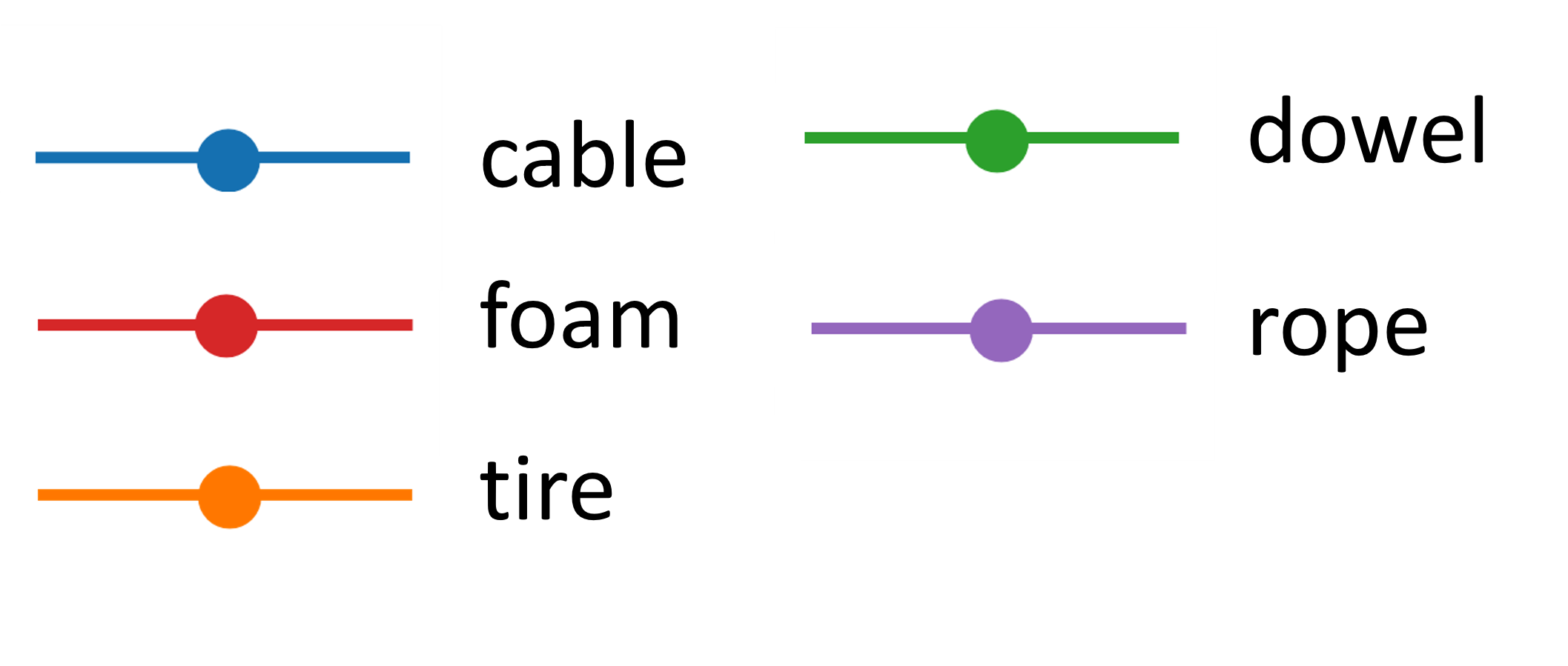}}                         \\
                        \vspace{-1.3cm}
                    \end{tabular}
                    \label{tab:sliding_pif_non-smooth_quantitative}
                \end{subtable}
                }
                \label{tab:sliding_pif_non-smooth}
            \end{table}
                
            As expected, \spif performs generally better with smaller window sizes $\omega$ (\cref{tab:sliding_pif_smooth_quantitative}), as local models tend to be more accurate when fitted to smaller regions, while large $\omega$ do not satisfy the locality principle.
            An exception is the cookie object, which contains a relatively large anomalous area. In fact, \spif performs well when genuine data are the majority in $W$, thus the best $\omega$ for cookie is the second or third-smallest value for planes and quadrics. Spheres, on the contrary, perform best when $\omega = \delta$, since they describe accurately the entire cookie, but worst for small windows as they tend to be biased towards the center of $W$, struggling to describe borders.
            Potato has a more ellipsoidal shape, and spheres struggle to describe it for large $\omega$. With medium-sized windows, spheres perform best, showing that locally they best capture potato shape. The higher flexibility of quadrics leads to high variance in the models fitted, resulting in poor performance for medium-sized $\omega$.
            Carrots are elongated and difficult to approximate with spheres, but they are well described by quadrics and planes, since the local curvature is not very pronounced.
            Bagels and peaches are difficult for all the model families when $\omega$ is large due to their toroidal shape. However, as $\omega$ decreases, all families perform better, since the toroidal shape no longer prevails in a small neighborhood.

            We further evaluated \spif on mechanical objects with non-smooth surfaces—namely \emph{cable gland}, \emph{dowel}, \emph{foam}, \emph{rope}, and \emph{tire}—to explore its limitations (\cref{tab:sliding_pif_non-smooth_quantitative}). As expected, performance generally declines in this scenario, due to the reduced compatibility with local structural priors.
            Among these objects, \emph{rope} and \emph{dowel} achieve relatively better performance, as their elongated, near-planar shapes resemble those of smoother objects like \emph{carrot}, and are therefore still amenable to local approximation with simple geometric models (e.g., planes or quadrics). In contrast, \emph{foam} represents a clear worst-case, \emph{Tire} follows closely in difficulty, while \emph{cable gland} lies in between. While the performance remains strong with smooth objects, it declines on highly irregular or non-smooth geometries (\emph{e.g.}, foam, tire), highlighting a key limitation when the model family $\mathcal{F}$ poorly matches the underlying data. Nonetheless, these failure modes are interpretable and reflect the model’s transparency. The reasons behind these performance drops can be systematically attributed to three fundamental challenges or limitations of our solution: \emph{i)} \emph{Violation of local structural assumptions}: in \emph{Foam} local structural regularity is systematically violated. Its highly irregular and non-smooth surfaces cannot be meaningfully captured by geometric primitives, resulting in consistently poor performance across all model families and window sizes. \emph{ii)} \emph{Mixed surface characteristics}: \emph{Tire} demonstrates the challenges arising when data exhibit spatially inconsistent local properties. While some regions are locally smooth, the presence of treads and sharp geometric transitions results in unstable model fitting, as individual local windows contain fundamentally incompatible structural characteristics that cannot be reconciled by a single geometric model. \emph{iii)} \emph{Partial structural compatibility}: \emph{Cable gland} exemplifies borderline scenarios for \pif. The combination of both smooth and irregular regions makes genuine data only partially approximable by local models, leading to unpredictable performance due to the spatial heterogeneity of structural properties, where some windows yield reliable models while others fail completely.
            Regarding model families, results reflect the inherent trade-offs between flexibility and stability. Planes perform well on nearly flat or elongated objects (\emph{e.g.}, \emph{dowel}, \emph{rope}, \emph{carrot}), due to their simplicity and low variance.
            Quadrics are more expressive and adaptable, but this flexibility results in higher variance across datasets. Spheres tend to perform consistently poor in this non-smooth setting, as their regular geometry rarely aligns with the actual surface structure, despite showing better behavior on smoother objects.
            These experiments demonstrate the robustness of our approach in complex real-world scenarios, where a global structural prior describing genuine data is unavailable or difficult to define. \spif proves effective even when structural assumptions hold only locally, achieving strong performance leveraging a prior based on an appropriate model family $\mathcal{F}$ on windows having size $\omega$. In contrast to global approaches (like \pif), \spif is more practical to analyze large-scale and geometrically heterogeneous datasets. Furthermore, the combination of \spif and \rzhiforest significantly improves runtime and memory efficiency, confirming the scalability of our framework for structure-based anomaly detection.

    \section{Limitations}
        \label{sec:limitations}
        
        The first limitation of our framework is its reliance on structural priors. The method is effective when some prior knowledge about the underlying structures characterizing genuine data is available, and can be expressed through a model family $\mathcal{F}$. While our results on \spif show that a local or coarse prior -- such as fitting planes to approximate smooth surfaces -- can in some cases be sufficient, some minimal understanding of the data's structure remains necessary. For example, it is crucial to know the intrinsic dimensionality of the structures: approximating a 2D surface with a 1D model such as a line would lead to unstable embeddings and poor separation, as the prior would fail to capture the relevant regularities and might instead be overly sensitive to noise. 
        
        The second limitation lies in the computational cost of the preference embedding. While the cost of the anomaly scoring stage (\pisolation) is mitigated through efficient mechanisms such as LSH partitioning and the sliding window strategy in \spif, the construction of the preference space remains computationally intensive, both in terms of runtime and memory. Although \spif reduces these demands by restricting computations to local regions, the process is still non-trivial when scaling to very large datasets. That said, this bottleneck could be addressed in future work through parallelized or GPU-accelerated implementations of the embedding step.

    \section{Conclusion}
        \label{sec:conclusion}
        
        Detecting anomalies in structured data is challenging, especially when genuine data lie on low-dimensional manifolds embedded in higher-dimensional ambient spaces. In such settings, traditional methods often fail, as they rely on ambient-space densities, which may not reflect structural inconsistency.
        
        To address this, we introduce \pif, which shifts the perspective from monitoring ambient-space density to structural consistency. Instead of  explicitly recovering the underlying models and their inliers --which is often more difficult than detecting anomalies -- we construct a preference embedding: each data point is mapped into a high-dimensional space that reflects how well it agrees with a pool of candidate models. In this \emph{preference space}, anomalies naturally appear as isolated points, lacking consensus with the genuine structures.
        
        The modular nature of \pif allows adapting the method to different scenarios:
        \begin{itemize}[noitemsep]
        \item When a global structure is assumed and arbitrary distances over preferences are needed, \viforest offers a general and accurate solution by leveraging Voronoi tessellations. 
        \item  When scalability is crucial, \rzhiforest replaces costly distance computations with efficient locality-sensitive hashing. It retains the benefits of structure-based detection while enabling application to large datasets.
        \item  When a locality prior is available, \spif allow to perform anomaly detection even  even when global structure is hard to define.
        \end{itemize}

        As future work, promising directions include integrating non-parametric or learned model families into the embedding process, and designing new distances or hashing schemes tailored to specific structural priors. 

    \section*{Acknowledgements}
        \label{sec:acknowledgments}
        This work is supported by GEOPRIDE under ID 2022245ZYB and CUP D53D23008370001 (PRIN 2022 M4.C2.1.1 Investment).

    \appendix

    \section{\rzhash is a \lsh for Ruzicka distance}
        \label{apx:rzhash}
        
        We present the proof of~\cref{thm:ruzhash_short}, introduced in~\cref{subsubsec:rzhash}, stating that \rzhash is a Locality Sensitive Hashing (LSH) for the Ruzicka distance. The proof builds on~\cref{lem:minhash}, a well-known result stating that \mhash is an LSH for the Jaccard distance. Additionally, \cref{fig:correlation} empirically illustrates the correlation between \rzhash and Ruzicka.
        
        \begin{lemma}[\mhash]
            \label{lem:minhash}
            Given $\vect{p}, \vect{q} \in \{0, 1\}^m$, let $h^{\vect{\pi}}_{min} : \{0, 1\}^m \rightarrow \{1, \dots, m\}$ be the \mhash mapping, where $\vect{\pi} = [\pi_1, \dots, \pi_m]$ is a random permutation of the $m$ dimensions $\{1, \dots, m\}$. It is known~\cite{BroderCharikarAl00} that
            
            \begin{equation*}
                \begin{split}
                    Pr[h^{\vect{\pi}}_{min}(\vect{p}) = h^{\vect{\pi}}_{min}(\vect{q})] &= \frac{\sum_{\vect{\pi}} \mathds{1}(h^{\vect{\pi}}_{min}(\vect{p}) = h^{\vect{\pi}}_{min}(\vect{q}))}{m!} =\\
                                                                                        &= \frac{\sum_{i=1}^{m} (p_i \wedge q_i)}{\sum_{i=1}^{m} (p_i \vee q_i)} = \frac{\sum_{i=1}^{m} \min\{p_i, q_i\}}{\sum_{i=1}^{m} \max\{p_i, q_i\}} = 1 - d_J(\vect{p},\vect{q}),
                \end{split}
            \end{equation*}
            where $\mathds{1}$ is the indicator function, $m!$ is the total number of permutations, and $\sum_{\vect{\pi}}$ sums over all the possible realizations of permutation $\vect{\pi}$.
        \end{lemma}

        \setcounter{theorem}{0}
        \begin{theorem}[\rzhash]
            \label{thm:ruzhash}
            Given $\vect{p}, \vect{q} \in [0, 1]^m$, let $\vect{\tau} = [\tau_1, \dots, \tau_m]$ be a vector of thresholds where $\tau_i \sim \mathcal{U}_{[0, 1)}$, let $\mathds{1}(\vect{p} > \vect{\tau}) = [\mathds{1}(p_1 > \tau_1), \dots, \mathds{1}(p_m > \tau_m)] \in \{0, 1\}^m$ be the binarization function, and let $h^{\vect{\pi}}_{min} : \{0, 1\}^m \rightarrow \{1, \dots, m\}$ be the \mhash mapping (\cref{lem:minhash}). Let \rzhash be the function $h^{\vect{\pi}, \vect{\tau}}_{ruz} = h^{\vect{\pi}}_{min} \circ \mathds{1}(\vect{p} > \vect{\tau}) :  [0, 1]^m \rightarrow \{1, \dots, m\}$. Then,
            \begin{equation}
                Pr[h^{\vect{\pi}, \vect{\tau}}_{ruz}(\vect{p}) = h^{\vect{\pi}, \vect{\tau}}_{ruz}(\vect{q})] = \frac{\sum_{i=1}^m \min\{p_i, q_i\}}{\sum_{i=1}^m \max\{p_i, q_i\}} = 1 - d_R(\vect{p}, \vect{q}).
                \label{eq:rzhash}
            \end{equation}
        \end{theorem}
        \begin{proof}
            Relying on the definition of \rzhash, it holds that
            \begin{equation*}
                \begin{split}
                    &P[h^{\vect{\pi}}_{min}(\mathds{1}(\vect{p} > \vect{\tau})) = h^{\vect{\pi}}_{min}(\mathds{1}(\vect{q} > \vect{\tau}))] = \frac{\int_{\vect{\tau}} \sum_{\vect{\pi}} \mathds{1}(h^{\vect{\pi}}_{min}(\mathds{1}(\vect{p} > \vect{\tau})) = h^{\vect{\pi}}_{min}(\mathds{1}(\vect{q} > \vect{\tau}))) d\vect{\tau}}{m!} =\\
                    &= \frac{\int_{\vect{\tau}} \sum_{i=1}^{m} (\mathds{1}(p_i > \tau_i) \wedge \mathds{1}(q_i > \tau_i)) d\vect{\tau}}{\int_{\vect{\tau}} \sum_{i=1}^{m} (\mathds{1}(p_i > \tau_i) \vee \mathds{1}(q_i > \tau_i)) d\vect{\tau}} = \frac{\int_{\vect{\tau}} \sum_{i=1}^m \min\{\mathds{1}(p_i > \tau_i), \mathds{1}(q_i > \tau_i)\} d\vect{\tau}}{\int_{\vect{\tau}} \sum_{i=1}^m \max\{\mathds{1}(p_i > \tau_i), \mathds{1}(q_i > \tau_i)\} d\vect{\tau}},
                \end{split}
            \end{equation*}
            where $\int_{\vect{\tau}}$ integrates over all the possible realizations of $\vect{\tau}$, and the rest follows from~\cref{lem:minhash}.
            Since each $\tau_i \sim \mathcal{U}_{[0, 1)}$ is an \textit{independent} realization, then
            \begin{equation*}
                \text{\hspace{-0.75cm}}
                \frac{\int_{\vect{\tau}} \sum_{i=1}^m \min\{\mathds{1}(p_i > \tau_i), \mathds{1}(q_i > \tau_i)\} d\vect{\tau}}{\int_{\vect{\tau}} \sum_{i=1}^m \max\{\mathds{1}(p_i > \tau_i), \mathds{1}(q_i > \tau_i)\} d\vect{\tau}} = \frac{\sum_{i=1}^m \int_0^1 \min\{\mathds{1}(p_i > \tau_i), \mathds{1}(q_i > \tau_i)\} d\tau_i}{\sum_{i=1}^m \int_0^1 \max\{\mathds{1}(p_i > \tau_i), \mathds{1}(q_i > \tau_i)\} d\tau_i}.
            \end{equation*}
            We have that
            \begin{equation*}
                \min\{\mathds{1}(p_i > \tau_i), \mathds{1}(q_i > \tau_i)\} =
                \begin{cases}
                    1 & \text{if $\min\{p_i, q_i\} > \tau_i$}\\
                    0 & \text{otherwise},
                \end{cases}
            \end{equation*}
            and analogously for $\max\{\mathds{1}(p_i > \tau_i), \mathds{1}(q_i > \tau_i)\}$.
            Therefore,
            \begin{equation*}
                \text{\hspace{-1cm}}
                \begin{split}
                    &\frac{\sum_{i=1}^m \int_0^1 \min\{\mathds{1}(p_i > \tau_i), \mathds{1}(q_i > \tau_i)\} d\tau_i}{\sum_{i=1}^m \int_0^1 \max\{\mathds{1}(p_i > \tau_i), \mathds{1}(q_i > \tau_i)\} d\tau_i} = \frac{\sum_{i=1}^m (\int_0^{\min\{p_i, q_i\}} 1 \,d\tau_i + \int_{\min\{p_i, q_i\}}^1 0 \,d\tau_i)}{\sum_{i=1}^m (\int_0^{\max\{p_i, q_i\}} 1 \,d\tau_i + \int_{\max\{p_i, q_i\}}^1 0 \,d\tau_i)} =\\
                    &= \frac{\sum_{i=1}^m \int_0^{\min\{p_i, q_i\}} 1 \,d\tau_i}{\sum_{i=1}^m \int_0^{\max\{p_i, q_i\}} 1 \,d\tau_i} = \frac{\sum_{i=1}^m \min\{p_i, q_i\}}{\sum_{i=1}^m \max\{p_i, q_i\}},
                \end{split}
            \end{equation*}
            and this proves~\eqref{eq:rzhash}.
        \end{proof}
        
        \begin{figure}[tb]
            \centering
            \begin{subfigure}[t]{.25\linewidth}
                \centering
                \includegraphics[width=\linewidth]{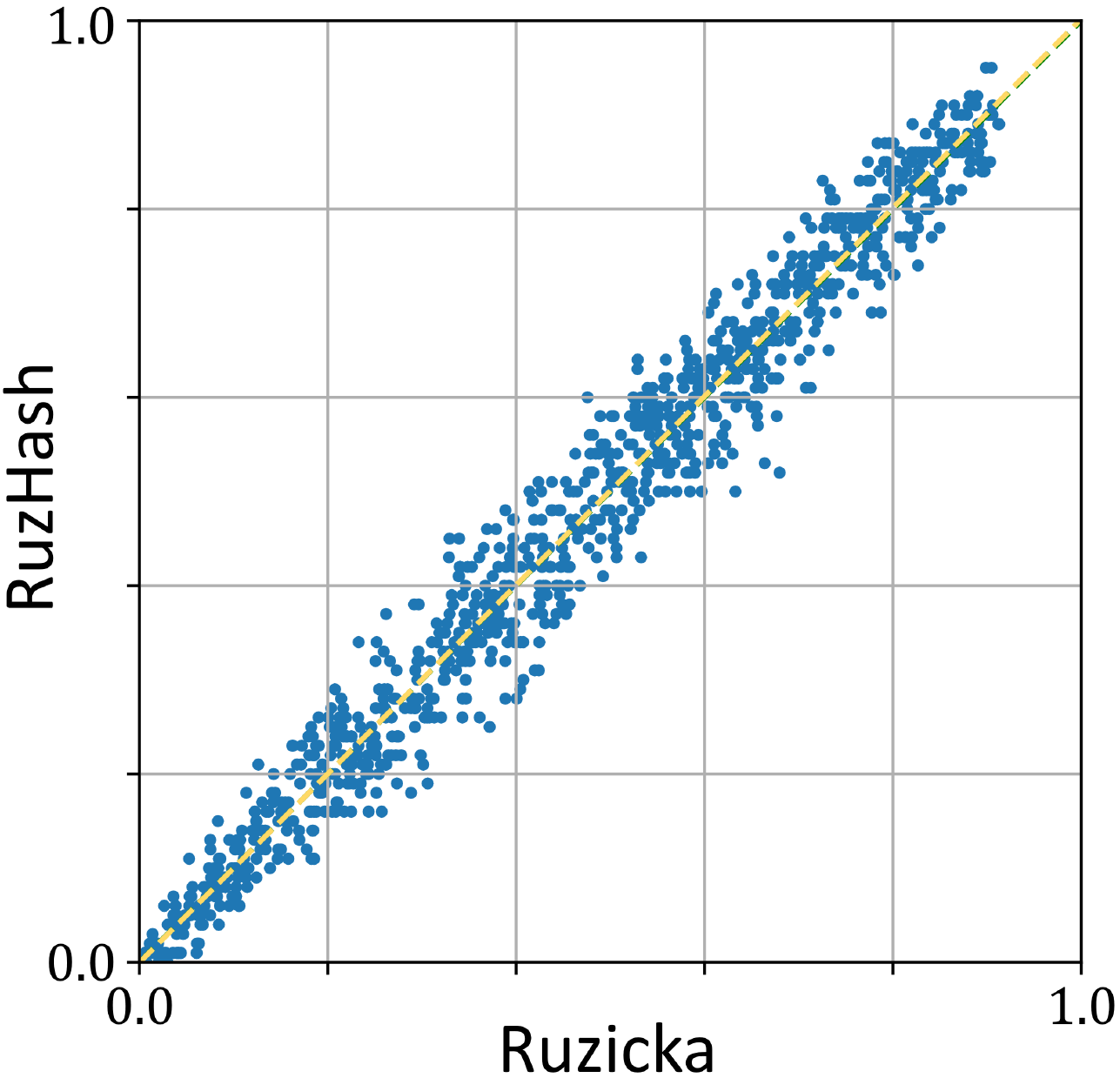}
                \caption{}
                \label{fig:ruzickacorr}
            \end{subfigure}
            \hspace{2cm}
            \begin{subfigure}[t]{.25\linewidth}
                \centering
                \includegraphics[width=\linewidth]{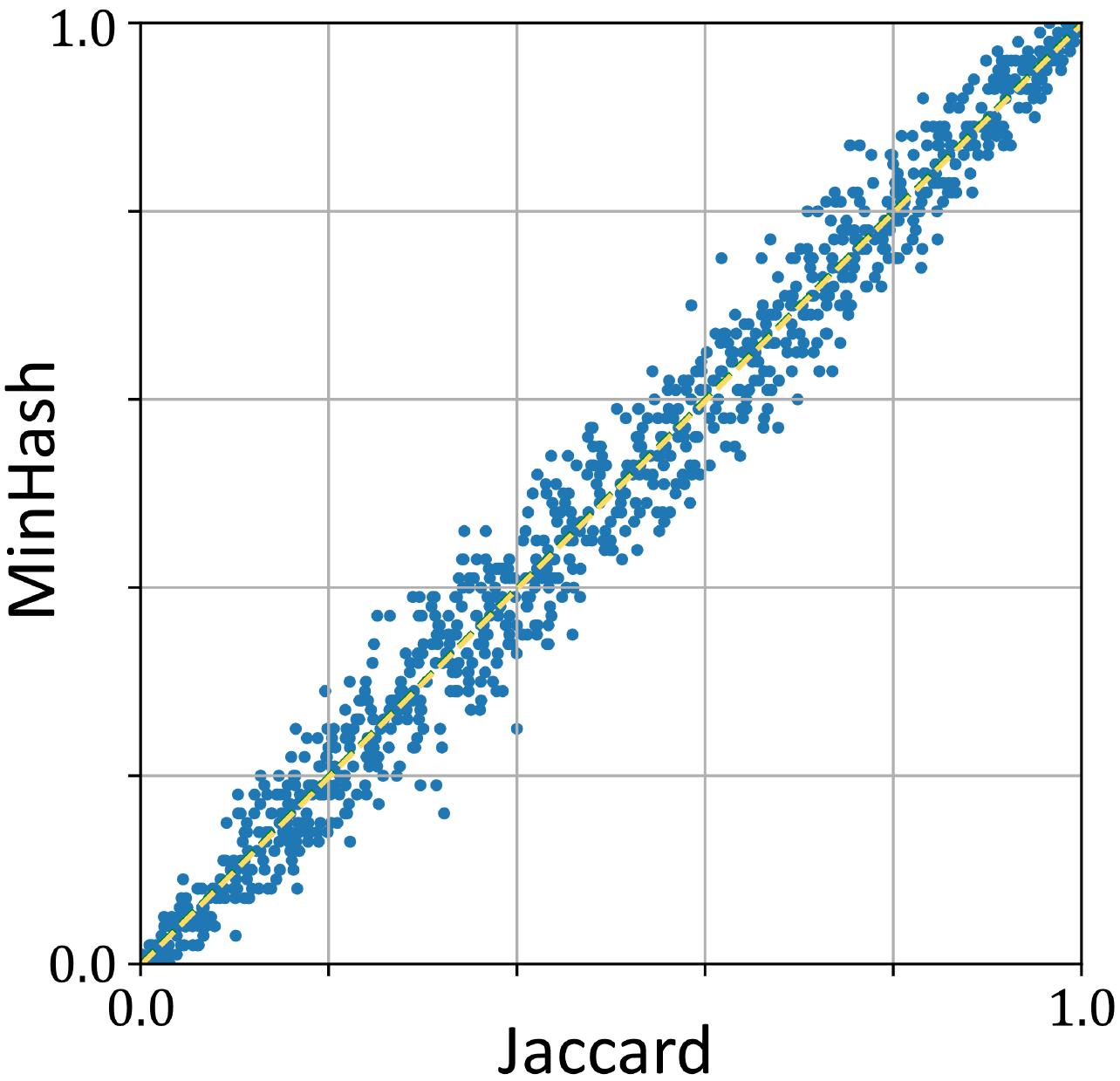}
                \caption{}
                \label{fig:minhashcorr}
            \end{subfigure}
            \caption{Correlation between Ruzicka and \rzhash estimated from $n = 1000$ random couples of points $\vect{p}, \vect{q} \in [0, 1]^m$, where $m = 10000$ and \rzhash is executed $k = 100$ times for each couple, compared to (b) the correlation between Jaccard and \mhash.}
            \label{fig:correlation}
        \end{figure}
        
        \cref{fig:correlation} shows the correlation between $1 - d_R(\vect{p}, \vect{q})$ and \rzhash for $n = 1000$ random couples of points $\vect{p}, \vect{q} \in [0, 1]^m$, when $m = 10000$ and \rzhash is executed $k = 100$ times for each couple, compared to the correlation between $1 - d_J(\vect{p}, \vect{q})$ and \mhash in the same settings.

    \section{\rzhash linearly correlates with Ruzicka distance}
        \label{apx:variable_split}
        
        We prove that, when the \rzhash mapping is modified as described in~\cref{subsubsec:rzhitree}, it maintains a correlation with the Ruzicka distance, with the slope being proportional to the new branching factor $b$. In~\cref{fig:ruzicka_variable}, we empirically illustrate the linear correlation for different branching factors.
        
        \begin{theorem}[$d_R(\vect{p}, \vect{q})$ linearly correlates with $b$]
            \label{thm:correlation}
            Given $\vect{p}, \vect{q} \in [0, 1]^m$, let $h^{\vect{\pi}, \vect{\tau}}_{ruz}(\vect{p}), h^{\vect{\pi}, \vect{\tau}}_{ruz}(\vect{q}) \in \{1, \dots, m\}$ their maps computed by \rzhash. Let $\vect{\beta} = [\beta_1, \dots, \beta_m]$ be a vector in $\{1, \dots, b\}^m$, where each $\beta_i \sim \mathcal{U}_{\{1, \dots, b\}}$.
            Then,
            \begin{equation}
                \label{eq:correlation}
                    P[\beta_{h^{\vect{\pi}, \vect{\tau}}_{ruz}(\vect{p})} = \beta_{h^{\vect{\pi}, \vect{\tau}}_{ruz}(\vect{q})}] = 1 + \frac{1 - b}{b}d_R(\vect{p}, \vect{q}).
            \end{equation}
        \end{theorem}
        \begin{proof}
            We split $P[\beta_{h^{\vect{\pi}, \vect{\tau}}_{ruz}(\vect{p})} = \beta_{h^{\vect{\pi}, \vect{\tau}}_{ruz}(\vect{q})}]$ in two terms
            \begin{equation*}
                \text{\hspace{-2.5cm}}
                \begin{split}
                    P[\beta_{h^{\vect{\pi}, \vect{\tau}}_{ruz}(\vect{p})} = \beta_{h^{\vect{\pi}, \vect{\tau}}_{ruz}(\vect{q})}] &= P[\beta_{h^{\vect{\pi}, \vect{\tau}}_{ruz}(\vect{p})} = \beta_{h^{\vect{\pi}, \vect{\tau}}_{ruz}(\vect{q})} | h^{\vect{\pi}, \vect{\tau}}_{ruz}(\vect{p}) = h^{\vect{\pi}, \vect{\tau}}_{ruz}(\vect{q})] \, P[h^{\vect{\pi}, \vect{\tau}}_{ruz}(\vect{p}) = h^{\pi, \tau}_{ruz}(\vect{q})] +\\
                                                                                                                                 &+ P[\beta_{h^{\vect{\pi}, \vect{\tau}}_{ruz}(\vect{p})} = \beta_{h^{\vect{\pi}, \vect{\tau}}_{ruz}(\vect{q})} | h^{\vect{\pi}, \vect{\tau}}_{ruz}(\vect{p}) \neq h^{\vect{\pi}, \vect{\tau}}_{ruz}(\vect{q})] \, P[h^{\vect{\pi}, \vect{\tau}}_{ruz}(\vect{p}) \neq h^{\vect{\pi}, \vect{\tau}}_{ruz}(\vect{q})].
                \end{split}
            \end{equation*}
            We know from~\cref{thm:ruzhash} that $P[h^{\vect{\pi}, \vect{\tau}}_{ruz}(\vect{p}) = h^{\vect{\pi}, \vect{\tau}}_{ruz}(\vect{q})] = 1 - d_R(\vect{p}, \vect{q})$, thus
            \begin{equation}
                \begin{split}
                    \text{\hspace{-0.75cm}}
                    P[\beta_{h^{\vect{\pi}, \vect{\tau}}_{ruz}(\vect{p})} = \beta_{h^{\vect{\pi}, \vect{\tau}}_{ruz}(\vect{q})}] &= P[\beta_{h^{\vect{\pi}, \vect{\tau}}_{ruz}(\vect{p})} = \beta_{h^{\vect{\pi}, \vect{\tau}}_{ruz}(\vect{q})} | h^{\vect{\pi}, \vect{\tau}}_{ruz}(\vect{p}) = h^{\vect{\pi}, \vect{\tau}}_{ruz}(\vect{q})] \, (1 - d_R(\vect{p}, \vect{q})) +\\
                                                                                                                                 &+ P[\beta_{h^{\vect{\pi}, \vect{\tau}}_{ruz}(\vect{p})} = \beta_{h^{\vect{\pi}, \vect{\tau}}_{ruz}(\vect{q})} | h^{\vect{\pi}, \vect{\tau}}_{ruz}(\vect{p}) \neq h^{\vect{\pi}, \vect{\tau}}_{ruz}(\vect{q})] \, d_R(\vect{p}, \vect{q}).
                \end{split}
            \end{equation}
            The probability that $\beta_{h^{\vect{\pi}, \vect{\tau}}_{ruz}(\vect{p})}$ equals $\beta_{h^{\vect{\pi}, \vect{\tau}}_{ruz}(\vect{q})}$ is $1$ when $h^{\vect{\pi}, \vect{\tau}}_{ruz}(\vect{p}) = h^{\vect{\pi}, \vect{\tau}}_{ruz}(\vect{q})$, and $\frac{1}{b}$ when $h^{\vect{\pi}, \vect{\tau}}_{ruz}(\vect{p}) \neq h^{\vect{\pi}, \vect{\tau}}_{ruz}(\vect{q})$.
            Therefore,
            \begin{equation*}
                    P[\beta_{h^{\vect{\pi}, \vect{\tau}}_{ruz}(\vect{p})} = \beta_{h^{\vect{\pi}, \vect{\tau}}_{ruz}(\vect{q})}] = 1 - d_R(\vect{p}, \vect{q}) + \frac{1}{b} \, d_R(\vect{p}, \vect{q}) = 1 + \frac{1 - b}{b}d_R(\vect{p}, \vect{q}),
            \end{equation*}
            and this proves~\eqref{eq:correlation}.
        \end{proof}
        
        \begin{figure}
            \centering
            \begin{subfigure}[t]{.25\linewidth}
                \centering
                \includegraphics[width=\linewidth]{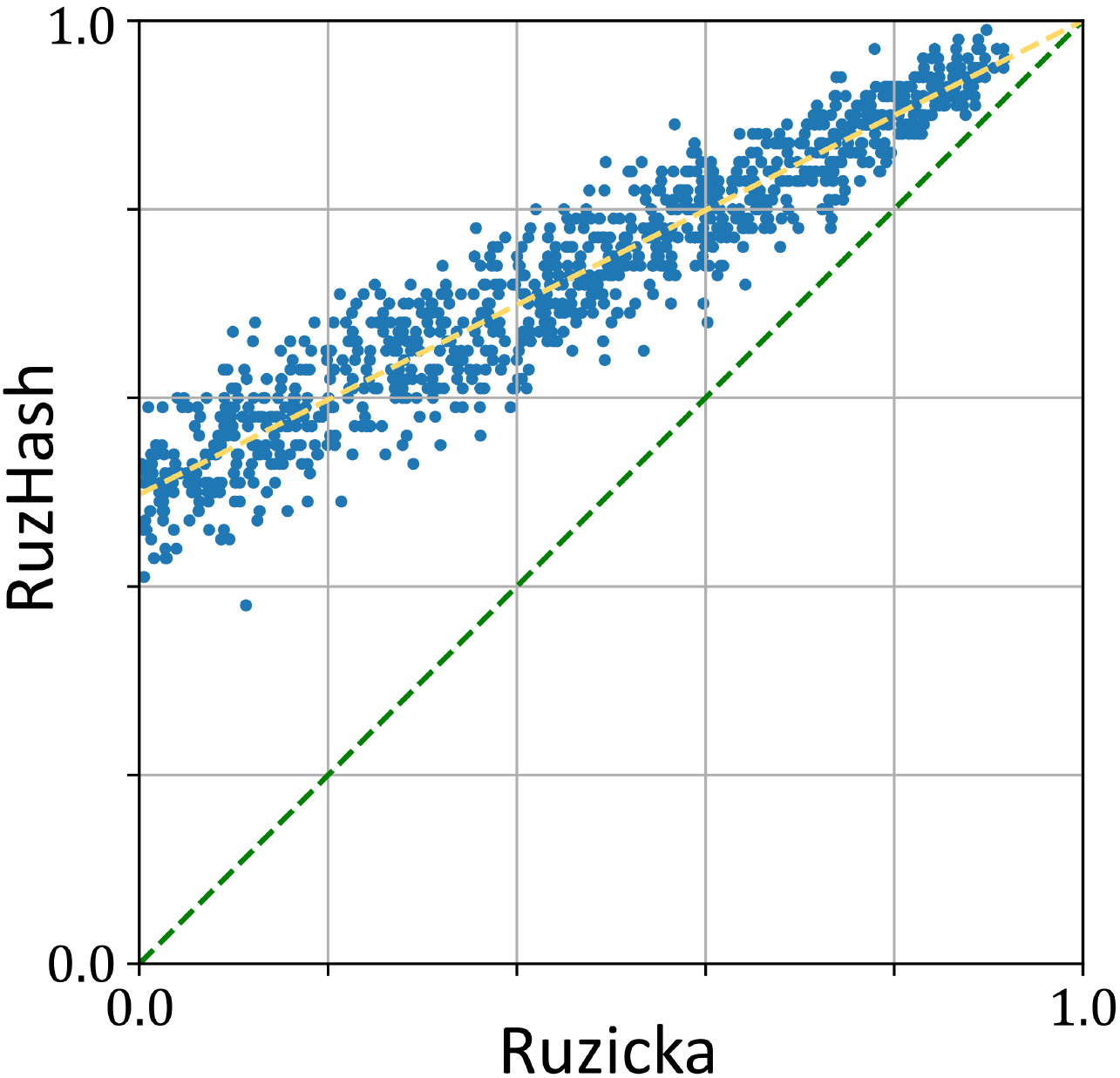}
                \caption{$b = 2$}
            \end{subfigure}
            \hfill
            \begin{subfigure}[t]{.25\linewidth}
                \centering
                \includegraphics[width=\linewidth]{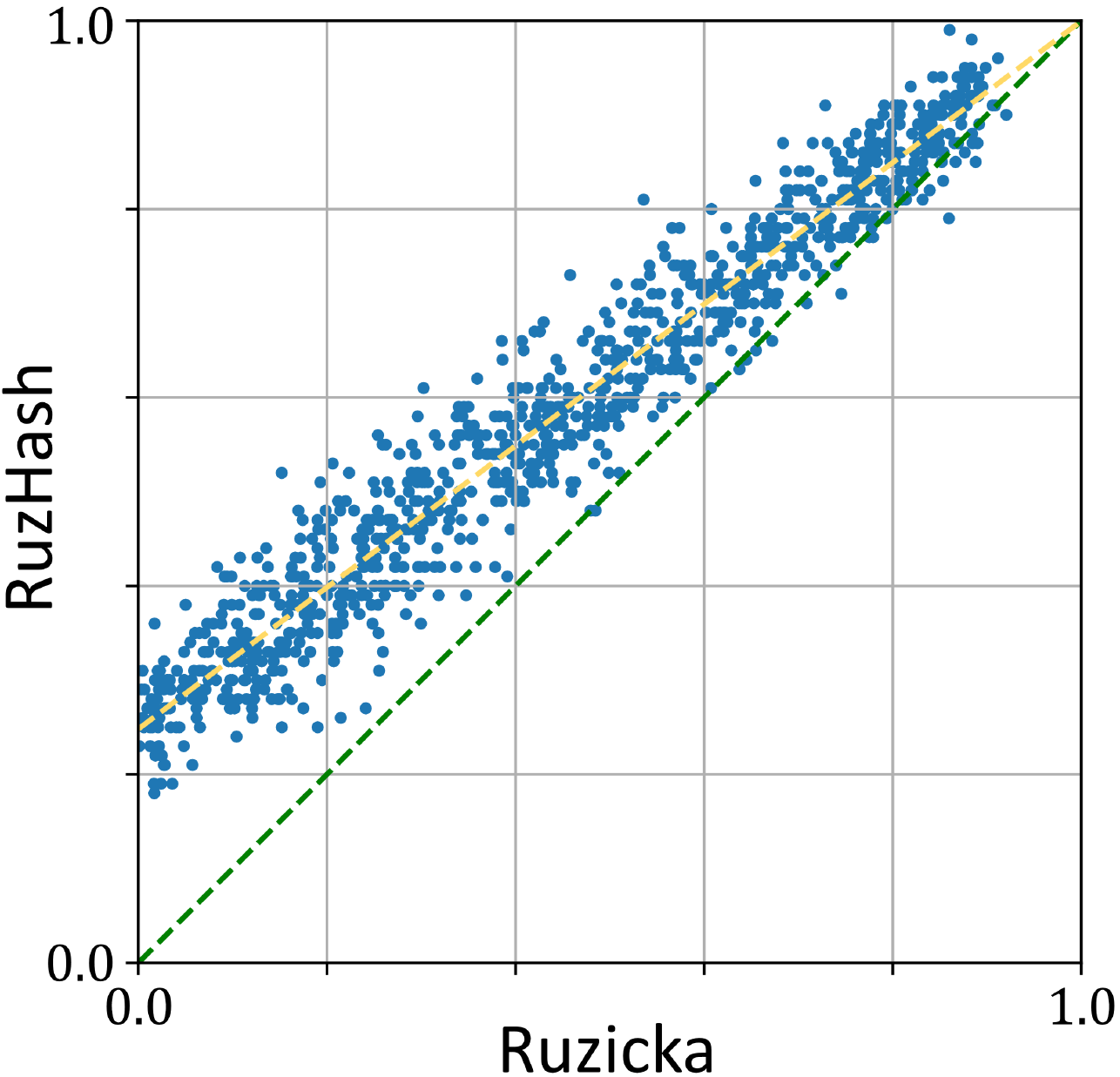}
                \caption{$b = 4$}
            \end{subfigure}
            \hfill
            \begin{subfigure}[t]{.25\linewidth}
                \centering
                \includegraphics[width=\linewidth]{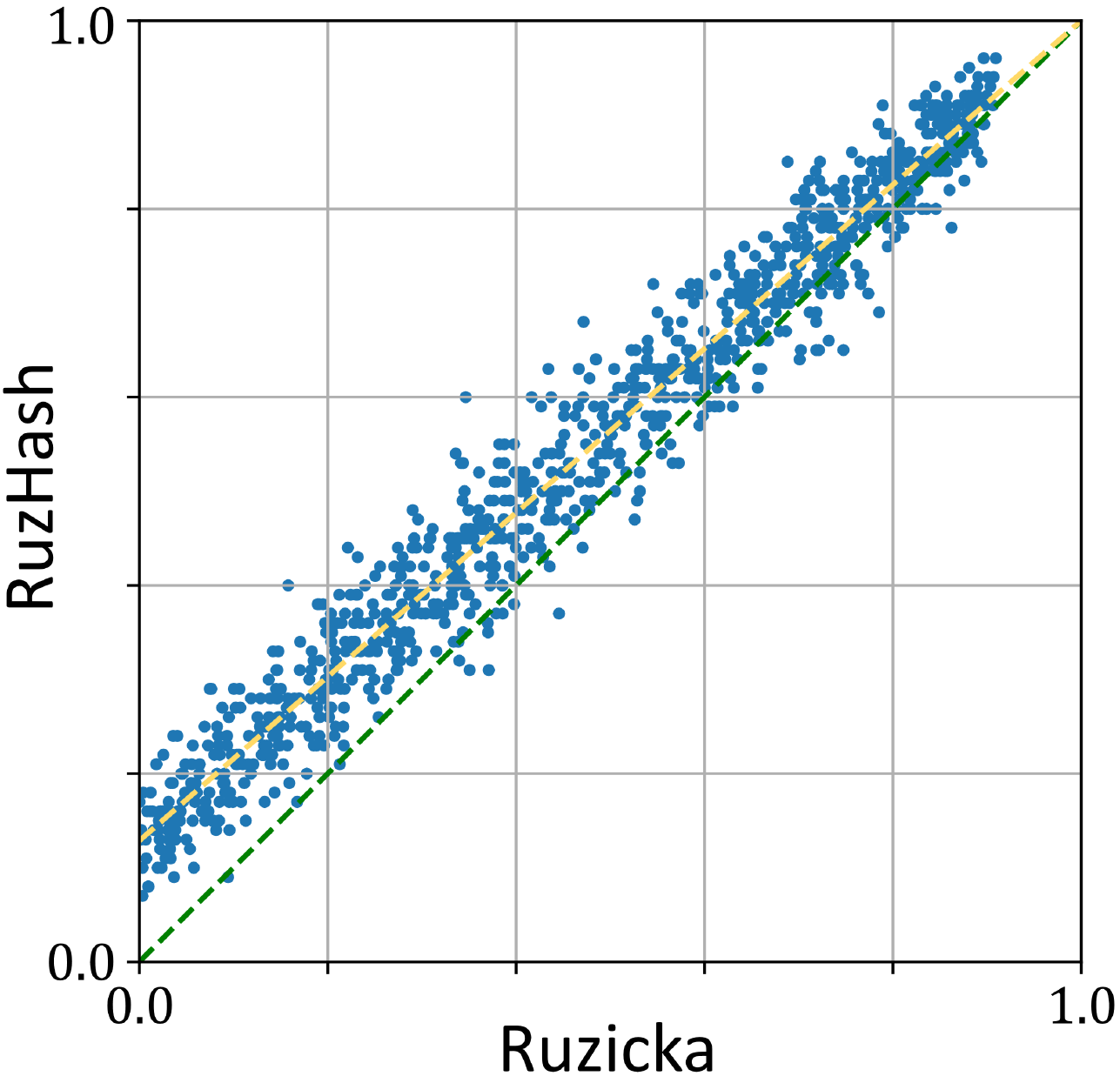}
                \caption{$b = 8$}
            \end{subfigure}
            \caption{Correlation between Ruzicka and \rzhash at various $b$ values, estimated from $n = 1000$ random couples of points $\vect{p}, \vect{q} \in [0, 1]^m$, when $m = 10000$ and $k = 100$.}
            \label{fig:ruzicka_variable}
        \end{figure}
        
        \begin{corollary}[$d_J(\vect{p}, \vect{q})$ linearly correlates with $b$]
            Substituting \rzhash maps $h^{\vect{\pi}, \vect{\tau}}_{ruz}(\vect{p}), h^{\vect{\pi}, \vect{\tau}}_{ruz}(\vect{q})$ with \mhash maps $h^{\vect{\pi}}_{min}(\vect{p}), h^{\vect{\pi}}_{min}(\vect{p})$ in~\cref{thm:correlation}, and given~\cref{lem:minhash}, the proof is equivalent. Therefore,
            \begin{equation*}
                    P[\beta_{h^{\vect{\pi}}_{min}(\vect{p})} = \beta_{h^{\vect{\pi}}_{min}(\vect{q})}] = 1 + \frac{1 - b}{b}d_J(\vect{p}, \vect{q}).
            \end{equation*}
        \end{corollary}

    \bibliographystyle{elsarticle-num}
    \bibliography{references}
\end{document}